\documentclass[runningheads]{llncs}
\usepackage{graphicx}

\usepackage{tikz}
\usepackage{comment}
\usepackage{amsmath,amssymb} 
\usepackage{color}
\usepackage{xspace}

\usepackage{enumitem}
\setitemize{noitemsep,topsep=0pt,parsep=0pt,partopsep=0pt}

\usepackage[accsupp]{axessibility}  

\newcommand{\omni}{$360^\circ$}

\newcommand{\cropim}[6]{\includegraphics[trim=#3 #4 #5 #6, clip, width=#1\linewidth]{#2}}

\newcommand{\argmax}[1]{\underset{#1}{\operatorname{arg}\!\operatorname{max}}\;}

\newcommand{\etal}{\textit{et al}.~}

\begin{document}
\pagestyle{headings}
\mainmatter
\def\ECCVSubNumber{5224}  

\title{SelectionConv: Convolutional Neural Networks for Non-rectilinear Image Data}

\titlerunning{SelectionConv: CNNs for Non-rectilinear Image Data}
%
\author{David Hart \and
Michael Whitney \and
Bryan Morse}

%
\authorrunning{Hart et al.}
%
\institute{Brigham Young University, Provo, Utah, USA\\
\email{\{davidmhart,mikeswhitney,morse\}@byu.edu}\\
}
\maketitle

\begin{abstract}
Convolutional Neural Networks have revolutionized vision applications. There are image domains and representations, however, that cannot be handled by standard CNNs (e.g., spherical images, superpixels). Such data are usually processed using networks and algorithms specialized for each type. 
In this work, we show that it may not always be necessary to use specialized neural networks to operate on such spaces. Instead, we introduce a new structured graph convolution operator that can copy 2D convolution weights, transferring the capabilities of already trained traditional CNNs to our new graph network. This network can then operate on any data that can be represented as a positional graph.
By converting non-rectilinear data to a graph, we can apply these convolutions on these irregular image domains without requiring training on large domain-specific datasets.
Results of transferring pre-trained image networks for segmentation, stylization, and depth prediction are demonstrated for a variety of such data forms.
\end{abstract}

\section{Introduction}
\label{sec:intro}
Convolution has been an important operator in image processing practically since its inception, long before the age of deep learning. 
It is the backbone of most modern deep neural networks, and learnable weights in convolution layers lead to incredible capabilities such as classification, object detection, segmentation, stylization, and many others.

Convolution is powerful, but the 
discrete form used for raster images requires dense rectilinear grids, typically Cartesian grids. 
For sparse, discontinuous, or irregular data, discrete raster convolution may not be applicable. 
Methods such as rasterization, interpolation, or padding are commonly used to convert the data into a form suitable for discrete convolution.

Graph convolution is more adaptable to less regularly structured data and is designed to mimic the process of 2D convolution. 
Instead of requiring spatial adjacency, it performs  convolution based on an adjacency matrix that describes the edges that connect nodes to each other in the graph. 
One key difference, however, between traditional and graph convolution is that graph convolution is assumed to be non-orientable, meaning that it cannot treat incoming edges differently based on spatial direction  or the order they are fed into the convolution. 
This is called the \emph{permutation-invariance} constraint of graph convolution~\cite{Bronstein}.


\begin{figure}[t]
\begin{center}
\begin{tabular}{cccc}
\includegraphics[width=.22\linewidth]{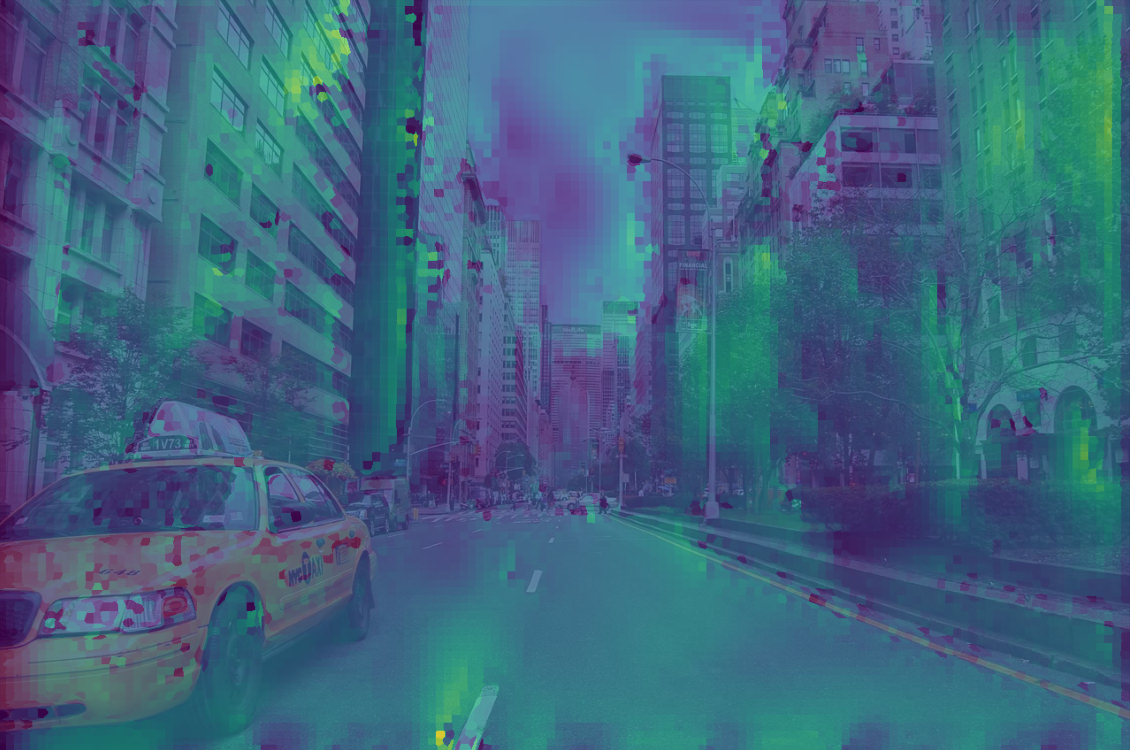} 
&
\hspace*{0.15in}
\includegraphics[width=0.14\linewidth]{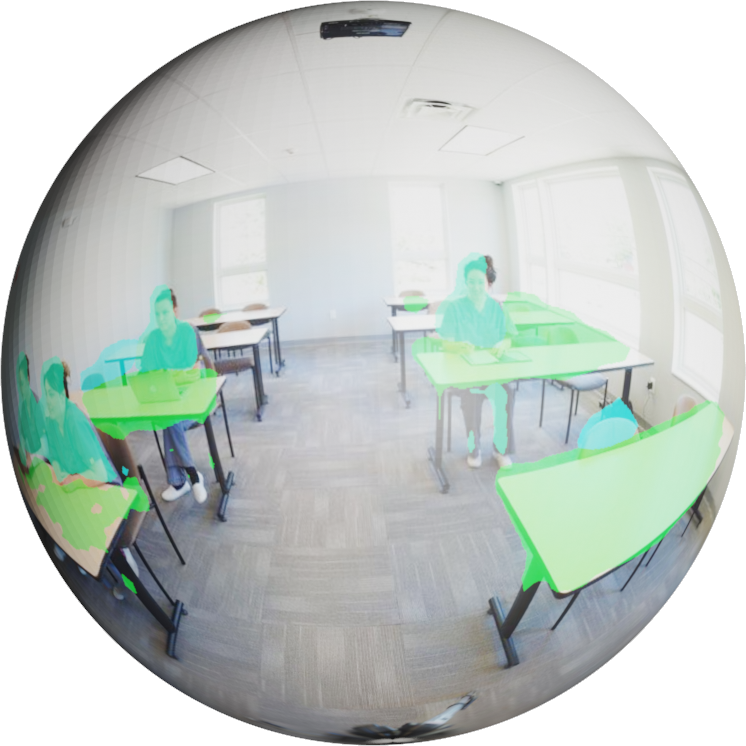} 
&
\hspace*{0.10in}
\cropim{.18}{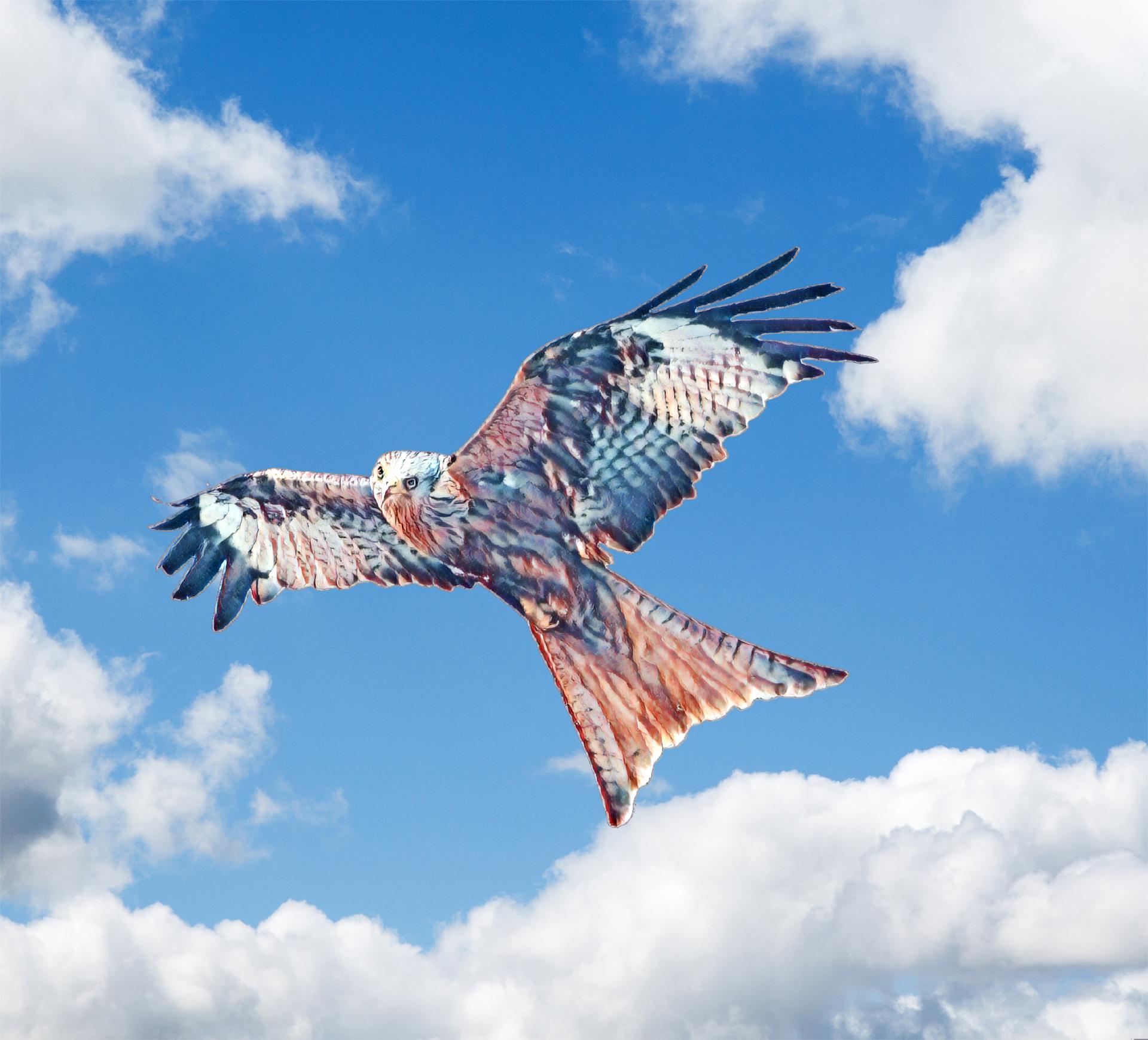}{150}{280}{350}{300} &
\hspace*{0.10in}
\includegraphics[width=.07\linewidth]{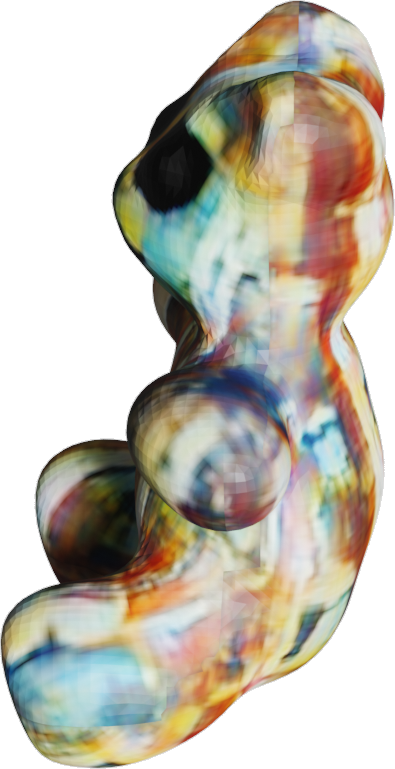}

\end{tabular}
\end{center}
   \caption{
   Our method allows pre-trained 2D CNNs to operate on non-rectilinear image domains such as
   superpixels, spherical images, masked images, and texture maps.
   }
\label{fig:overview}
\end{figure}

In image convolution, neighboring pixels are commonly given different weights to help detect shapes and other patterns. 
In graph convolution, all neighbors are aggregated in the same way, removing any location-based structure in the process. 
Thus, the weights learned in a 2D convolutional neural network are incommensurate with the weights learned in a graph convolution neural network.

Graphs are commonly used to model abstract data that do not have positional information (social networks, individual media ratings, etc.), but when the graph data is image-based,
we still wish to leverage positions, shapes, and patterns in the same manner as traditional image convolution.

This paper presents a framework for working with non-rectilinear image-based data that both traditional convolutional networks and graph networks are
ill-equipped to handle natively, including the types shown in Fig.~\ref{fig:overview}.
We do this using a new type of selection-based graph convolution, which we name \emph{SelectionConv}, that can retain the same shapes and patterns that a traditional convolution learns. 
In so doing, traditional convolution weights can be made commensurate with SelectionConv weights, allowing the transfer of weights directly from networks previously trained on standard image datasets. 
Thus, no special training or fine-tuning is necessary to run the network on less conventional image types.
This is particularly significant because less common image types usually have far less available training data than typical image datasets.


Through this method, any network that operates on images can operate on any form of data that can be represented as a positional graph. 
This allows one framework to perform multiple tasks, such as depth prediction on superpixels, segmentation of spherical images, and stylization of texture maps for 3D meshes, as described in Sec.~\ref{sec:Examples} and demonstrated in Sec.~\ref{sec:Experiments}. 
This technique opens up a realm of possibilities for previously underused data sources.

In summary, our contributions are as follows:
\begin{itemize}
    \item We present a selection-based graph convolution operator that assigns different weights to incoming edges without violating  permutation invariance.
    \item We demonstrate how to transfer pre-trained 2D convolution weights to our new graph operator, thus removing the need to retrain the graph network.
    \item We apply this new method to various non-rectilinear image applications to demonstrate its effectiveness.
\end{itemize}

\section{Related Work}

\subsection{Graph Convolution Networks}
The explosion of deep learning in recent years has led to many developments in both Convolutional Neural Networks (CNNs) and Graph Neural Networks (GNNs). 
Graph Convolution Networks (GCNs) started with the work of Kipf \etal, which extended the ideas of CNNs to a general graph structure~\cite{Kipf2017}. 
Improvements on the original method have been proposed including higher-order aggregation structures~\cite{Defferrard2016,Morris2019} and incorporating MLPs in the aggregation step~\cite{Wang2019}. 
For a more complete overview of this line of work, we direct the interested reader to recent surveys of both CNNs~\cite{Li2020} and GNNs~\cite{Wu_2020}.

Interpolated Convolution Networks~\cite{Mao2019} and Spline Convolution Networks~\cite{Fey2018} are designed, like this work, to mimic the process of a traditional CNN on a point cloud or graph. 
However, these two approaches do not provide an explicit method for transferring weights to the new network. 
Additionally, both of these approaches require traditional position-based point clouds, whereas we  show that our method is adaptable to many different data types and can be modified flexibly according to the users' specifications.

The works by Xu \etal~\cite{xu2021} and Zhou \etal~\cite{Zhou2021}
both aim to use the structural and positional information inherent in graphs to improve graph learning. 
These are learnable components that can improve the training and performance of graph networks given sufficient suitable training data.
Our approach aims to explicitly define the graph structure so that additional training is not needed.

\subsection{Transfer Learning}

The goal of transfer learning is to take the information or weights learned from one network and utilize them in another network in some way. 
One common example of transfer learning is to use a CNN backbone trained for a classification network such as VGG-19~\cite{Simonyan15} for another task such as segmentation. 
Many researchers have explored the effectiveness of networks trained for one task when performing another task~\cite{taskology,taskonomy}, and a recent survey of transfer learning techniques can be found in~\cite{Zhuang2021}.
Our work differs from the goal of previous transfer learning literature since our focus is not to transfer weights to a different task, but to a network that operates on a different domain.

It is worth noting that there has been an effort to theoretically unify the various types of neural networks and their various domains by focusing on their invariance and equivariance properties~\cite{Bronstein}. 
To our knowledge, though, no attempt has been made to state these operations in terms of each other and thus make them transferable.

\subsection{Spherical Images, Superpixels, and Texture Maps}\label{sec:rel_tasks}

This paper demonstrates the effectiveness of selection-based graph convolution
on various forms of non-raster data. 
Here we include work relevant to the tasks we perform and the types of data used.

We demonstrate working with spherical images by performing both semantic segmentation and stylization. 
Several groups have worked on performing semantic segmentation on spherical images~\cite{Jiang2019,Tateno_2018_ECCV,Zhang2019}.
Notable is the work of Tateno \etal\cite{Tateno_2018_ECCV} who developed distortion aware convolutions that operate in spherical space and also have the ability to transfer weights from a standard 2D CNN.
However, their method is specific to spherical images where ours extends to other image domains.

Ruder \etal\cite{Ruder2018} present a method for performing style transfer on 360$^\circ$ images by taking the six cube-projected views and stylizing each one in turn while enforcing consistency between each previously stylized view. 
We also perform spherical image stylization in Sec.~\ref{sec:SphereStyle}, but we do so in a single feed-forward step without the need of fine-tuning a specialized network.

The aim of superpixels is to group similar pixels in an image into regions, simplifying the representation of the whole image. In this work, we use SLIC, a standard baseline algorithm for generating superpixels~\cite{SLIC}. Many other classical approaches exist for generating superpixels~\cite{SuperpixelsSurvey}.
Recent deep learning techniques have been proposed to improve superpixels \cite{Lin2021DeepSC,Verelst2019}. Superpixels have also been used to improve modern detection and segmentation methods~\cite{jampani18,Zhao2021ICCV}. Yang \etal\cite{Yang2020Superpixel} use superpixels to take low-resolution results into higher resolutions, similar to the task we perform in Sec.~\ref{sec:superdepth}, but they do so by pretraining a separate network that can predict superpixel associations on a grid-based structure.

Most work on meshes has focused on learning from the geometry rather than from the color information that is provided in the texture map. 
Some have explored generating new texture from a smaller example texture through classical texture synthesis methods~\cite{ashikhmin2001,efros2001,wei2000} and neural approaches~\cite{fruhstuck2019,gatys2015texture,shi2020,snelgrove2017,zhou2018texture}. 
Textures have also been manipulated through lighting-based style transfer~\cite{StyLit,StyBlit} rather than through a purely image-based approach like the one we present in Sec.~\ref{sec:TextureStyle}. 
Yin \etal\cite{yin2021} recently proposed a geometry and texture stylization approach that is optimization based and uses differentiable rendering, a fundamentally different approach to operating on texture map data than the one we explore in this work.

\section{Selection-Based Convolution}\label{sec:Selection}
Our method requires designing a graph convolution operator that treats incoming edges differently from one another during the aggregation step. 
Traditionally in graph convolution, all connecting edges are specified in a single adjacency matrix, and this matrix is used to described which nodes can influence each other after some set of transformation operations. 
For example, the original Graph Convolution Layer defined in~\cite{Kipf2017} can be described as
\begin{equation}
    \mathbf{X}^{(k+1)} = \tilde{\mathbf{A}} \mathbf{X}^{(k)} \mathbf{W}
\end{equation}
where $\mathbf{X}^{(k)}$ is the current node activations, 
$\tilde{\mathbf{A}}$ is a normalized adjacency matrix, 
and $\mathbf{W}$ is the learned weights. 
Note that the weight matrix is applied equally to all nodes, making the result invariant to the order that nodes are enumerated in $\mathbf{X}^{(k)}$ as long as $\tilde{\mathbf{A}}$ changes correspondingly. 
This is an example of the permutation invariance constraint to which all graph convolution operations must adhere.

In comparison, while standard 2D convolution is shift invariant, it is not permutation invariant, relying heavily on orientation when assigning the weight of each connecting pixel. 
For example, the pixel directly above the current one might be given different weight than the pixel to the bottom right, and so on.
Graph convolutions are able to assign edge weights, but they are generally static or on a node-by-node basis using a mechanism such as attention~\cite{velickovic2018}.
Thus, the weights learned during a 2D convolution are incommensurate with the weights learned during a graph convolution.

In order to leverage the benefits of pretrained 2D convolutional networks while having the structural flexibility of graphs, we introduce a new graph convolution layer
that can preserve location information.
We do so by preprocessing the graph
into multiple adjacency matrices, 
selecting edges to be assigned to different matrices based on the spatial relationship between their two nodes.
This is similar to the way we can think of different adjacency relationships between pixels and their directional neighbors.
There is also a unique weight matrix for each corresponding adjacency matrix so only those edges are affected.
The results for the set of adjacency and weight matrices are summed together to make the final activation. 
This selection-based convolution is what we call \emph{SelectionConv}. 

\begin{figure}[t]
\begin{center}
\includegraphics[width=0.50\linewidth,trim=0 10 0 105,clip]{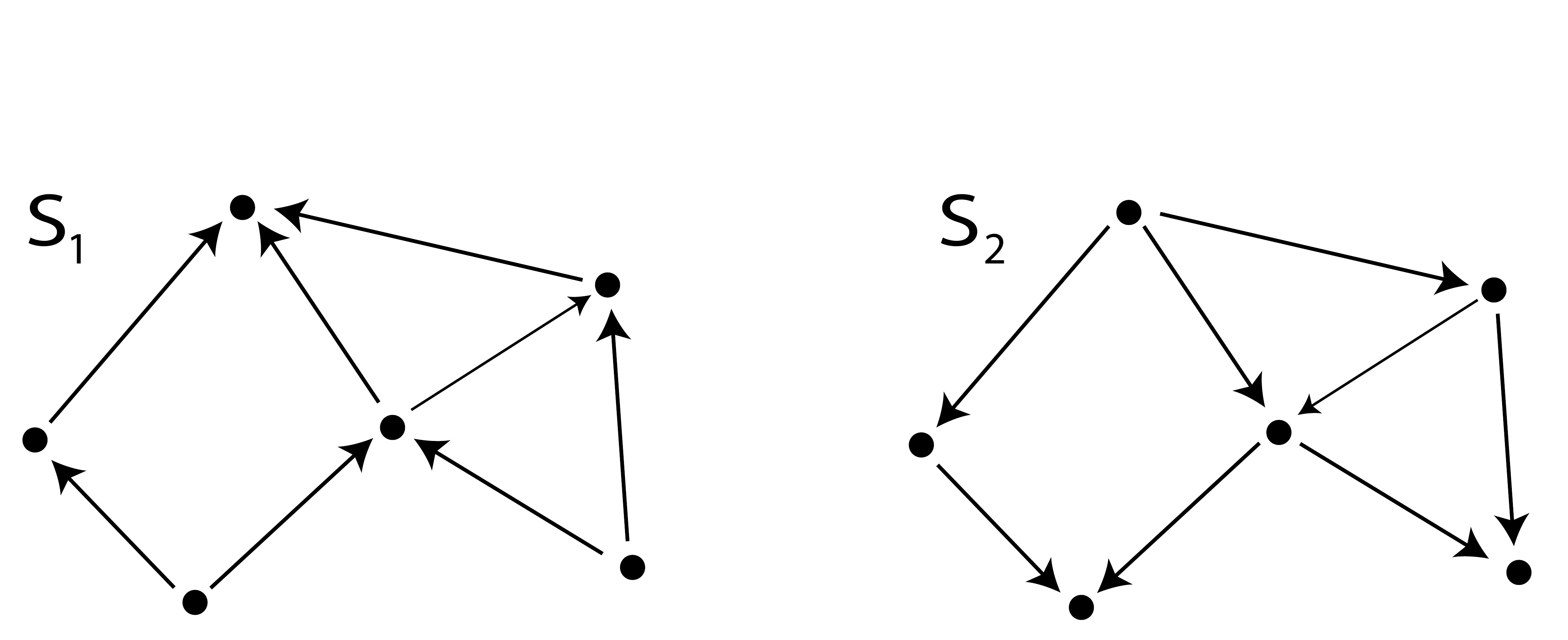}
\end{center}
   \caption{
   A graph with a selection function that weights upwards edges differently from downwards edges. Such a selection function would give two adjacency matrices, $\mathbf{S}_1$ and $\mathbf{S}_2$, that would be applied to two different weight matrices, $\mathbf{W}_1$ and $\mathbf{W}_2$.
   }
\label{fig:SelectionGraph}
\end{figure}

For each graph, a given edge $e_{ij}$ needs to be assigned to its specific adjacency matrix.
We do so using a selection function $s(v_i,v_j)$ for vertices $v_i$ and $v_j$ with spatial positions $\mathbf{x}_i$ and $\mathbf{x}_j$ respectively, which indicates which adjacency matrix includes the edge $e_{ij}$ between these vertices. 
For $m$ possible selections this gives $m$ adjacency matrices respectively defined as
\begin{equation}
    \label{eq:selection-adj}
    \mathbf{S}_{m_{ij}} = 
        \begin{cases}
                 1 & \text{ if } s(v_i,v_j) = m \\
                 0 & \text{otherwise}
        \end{cases}
\end{equation}
Each selection has a corresponding weight matrix. Thus our convolution becomes
\begin{equation}
    \label{eq:selection-conv}
    \mathbf{X}^{(k+1)} = \sum_m{\tilde{\mathbf{S}}_m \mathbf{X}^{(k)} \mathbf{W}_m}
\end{equation}
where $\tilde{\mathbf{S}}_m$ is the normalized version of $\mathbf{S}_m$ to account for nodes having multiple edges with the same selection $m$.
With this structure, nodes can be treated differently based on location and other features relative to the current node without breaking the permutation invariance constraint. An example of this process is illustrated in~Fig.~\ref{fig:SelectionGraph} .


We use PyTorch Geometric~\cite{Fey2019} to implement this process and use slices from 3D tensors rather than separate weight and adjacency matrices for efficiency, but the result is mathematically equivalent.

\section{Selection-based Convolution for Images}\label{sec:ImageSection}

Though the process described in Sec.~\ref{sec:Selection} is general enough to work with any number of selections based on any number of node attributes, we now move to the specific case of working in the image domain. 
To start, we will establish a baseline for our method by looking at a regular image and showing that we get results identical to a 2D convolution with our selection-based convolution.

\subsection{Setting Up Image Graphs}\label{sec:Images}

An image with pixels in a Cartesian grid can be thought of as a set of nodes at equally spaced distances. 
Many neural networks use $3 \times 3$ convolutions as their primary image feature extractor, which look at the pixel and its 8-connected neighbors. Thus, when we construct our graph, we will add an edge from each pixel to its 8-connected neighbors and one to itself. 
This means we need $m = 9$ possible selections. 

For pixels in a Cartesian grid arrangement, value assignment for the respective selection functions is straightforward.
For the general spatial case, we project the vector defined by the position of the two nodes onto the set $\mathbf{D}$ of unit vectors in each of the respective neighbor directions. Specifically, 
{\small
\begin{equation}
    \label{eq:directions}
    \mathbf{D} := 
    \begin{matrix} 
        \left<-\sqrt{2}/2,-\sqrt{2}/2\right> & \left<0,-1\right> & \left<\sqrt{2}/2,-\sqrt{2}/2\right>\\ 
        \left<-1,0\right> &   & \left<1,0\right> \\  
        \left<-\sqrt{2}/2,\sqrt{2}/2\right> & \left<0,1\right> & \left<\sqrt{2}/2,\sqrt{2}/2\right>
    \end{matrix}
\end{equation}}

\noindent Whichever directional unit vector results in the largest projection (resulting dot product) corresponds to the assigned selection. 
Additionally, if the positions are the same (or within some small radius), the central selection is made. Thus, our selection function becomes
{
\small
\begin{equation}
    \label{eq:selection}
    s(v_i,v_j) =
        \begin{cases}
            0 \text{ if } \|\mathbf{x}_j - \mathbf{x}_i\| < \epsilon \\
            \argmax{k} \mathbf{D}_k \cdot \left( \mathbf{x}_j - \mathbf{x}_i \right) \ \mbox{otherwise}
        \end{cases}
\end{equation}
}

\noindent For simplicity, when assigning a selection number or index to each direction, we follow the mathematical convention of angles by making the direction to the right be the first selection and moving in the counterclockwise direction for assigning each subsequent direction. This is visualized on the left and right sides of Fig.~\ref{fig:SelectionWeightTransfer}.

\subsection{Weight Transfer from 2D Convolutions}

\begin{figure}[t]
\begin{center}
\includegraphics[width=0.65\linewidth,trim=0 25 0 25,clip]{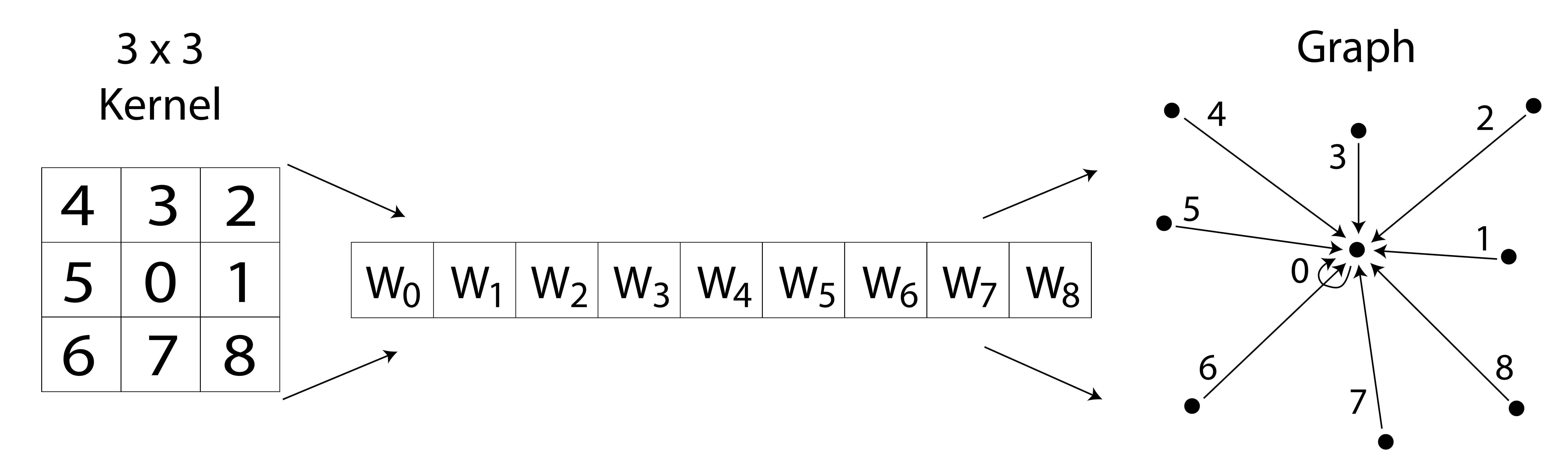}
\end{center}
   \caption{
   Elements of a $3 \times 3$ convolution kernel are enumerated with zero for the center weight and neighboring weights from one to eight in counter-clockwise direction. 
   The different weights from the kernel are transferred to associated weight matrices. Those weight matrices are then applied to the selected edges on the graph. Note that the points in the graph do not need to be equally spaced as in regular images. There can also be more than one node per selection.
   }
\label{fig:SelectionWeightTransfer}
\end{figure}

Once the appropriate selections have been made, transferring the weights is simply copying the appropriate slice of the convolution kernel weights to its assigned selection. 
For example, if selection $5$ represents an edge going to the left, the left kernel convolution weights would be copied to~$\mathbf{W}_5$. 
This process is illustrated in Fig.~\ref{fig:SelectionWeightTransfer}.
When applied to raster data, this process leads to results that are identical to those using an image-based convolutional network.

\subsection{Handling Larger Kernels} \label{sec:LargeKernel}
If a network uses a kernel that is larger than $3 \times 3$, more weights need to be copied over, but the same graph structure can still be used. 
Rather than using a larger graph 
that is memory inefficient,
we use the 
edges 
of the simpler $3 \times 3$ graph.
Through successive multiplications of the adjacency matrices, multiple edge hops (traversals) can be performed until specific kernel locations are reached.
For example, if a $5 \times 5$ kernel is used, the weight associated with the bottom middle pixel would be assigned to the action of taking the bottom selection's bottom selection.

\subsection{Pooling Operators and Upsampling} \label{sec:Pool}

Many CNNs use pooling layers throughout the network to combine nearby features. While GCNs have similar pooling features, it is important that our SelectionConv network's pooling layers mimic the downsampling nature of those used in traditional CNNs. 

If the network 
contains fully connected layers, we impose a spatial grid onto the set of nodes when pooling. 
This grid matches the spatial size that the original image data reduces to after pooling.
If the network is fully convolutional, pooling does not require a regular grid (especially if one cannot be imposed, such as on a texture map), and any pixel-clustering algorithm can be used. In both cases, any node within a cell or cluster is made into a single node during the pooling step.
The average of the positions of nodes in that cell becomes the position of the pooled node. This is similar to the meta-node approach used in \cite{Qi2017pnplusplus}.


Pooled nodes also need to reestablish edge connections and selections. To do so,
we implement a post-pooling function that makes new edges between the aggregated cluster nodes by using the previous layer's graph edges. 
If any edge exists between two nodes in different clusters, the aggregated nodes of those clusters will have a corresponding edge between them.
That edge is assigned the average of the previous selection values of all the edges between the original nodes in the two clusters (while properly accounting for the selection value wrap-around between 1 and 8).

If the network requires upsampling steps, we save each version of the graph before it is downsampled. When upsampling, we revert back to a previous version of the graph and copy appropriate values to each node using the defined clusters. Bilinear upsampling can also be approximated by using the average value of the connections to new nodes or through other point cloud interpolation methods.

\subsection{Strides, Dilation, Padding} \label{sec:Stride}
Traditional 2D convolution layers often have additional parameters such as the stride of the kernel, the dilation of the kernel, and the padding to be used on the border of the image. 
Selection-based convolution as described so far is equivalent to a stride of 1, a dilation of 1, and zero padding, but we have the ability to mimic these additional features in our SelectionConv network when needed.

Strides larger than 1 in a regular convolution layer are equivalent to a stride equal to 1 followed by downsampling by the stride amount (with no antialiasing). 
We use this same idea to implement large strides in the SelectionConv network. 
The convolution is performed as usual (equivalent to a stride of 1), then the graph is pooled using the method described in Sec.~\ref{sec:Pool}, but rather than using a max or average pooling operator, a predetermined central node in each cluster is always used as the pooled value. 

Dilation is handled in a way similar to the larger kernels described in Sec.~\ref{sec:LargeKernel}. 
The dilation amount defines how many times a selection is edge hopped. 
For example, a dilation of two would indicate that instead of taking the left selection, the left selection's left selection should be used instead. 
This process is repeated for each selection for the same number of times as the dilation's value.

Padding in traditional 2D CNNs helps control the size of output layers.
Graph convolution layers do not change the size of the output since the number of nodes will stay the same, so padding is not usually needed. 
Some 2D CNN padding schemes, however, are used to help information propagate correctly along the borders of an image (such as reflective padding in stylization networks). 
We handle these situations not only by looking at nodes along a border, but by determining what to do with any missing selections. 
The following padding methods can be implemented effectively and approximate padding schemes used for images:
\begin{itemize}
    \item \textbf{Zero}: Missing selections are not considered (default).
    \item \textbf{Constant}: Missing selections are assigned a predetermined value.
    \item \textbf{Replicate}: Missing selections are assigned the value of the current node.
    \item \textbf{Reflect}: Missing selections are assigned the value of the selection in the opposite direction.
\end{itemize}

\noindent Some of these steps we mimic from CNNs currently have nondifferentiable implementations. 
Though not needed for the scope of this work, which focuses on transferring weights from pre-trained networks, all parts of our SelectionConv network would need to be differentiable if any form of training or fine-tuning was desired for the network after weights have been transferred. 

\subsection{Verification}

To verify that the SelectionConv network can truly be equivalent to a traditional CNN, we used a pre-trained VGG-11 network~\cite{PytorchModels,Simonyan15} on CIFAR-10~\cite{CIFAR10} and transferred the original weights to our network.
We compared this to the original image-based network using the 10,000-image validation set. 
As expected, the two methods resulted in identical predictions and identical accuracies of 84.5\%. 
This remains true even when small random spatial perturbations are introduced to the points in the graph structure.

\section{Example Non-rectilinear Configurations}\label{sec:Examples}

Sec.~\ref{sec:ImageSection} describes how to configure a SelectionConv network to work on images in exactly the same way as a regular CNN. 
This works as a baseline but does not provide any additional power over a regular CNN. 
In this section, we give examples of the flexibility of our method to work with data that cannot be processed with a traditional CNN due to its irregular structure. 
{\em Importantly, our method can use the weights from a CNN pre-trained using standard image datasets without the need to retrain for specific data types.}



\subsection{Panoramic and Spherical Images}
\label{subsec:360}

Many smartphones and cameras allow users to take single or multiple pictures of their surroundings to acquire panoramic or even up to the full 360$^\circ$ $\times$ 180$^\circ$ view of their environment. 
These panoramic and spherical images have non-simple topologies but are typically stored as simple planar images. 
This requires projection of the content to a surface of some kind. 
While there are many ways to project, including spherical, equirectangular (cylindrical), and cubic, each of these have distortion or seams of some nature that would be difficult for a traditional CNN to handle due to irregular spatial sampling or topological considerations. 
With our method, such seams and distortion can be handled by proper construction of the graph.

\begin{figure}[t]
\begin{center}
\begin{tabular}{rccrc}
\raisebox{0.31\linewidth}{a)}\! &
\includegraphics[width=0.45\linewidth]{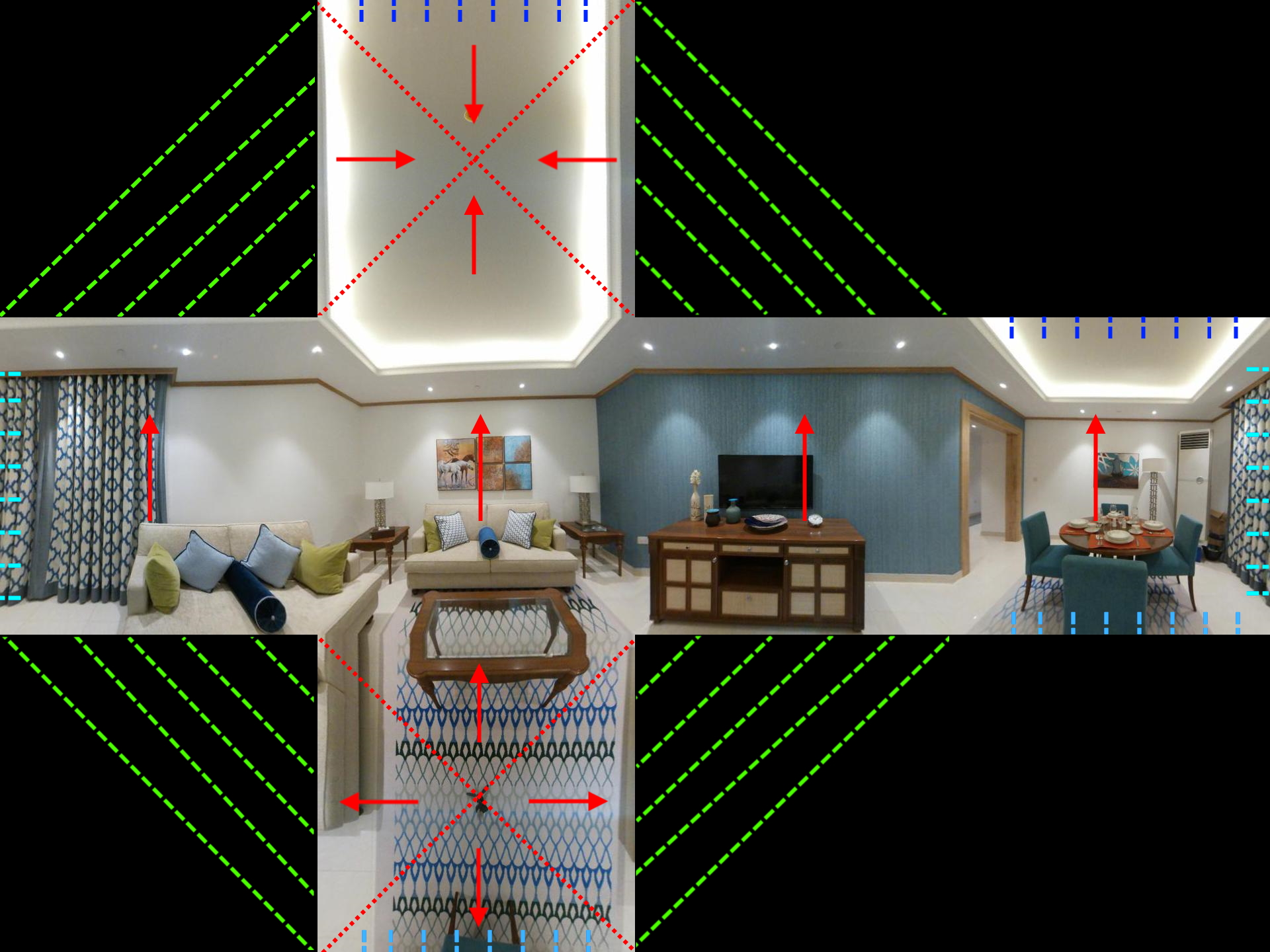} &
\hspace{0.3 in}
\raisebox{0.31\linewidth}{b)}\! &
\includegraphics[width=0.28\linewidth]{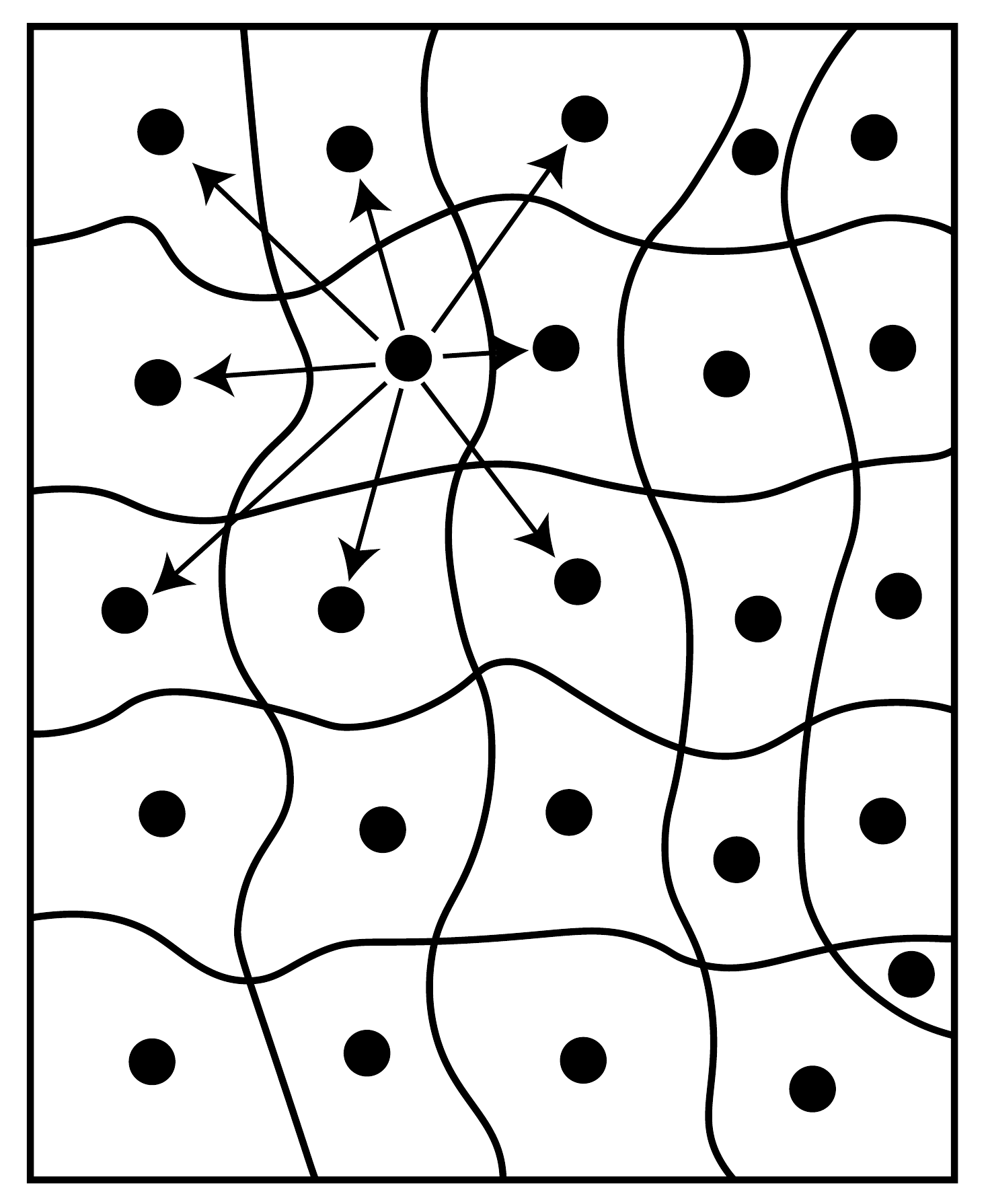}
\end{tabular}
\end{center}
   \caption{a) Illustration of the graph connections for a cube map. Red arrows represent the upwards selection for nodes in that part of the graph. Green and blue lines represent connections made in the graph between faces. b) Illustration of the graph connections between the centroids of superpixels in an image.
   }
\label{fig:cubeseams}
\end{figure}

As an example, we show how to construct the graph for a cubic projection often used for environment maps in computer graphics, which we have found to be effective to work with since it has low levels of distortion compared to other projections. 
Seams are naturally present along each edge of the cube, and a 2D image can only represent a few of those connections. 
In our graph, we simply need to make the connections between the rest of the seams, as illustrated in Fig.~\ref{fig:cubeseams}.a. 
Additionally, we orient our selection function in such a way so that the upwards selection is always pointed towards the top pole of the map. 

\subsection{Superpixel Images}
Using superpixels is a common approach for simplifying a high-resolution image by representing it as a smaller set of similar regions. Because of the irregular structure of the regions, superpixels cannot be used for standard CNNs. They can, however, be easily represented as a graph and used with our approach.

To construct the graph from the set of superpixels, the centroid of each region is treated as a node, and edges are selected using a K-nearest neighbors method. Selections are then made using the dot product method described in Sec.~\ref{sec:Images} and the graph is pruned so only the closest neighbor to each node for a given selection is used. This process is illustrated in Fig.~\ref{fig:cubeseams}.b.

\subsection{Masked Images and Texture Maps}
\label{subsec:Texture}

Many applications require operating only on the foreground or a specified region of the image. 
Our graph construction can naturally handle these cases by simply dropping nodes and edges that are not part of the desired region, and any desired padding mode described in Sec.~\ref{sec:Stride} can be used to handle the irregular border.

\begin{figure}[t]
\begin{center}
\hspace*{-0.11in}
\begin{tabular}[t]{rcrcrc}
\raisebox{0.14\linewidth}{a)}\!\! &
\includegraphics[width=.14\linewidth,trim=0 10 0 0,clip]{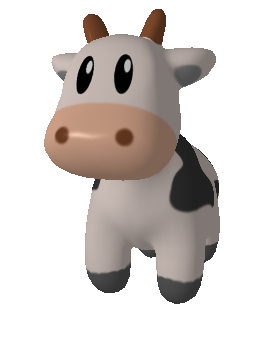} 
\hspace*{0.025in}
&
\raisebox{0.14\linewidth}{b)\!} &
\includegraphics[width=.165\linewidth]{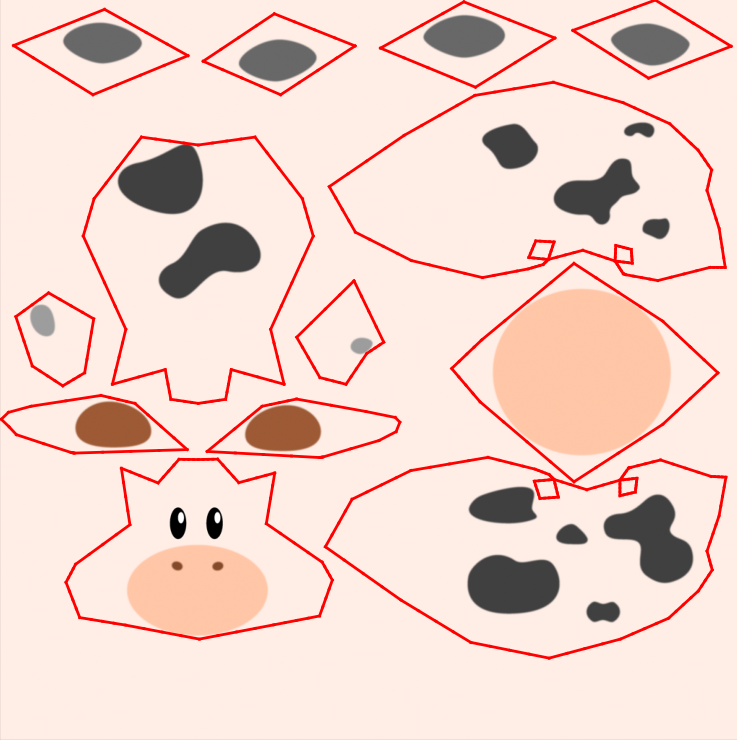} 
\hspace*{0.1in}
&
\raisebox{0.14\linewidth}{c)\!} &
\includegraphics[width=.165\linewidth]{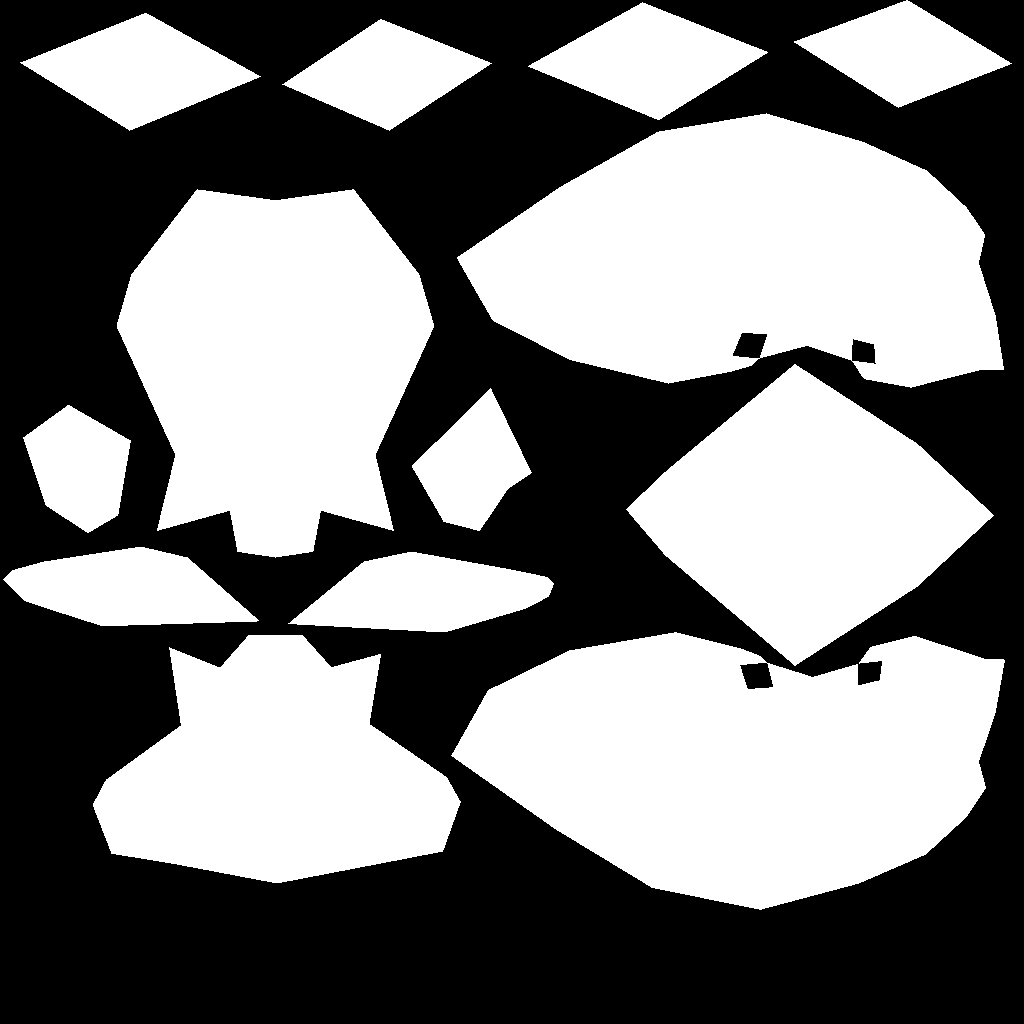}
\end{tabular}
\end{center}
   \caption{A 3D model~(a) and its texture map. The model's texture seams can be determined from UV coordinates and represent boundaries on the texture map~(b). From this, we construct a mask of relevant pixels~(c) and connect discontiguous regions.
   }
\label{fig:texturemap}
\end{figure}

Texture maps for 3D meshes can also be thought of as masked regions. 
Not all pixels present on a texture map image will be used to determine colors on the actual mesh, so we can mask out the regions that contain pixels that represent some part of a face on the mesh. 
If we further connect these faces in the graph, we can operate on the texture map in the same fashion as any image.
To do this in general, we start by determining which edges are only referenced once in the UV map, since these represent the boundaries of groups of faces on the UV map. 
We then find all closed loops of edges to separate each boundary. 
Finally, we do a polygon-contains-point test for each pixel to see if the pixel is inside any of our boundaries. 
This becomes the mask of relevant pixels on the texture map. 
An overview of this process is shown in Fig.~\ref{fig:texturemap}. 

If it is known which geometric faces are connected to each other, this can be built into the graph construction. 
Otherwise, each edge boundary is paired with a corresponding edge on another face that is used to make the connections. 
All regions are connected regardless of where they are located on the image.

\section{Results}\label{sec:Experiments}


To demonstrate transfer from image networks to SelectionConv networks,
we present the results of applying this and the graph-construction methods from the previous section to various applications and image types.
Additional results can be found in the accompanying supplemental materials.

\subsection{Spherical Style Transfer} \label{sec:SphereStyle}

\begin{figure}[t]
\begin{center}
\hspace*{-0.25in}
\begin{tabular}{rcrc}
\raisebox{0.12\linewidth}{a)} &
\includegraphics[width=.30\linewidth]{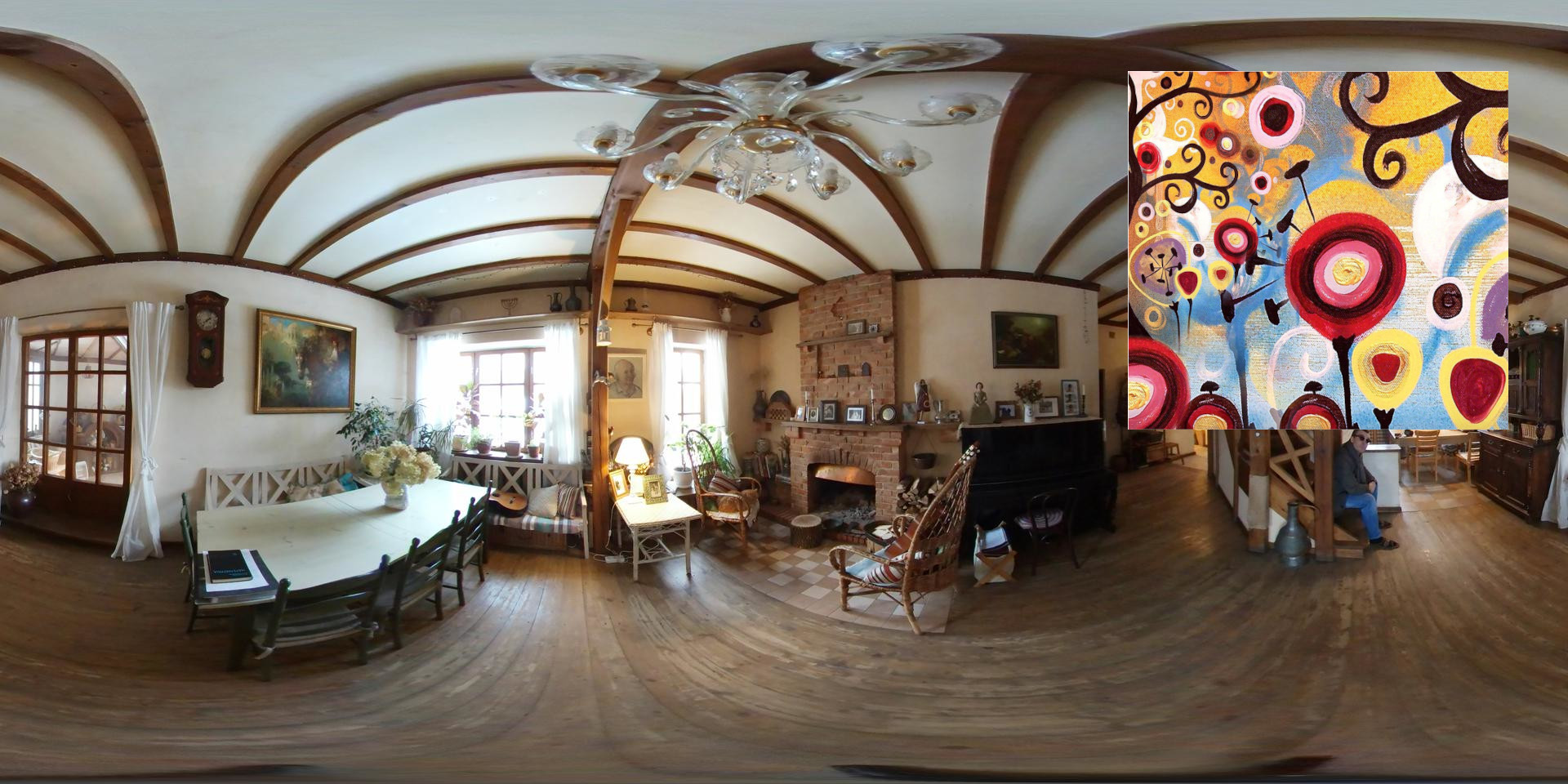} &
\raisebox{0.12\linewidth}{b)} &
\includegraphics[width=.30\linewidth]{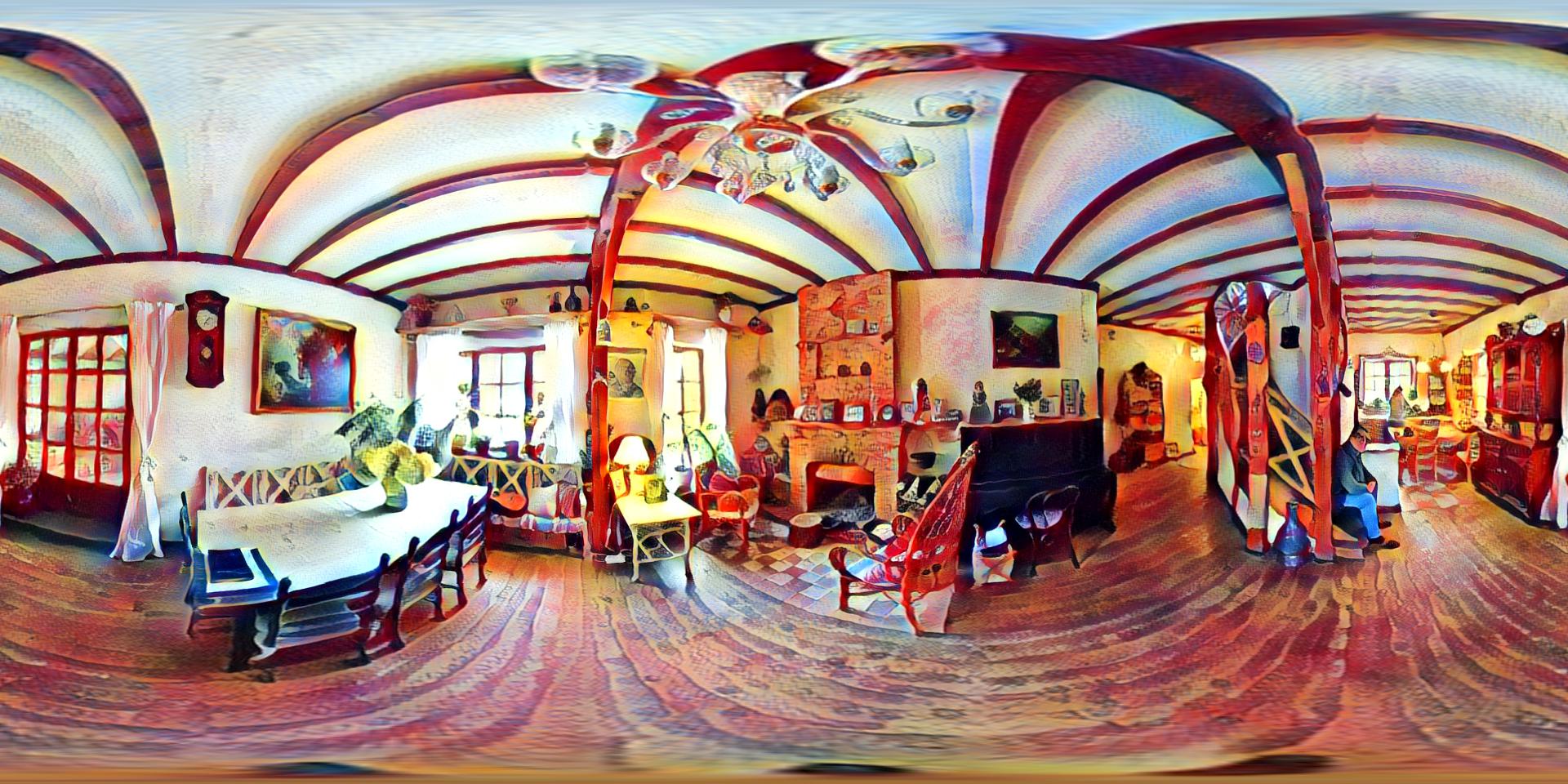} 
\\ 
\end{tabular}
\begin{tabular}{ccc}


\cropim{.26}{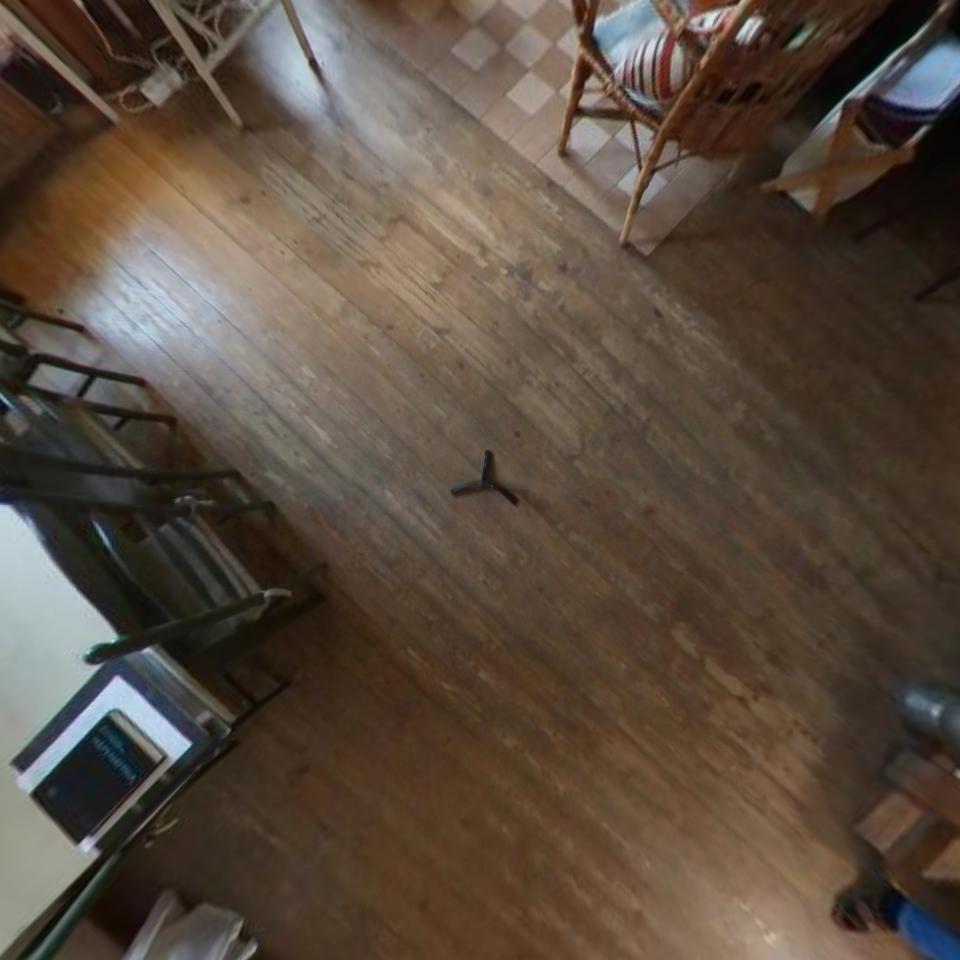}{150}{250}{150}{250} &
\cropim{.26}{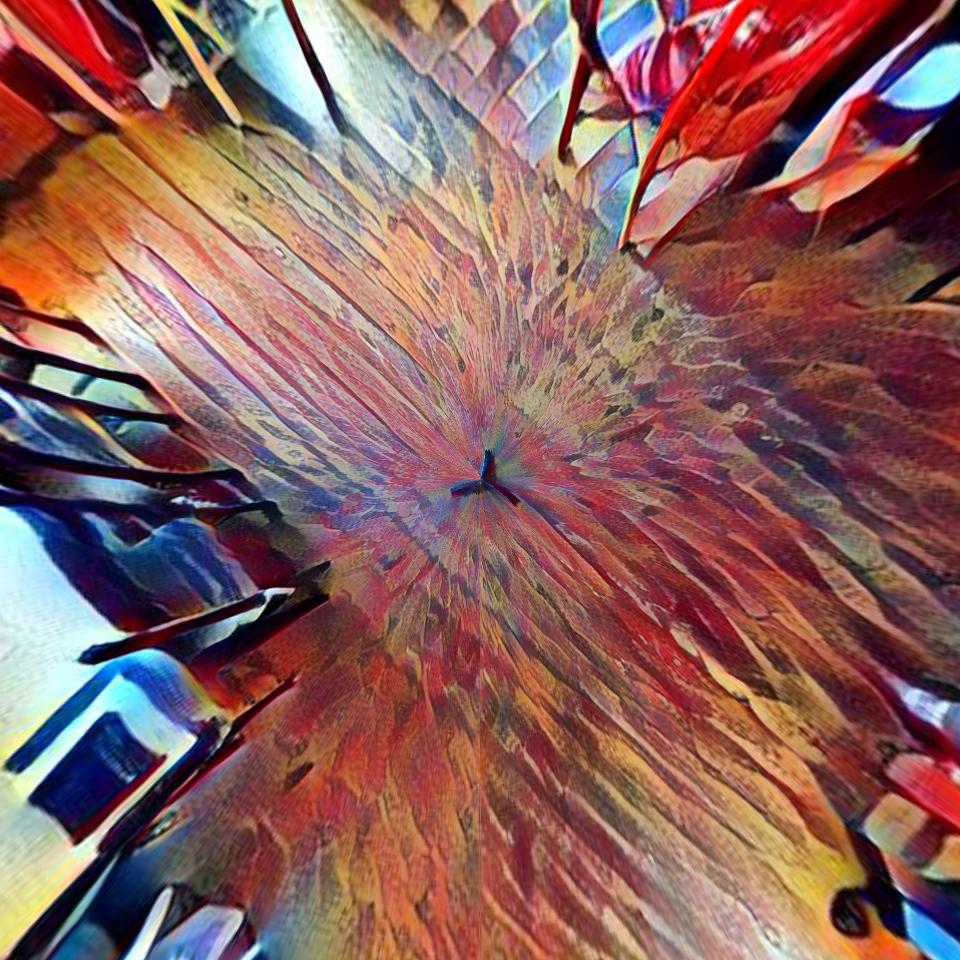}{150}{250}{150}{250} &
\cropim{.26}{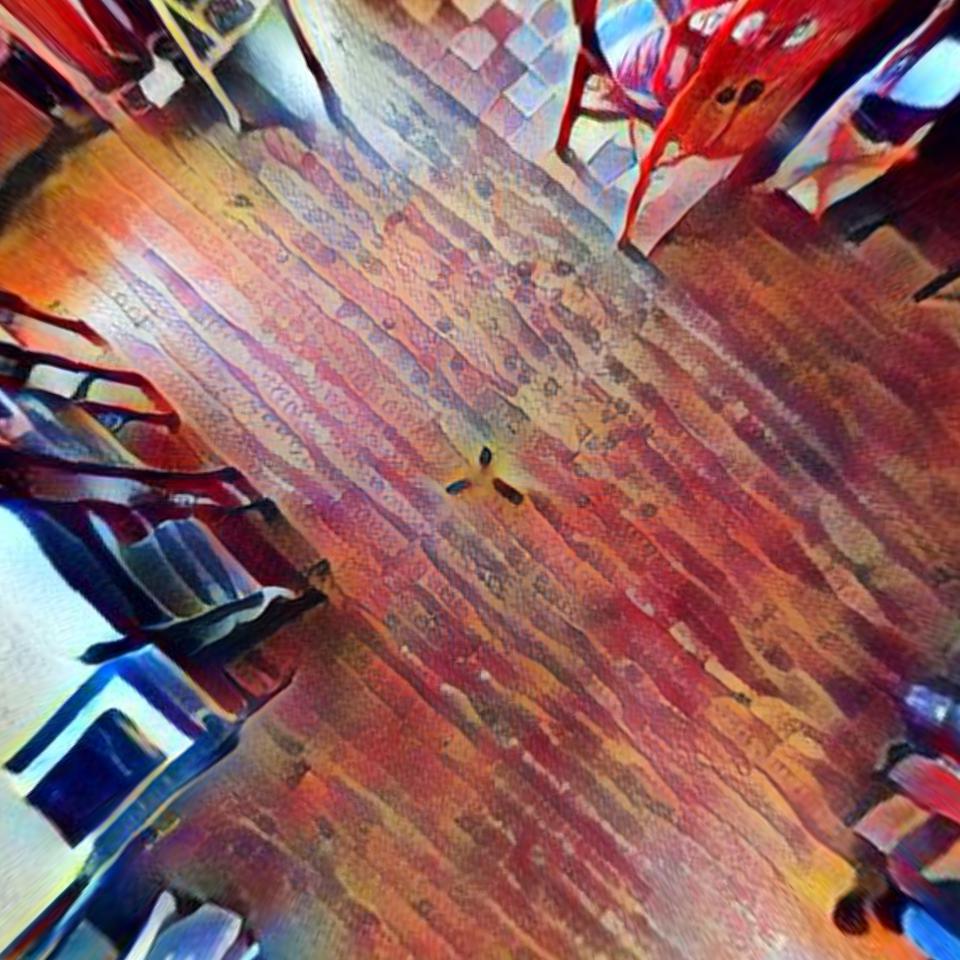}{150}{250}{150}{250}
\\
{(c) Ground View} &
{(d) Naive} &
{(e) SelectionConv } 
\end{tabular}
\end{center}
   \caption{A 360$^\circ$ image (a) and its stylization using our feed-forward method (b). An example view looking downward at the lower pole of the image (c) has seams and distortion when naively stylizing the rectangular image (d), but those seams and distortion are minimized with our method  (e). Image taken from~\cite{indoorDataset}.
   }
\label{fig:spherestyle}
\end{figure}

A simple but  effective illustration of the seamless nature of a selection-based graph convolution is to perform style transfer on a spherical image.
For this we use the feed-forward style transfer approach recently proposed by Li \etal\cite{Li2019}. 
Usually for this task, a special piece-wise optimization approach or fine-tuned network would be needed, such as that proposed by Ruder \etal\cite{Ruder2018}. 
However, by generating our spherical graph using the method shown in Sec.~\ref{subsec:360}, we naturally avoid distortion, minimize seams, and can stylize in a single feed-forward pass. 
An example is shown in Fig.~\ref{fig:spherestyle}. 
The whole process of generating the graph, transferring the weights, and running the graph convolution can run in 15-20 seconds on a consumer GPU, while \cite{Ruder2018} would take 8-10 minutes per \omni{} image. Even the faster approach suggested in \cite{Ruder2018} requires fine-tuning a network for 120,000 iterations per style image. Our approach uses a state-of-the-art feed forward style transfer method, 
can be used for any style image, and still enforce consistency across the seams of the cube map.


\begin{figure}[b!]
\begin{center}
\begin{tabular}{cccc}
\includegraphics[width=.24\linewidth]{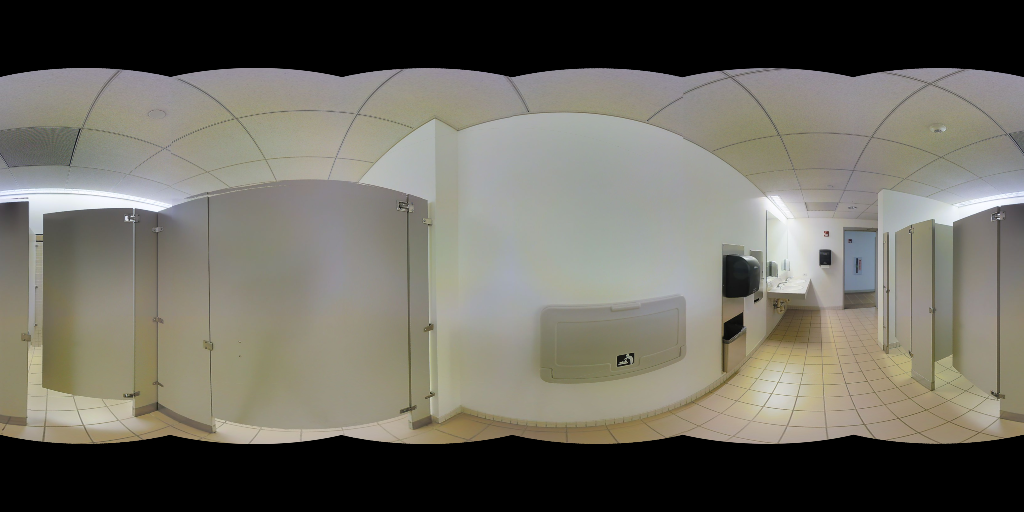} 
&
\includegraphics[width=.24\linewidth]{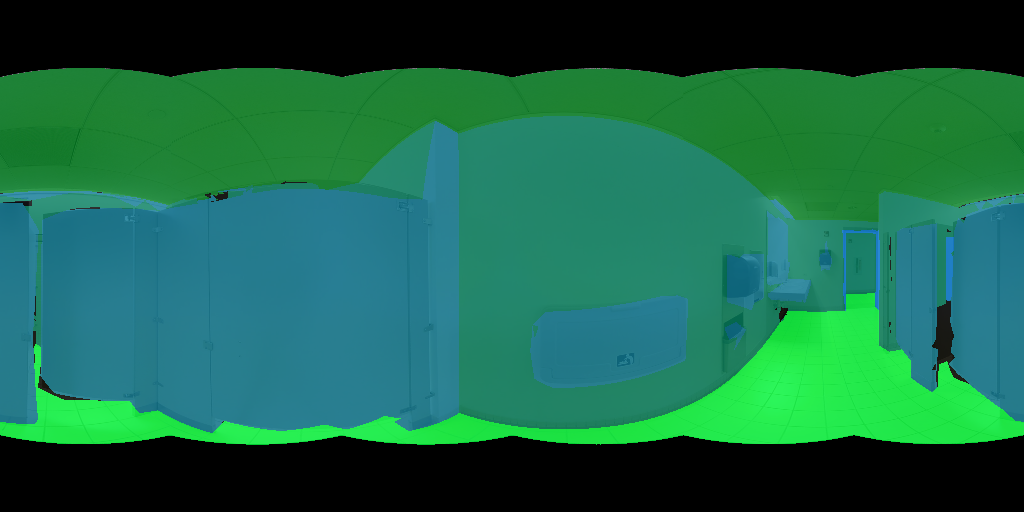} 
&
\includegraphics[width=.24\linewidth]{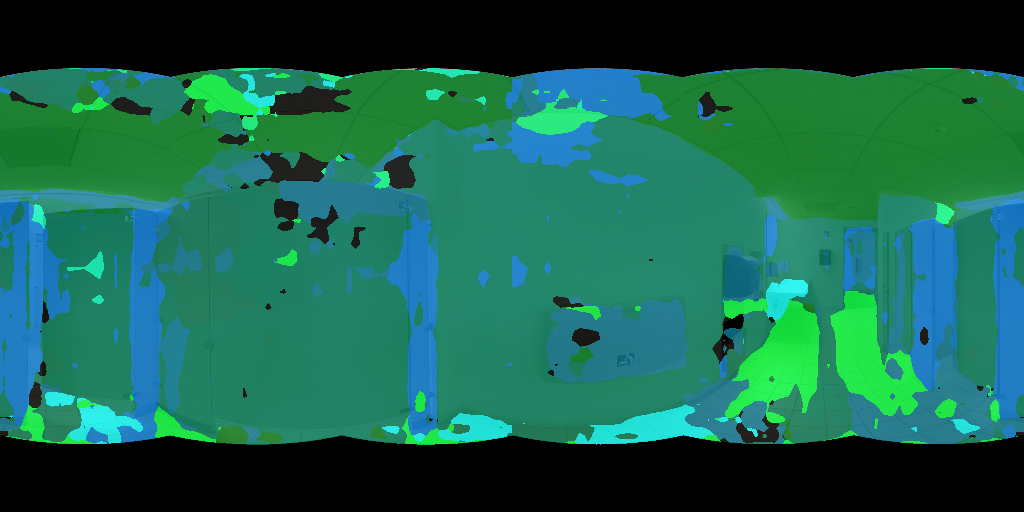} 
&
\includegraphics[width=.24\linewidth]{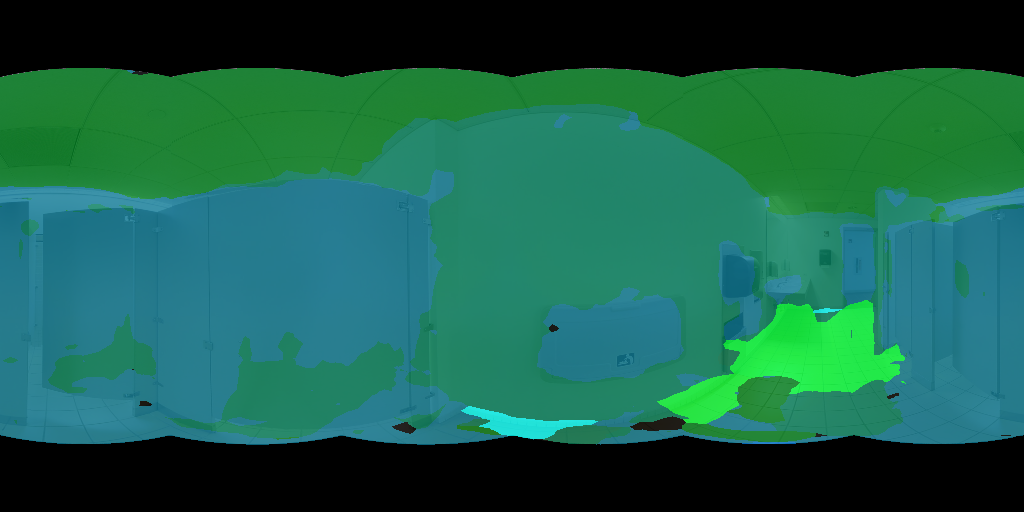} 
\\
\includegraphics[width=.24\linewidth]{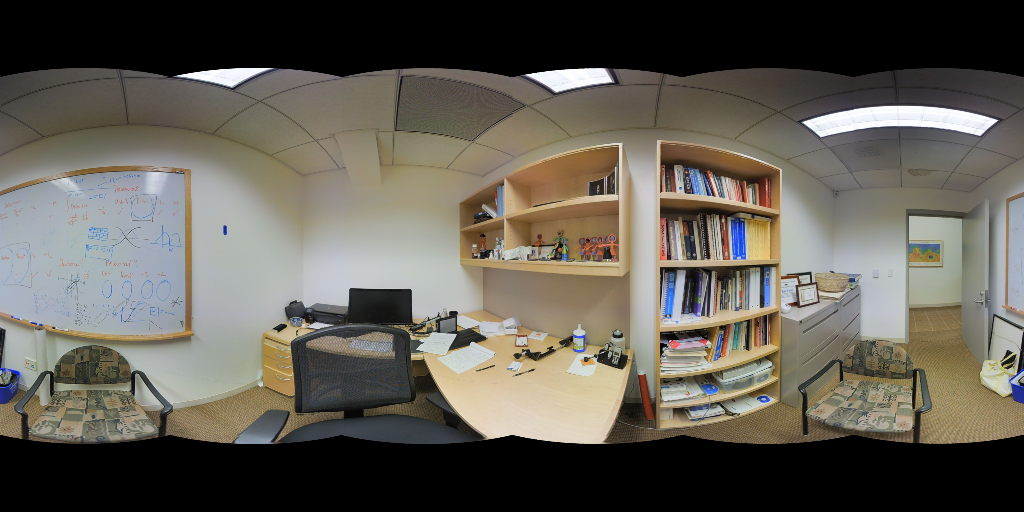} 
&
\includegraphics[width=.24\linewidth]{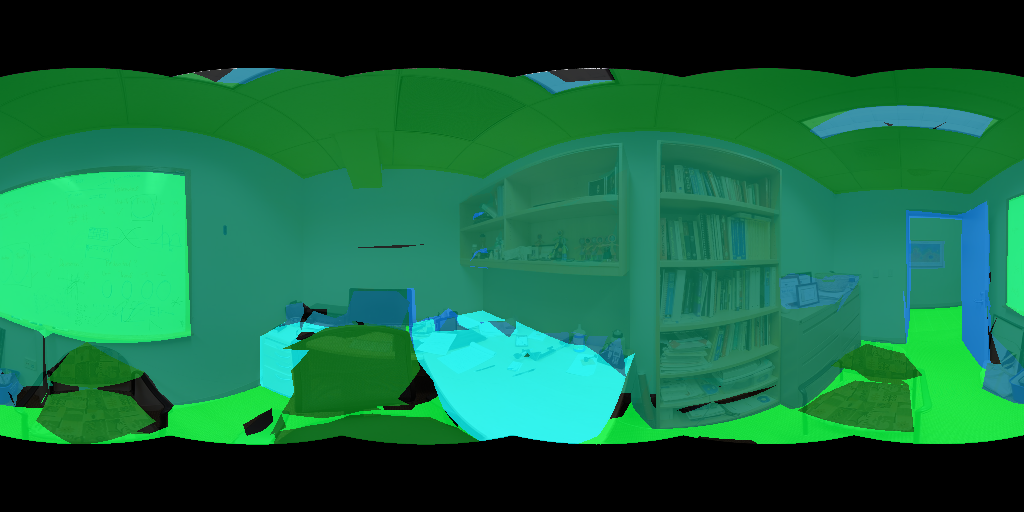} 
&
\includegraphics[width=.24\linewidth]{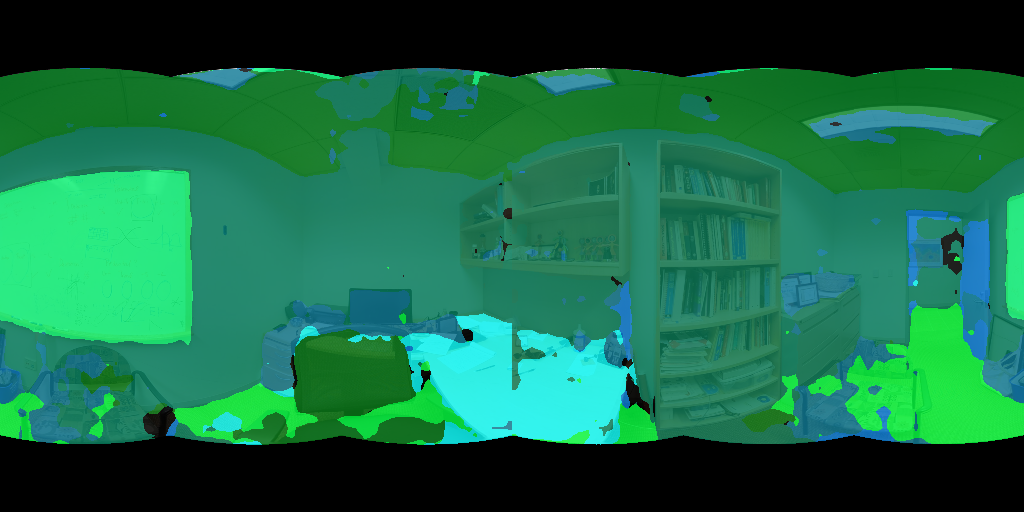} 
&
\includegraphics[width=.24\linewidth]{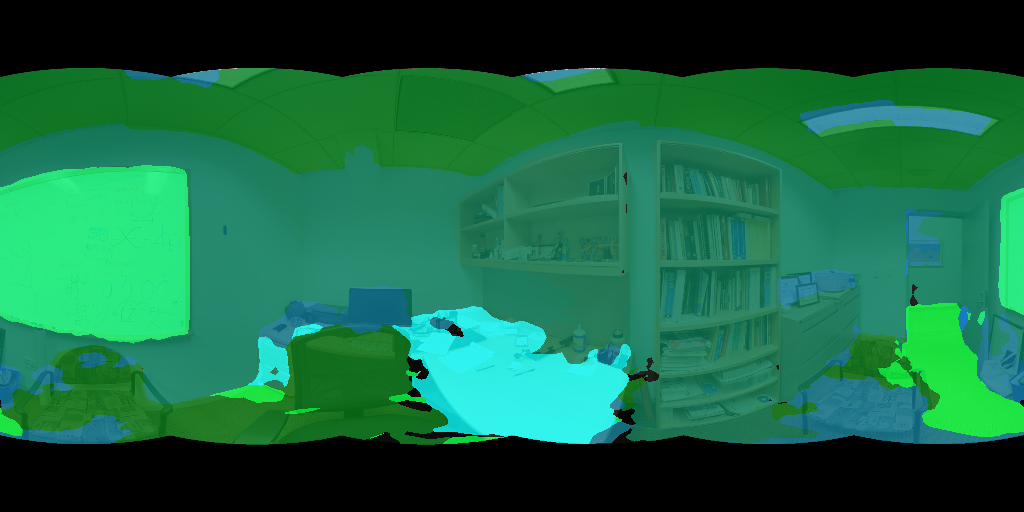} 
\\
(a) Orig. Image
&
(b) Ground Truth
&
(c) Naive 
&
(d) SelectionConv 
\end{tabular}
\end{center}
   \caption{A visual comparison of semantic segmentation of images from the Stanford 2D-3D-S~\cite{Armeni2017} dataset (a,b) between an FCN~\cite{Long_2015_CVPR} with a ResNet-50~\cite{He2016} backbone using standard convolutions (c) versus our SelectionConv operations (d).
   Note that the use of SelectionConv gives cleaner segmentation results along the poles of and seam of the image (located in the center of this representation).}
   
   

\label{fig:segmentation}
\end{figure}

\subsection{Spherical Segmentation}

We apply SelectionConv to the task of semantic segmentation on images in the Stanford 2D-3D-S~\cite{Armeni2017} dataset.
To do so, we first trained a standard 2D FCN~\cite{Long_2015_CVPR} using a ResNet-50~\cite{He2016} backbone on 2D projected views.
We then transfer the weights to a SelectionConv-based version of FCN~\cite{Long_2015_CVPR} and apply the network, with no additional training, to the standard test set,
converting the data to spherical graphs using the method described in Sec.~\ref{subsec:360}.
This gives an improvement over naively applying the network to the equirectangular images, as shown in Fig.~\ref{fig:segmentation}.
When operating on the validation set, the naive approach gives an average IOU score of 32.57\%.
We compare our results with that of another transfer-based method, distortion-aware convolutions~\cite{Tateno_2018_ECCV}, who reported an average IOU score of 34.56\% on the same dataset.
Our results show an average IOU score of 36.29\%.
Although this is only a small improvement and is still below state-of-the-art performance for RGB spherical segmentation
(45.6\% \cite{Eder2020}),
we again note that other methods are designed and fine-tuned specifically for spherical tasks, whereas we can achieve a performance boost through a simple design of our graph structure.


\subsection{Superpixel Depth Prediction} \label{sec:superdepth}

\begin{figure}[t]
\begin{center}
\hspace*{-0.1in}
\begin{tabular}{cccc}

\includegraphics[width=.23\linewidth]{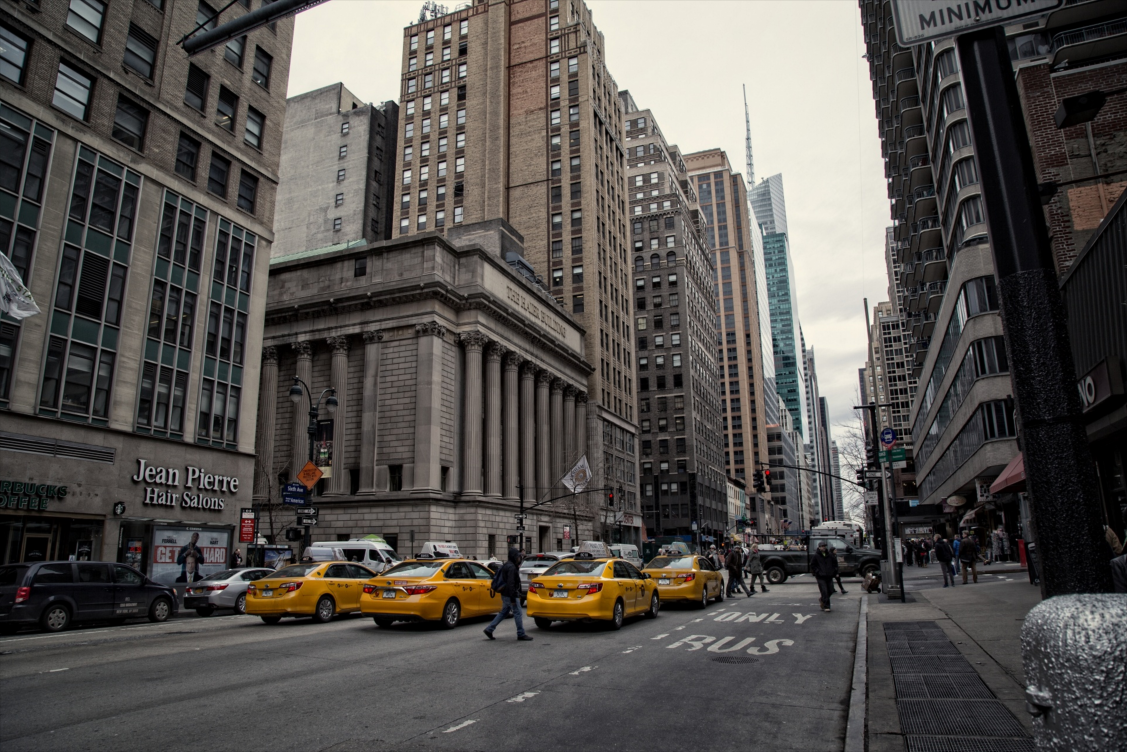} &
\includegraphics[width=.23\linewidth]{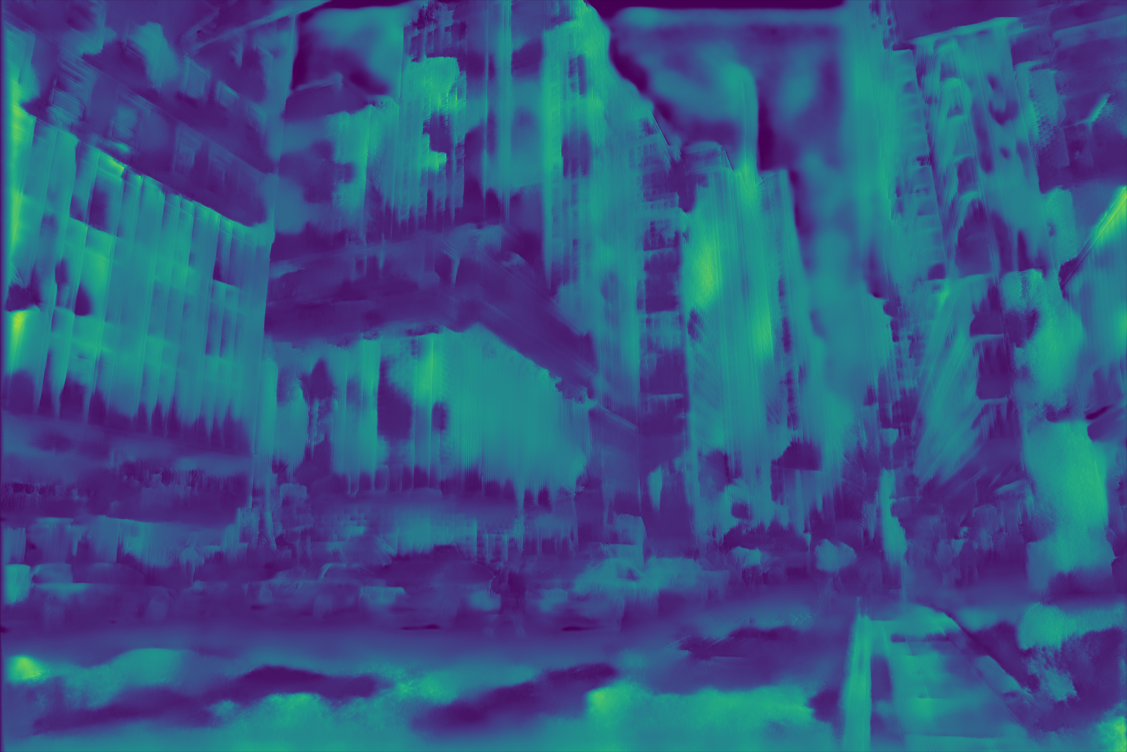} &
\includegraphics[width=.23\linewidth]{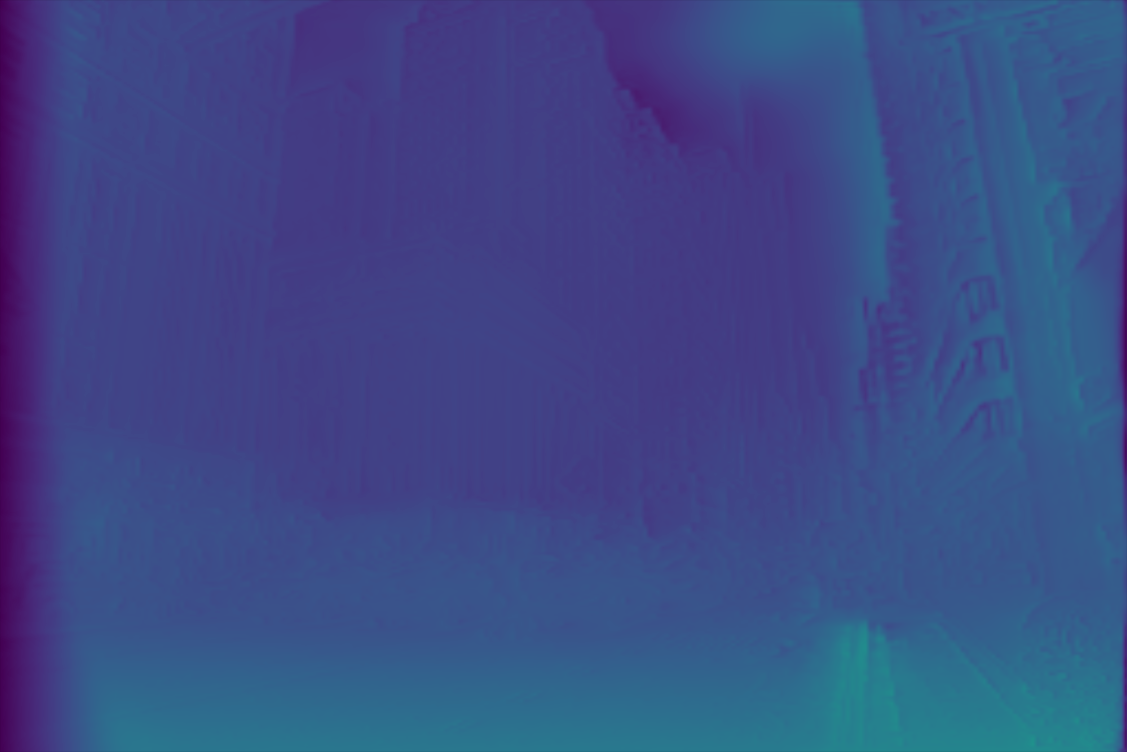} &
\includegraphics[width=.23\linewidth]{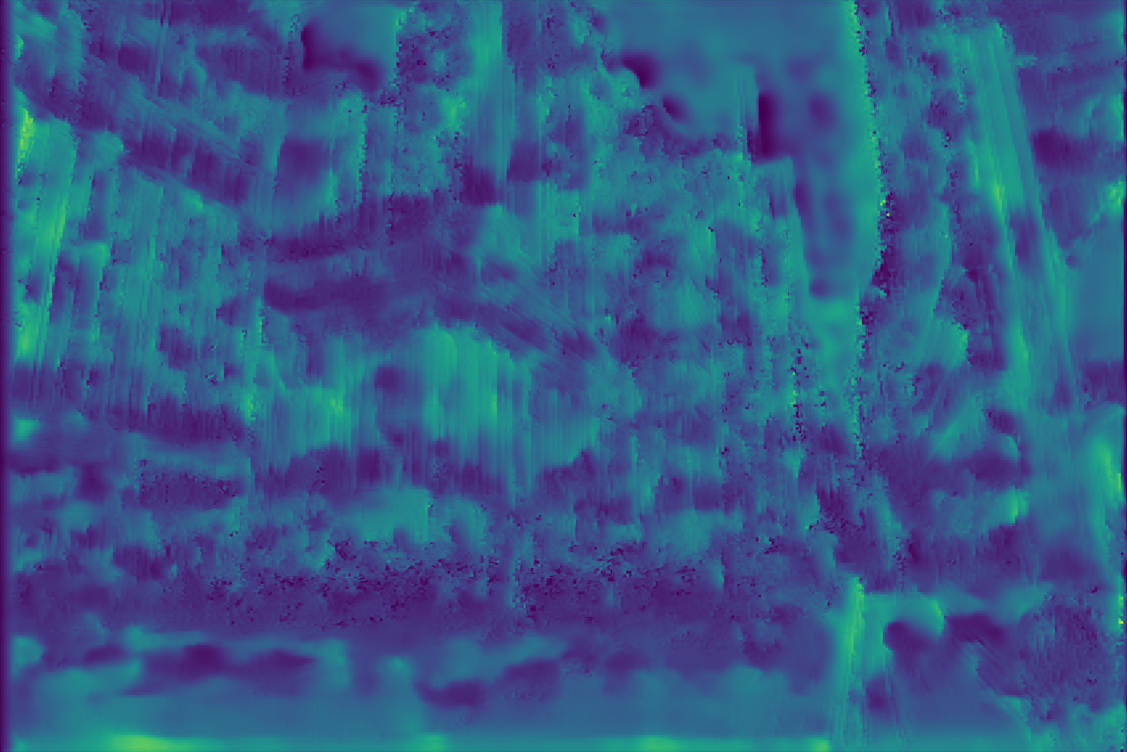} 

\\
(a) Original &
(b) Prediction &
(c) Low-Res &
(d) SelectionConv 

\end{tabular}
\end{center}
   \caption{ A high-resolution image (a) requires $\thicksim$25.9 seconds on a CPU to create a predicted depth map (b). A lower-resolution $256 \times 256$ version can be processed by a network in $\thicksim$0.8 seconds on a GPU, but with low-fidelity results when upscaled to the same resolution~(c). Generating approximately the same number of superpixels as the low-resolution image then using our graph-based network requires only $\thicksim$5.1 seconds on a GPU with higher-fidelity results (d).
   }
\label{fig:superpixel}
\end{figure}

We now illustrate possible applications using our method with superpixels for efficient processing of high-resolution images. We use a Pytorch implementation of a monocular depth estimator \cite{monodepth17,depthestimatorcode}. When operating on a 4K image, the amount of data is too large for a consumer-grade GPU, necessitating downsampling the input before processing. 
By comparison, if the image is first converted into a graph of neighboring SLIC superpixels~\cite{SLIC}, SelectionConv can process it on a GPU
and output results that are of much higher quality than using a downsampled image.
An example of such cases is shown in Fig.~\ref{fig:superpixel}. We again note that Yang \etal\cite{Yang2020Superpixel} complete a similar task with state-of-the-art performance, but their method requires using superpixels generated by their trained network. The SelectionConv network, in comparison, can utilize any superpixel method.

\subsection{Masked Image and 3D Mesh Style Transfer} \label{sec:TextureStyle}

Lastly, 
we demonstrate the ability of our network to work on data with many discontinuities by performing style transfer on masked images and texture maps.

\begin{figure}[t]
\begin{center}
\begin{tabular}{cccc}

\includegraphics[width=.21\linewidth]{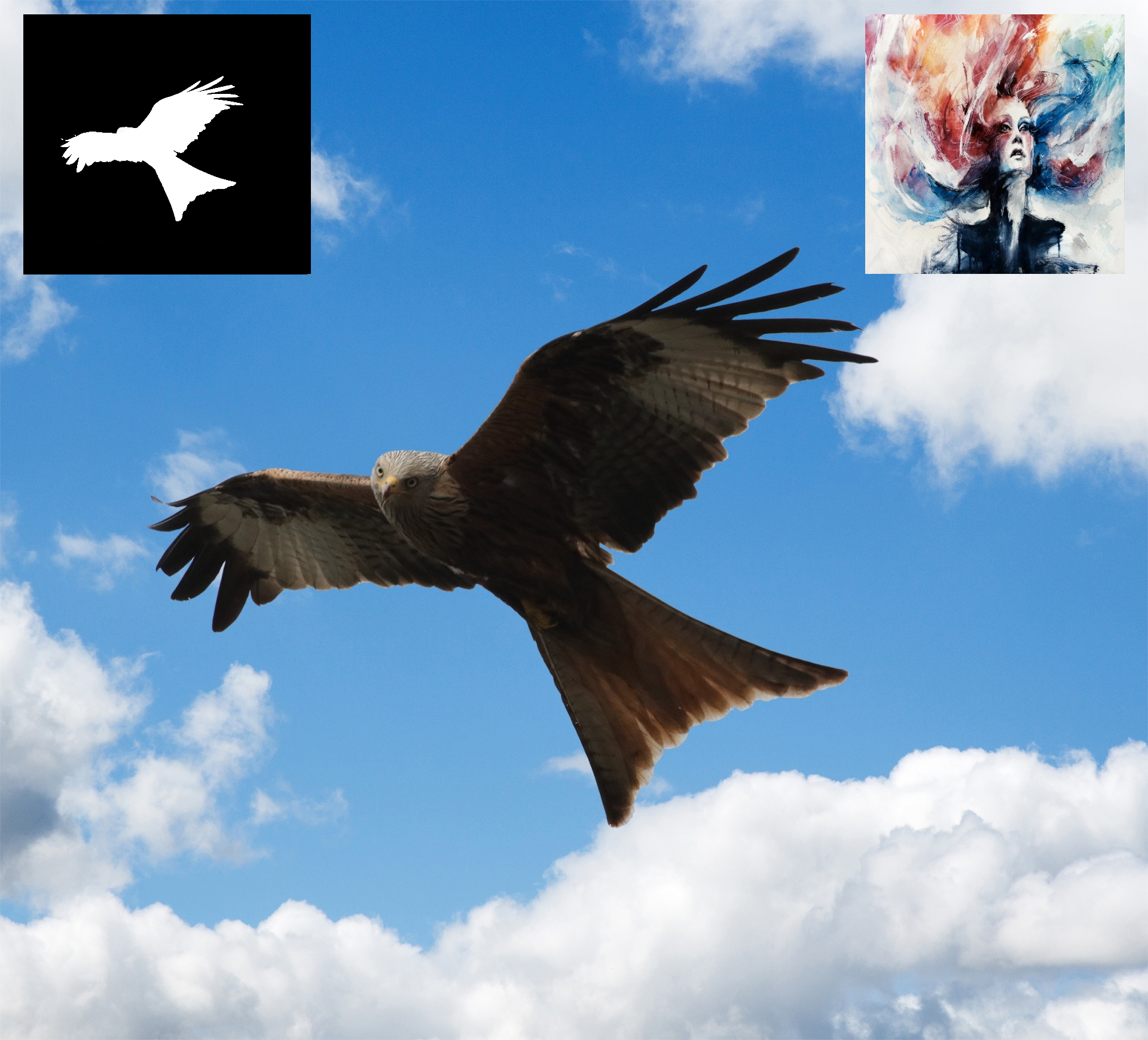} &
\includegraphics[width=.21\linewidth]{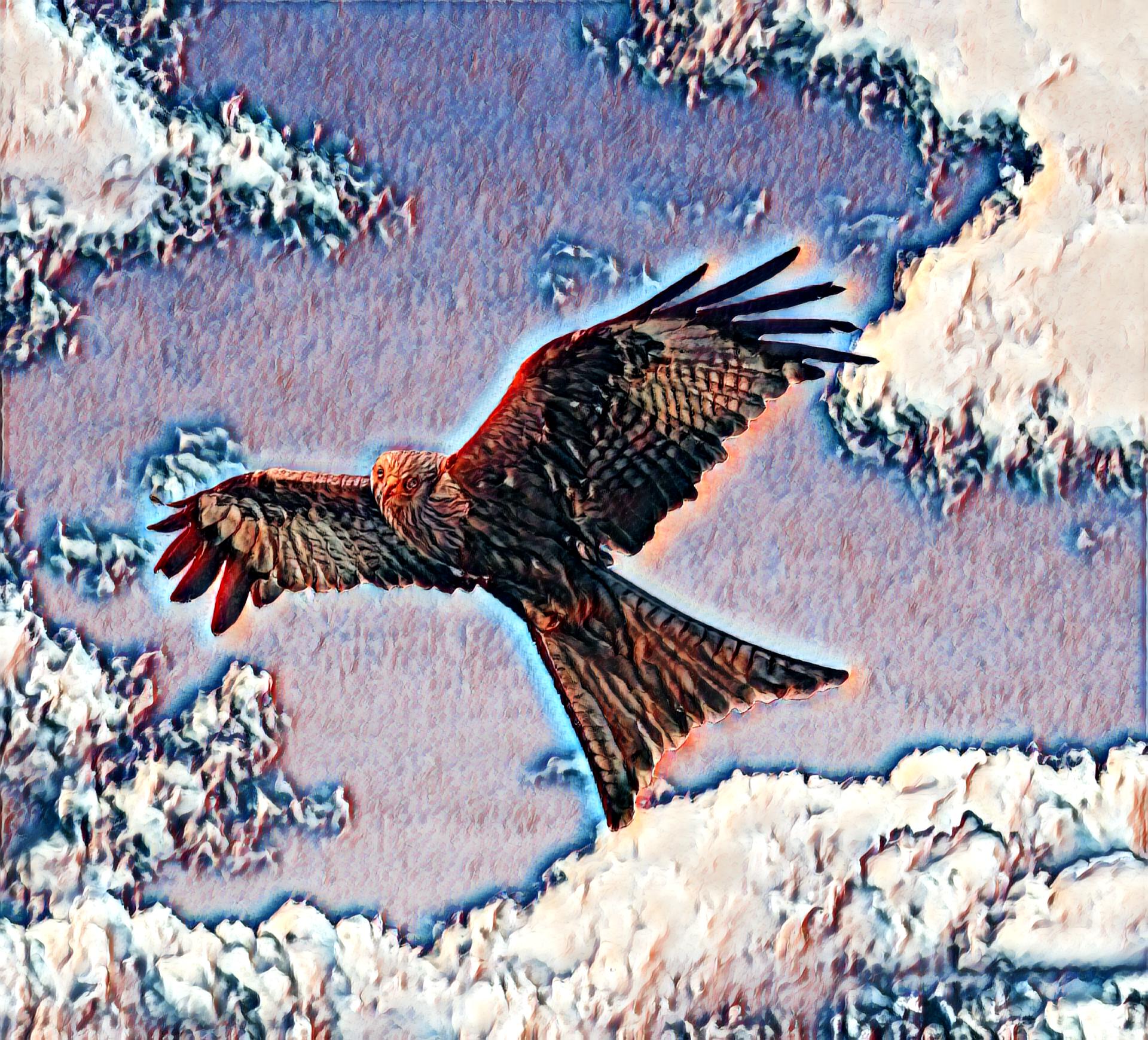} &
\includegraphics[width=.21\linewidth]{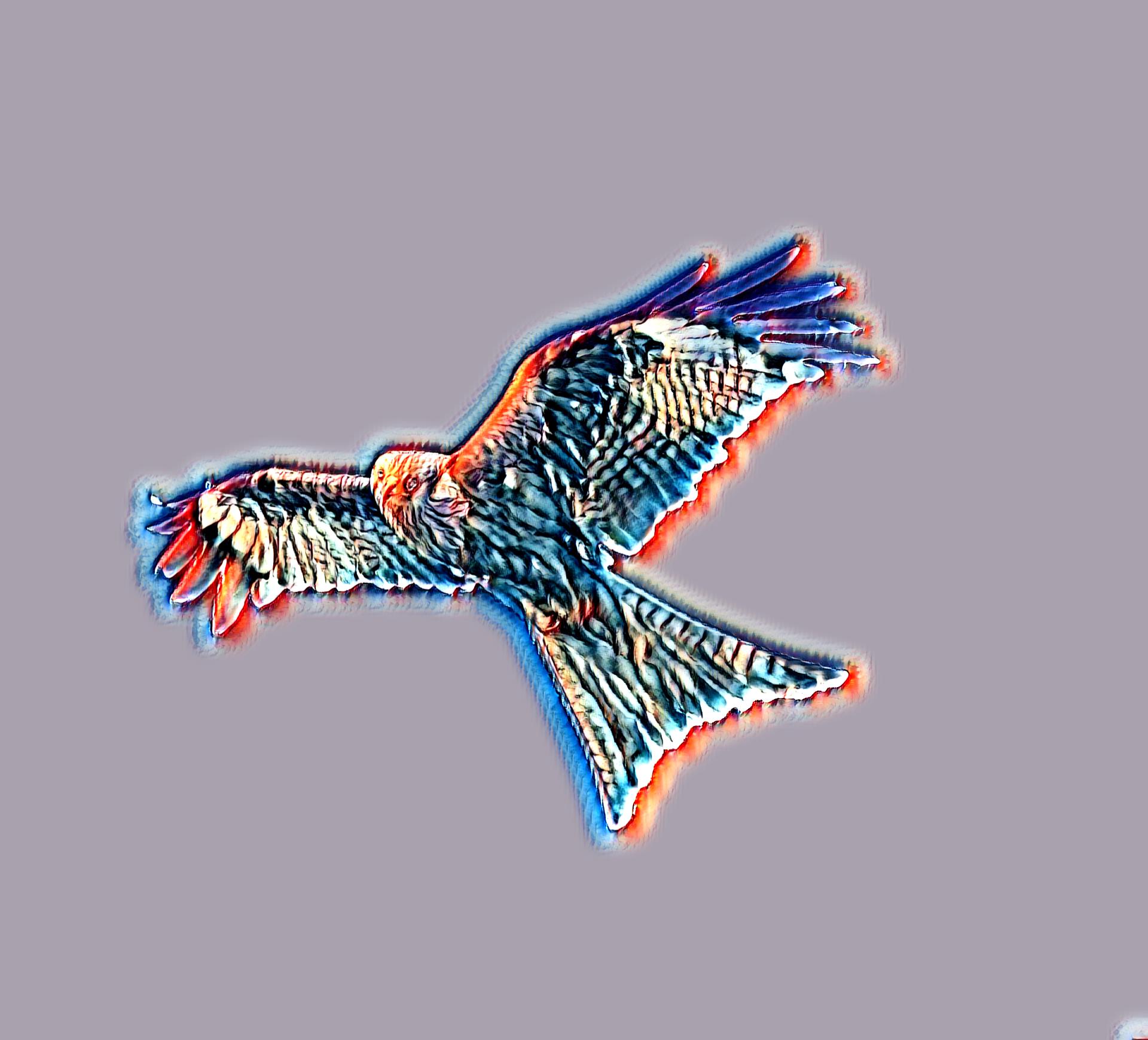} &
\includegraphics[width=.21\linewidth]{Figures/masked_output_style6.jpg}

\\
(a) Original &
(b) Post-masking &
(c) Pre-masking &
(d) Ours

\end{tabular}
\end{center}
   \caption{A content image, a masked region of interest, and a given style image~(a). 
   To stylize the masked region with a traditional CNN, the entire image can be stylized~(b) or the image can be masked before stylization~(c) and then the masked result can be applied back to the original. In both cases, outside statistics influence the stylization inside the region of interest (making (b) darker than expected and (c) brighter than expected).
   In comparison, our method (d) can generate a graph just for the masked region, which more closely matches the style image statistics in the region of interest.} 
\label{fig:maskedregion}
\end{figure}

To achieve style transfer on a masked region with a regular CNN would require stylizing the entire image 
or a zero-padded masked image 
and then reapplying the unmasked region.
This means that background features can affect stylization of the foreground. 
In comparison, our method can handle these scenarios natively, leading to a stylization that only depends on the foreground statistics. 
Comparisons of these approaches are shown in Fig.~\ref{fig:maskedregion}.

\begin{figure}[t]
\begin{center}
\begin{tabular}{ccccc}


\includegraphics[width=.12\linewidth]{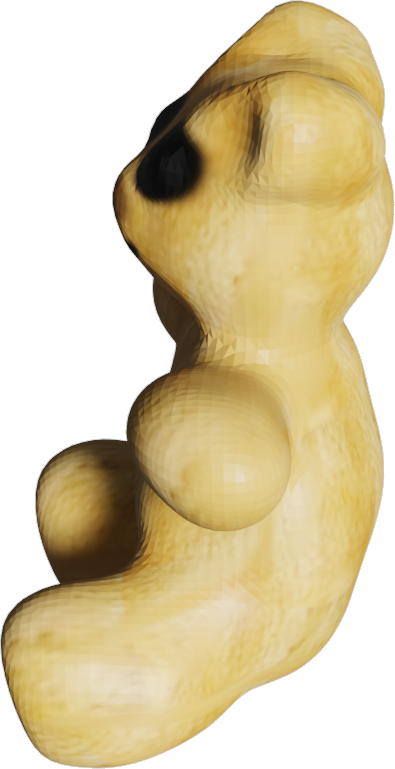} &
\hspace*{0.1in}
\includegraphics[width=.12\linewidth]{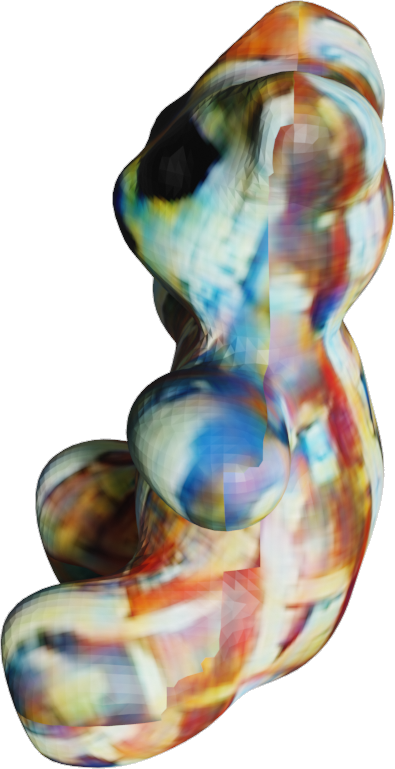} &
\hspace*{0.1in}
\cropim{.15}{Figures/teddy2naive.png}{170}{400}{25}{50} &
\hspace*{0.1in}
\includegraphics[width=.12\linewidth]{Figures/teddy2ours.png} & 
\hspace*{0.1in}
\cropim{.15}{Figures/teddy2ours.png}{170}{400}{25}{50}
\\


(a) 3D Mesh &
(b) Naive &
(c) Close-up&
(d) SelectionConv &
(e) Close-up

\end{tabular}
\end{center}
   \caption{ 
        3D mesh~(a), the result of naively stylizing the texture map~(b) and a magnification (c), and the result of using our method~(d) and a magnification (e). Note the visible seams shown in the magnifications of the naive method (c), whereas our method in~(e) minimizes the visibility of those seams.
   }
\label{fig:meshstyles}
\end{figure}

As another illustration, treating a texture map as a 2D image and naively performing style transfer leads to noticeable seams in the mapped texture. 
In comparison, using the graph structure proposed in Sec.~\ref{subsec:Texture} leads to more continuous patterns. 
Visualizations of these two methods for various stylizations and meshes are shown in Fig.\ref{fig:meshstyles}. 

Others have attempted style transfer between two different 3D objects~\cite{StyLit,StyBlit,yin2021}, but we are not aware of other work attempting direct style transfer between the texture map of the 3D mesh and a 2D image. 


\section{Conclusion}

We have presented a method that allows for information from pre-trained traditional convolutional neural networks to be transferred directly to a new kind of graph convolutional network. 
This makes it possible for these previously trained networks to operate on data that they could not before, such as superpixels, spherical images, and texture maps. 
We have demonstrated various use cases and given the general framework so that others can continue to extend this method for their needs.
In theory, any set of adjacency matrices could be designed to work with the particular data of a graph. 
Future research could also use selection-based convolution to improve applications outside of the image domain.

%
%
\bibliographystyle{splncs04}
\bibliography{selectionconv}
\end{document}


\pagestyle{headings}
\mainmatter
\def\ECCVSubNumber{5224}  

\title{SelectionConv: Convolutional Neural Networks for Non-rectilinear Image Data \\ Supplemental Material}

\titlerunning{SelectionConv: CNNs for Non-rectilinear Image Data}
%
\author{David Hart \and
Michael Whitney \and
Bryan Morse}

%
\authorrunning{Hart et al.}
%
\institute{Brigham Young University, Provo, Utah, USA
\email{\{davidmhart,mikeswhitney,morse\}@byu.edu}\\}

\maketitle

\section{Code}

We provide our code and the network weights used for our various experiments. These are found at \url{http://github.com/davidmhart/SelectionConv} along with instructions and more details.

\section{Masked Region Stylization}

\begin{figure}
\newcommand{\figwidth}{0.40\linewidth}
\begin{center}
\begin{tabular}{cc}
\includegraphics[width=\figwidth]{Figures/FlyingBird_withStyle.jpg} &
\hspace*{0.2in}
\includegraphics[width=\figwidth]{Figures/reference_style6.jpg} 
\\
a) Original with Mask and Style &
\hspace*{0.2in}
b) Entire stylized image 
\\
\\
\includegraphics[width=\figwidth]{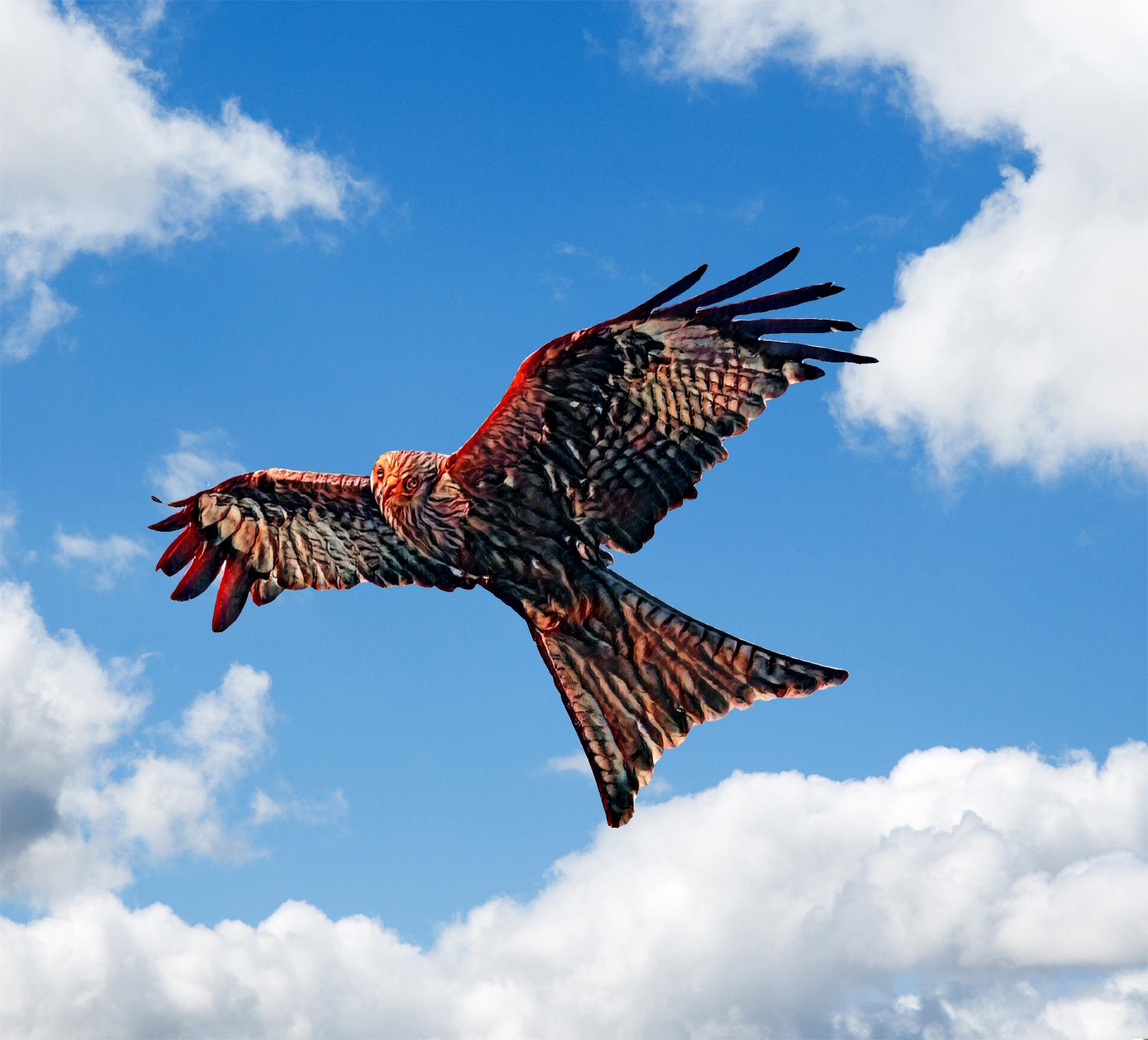} &
\hspace*{0.2in}
\includegraphics[width=\figwidth]{Figures/mreference_style6.jpg} 
\\
c) Composition of (a) and (b) &
\hspace*{0.2in}
d) Masked then stylized 
\\
\\

\includegraphics[width=\figwidth]{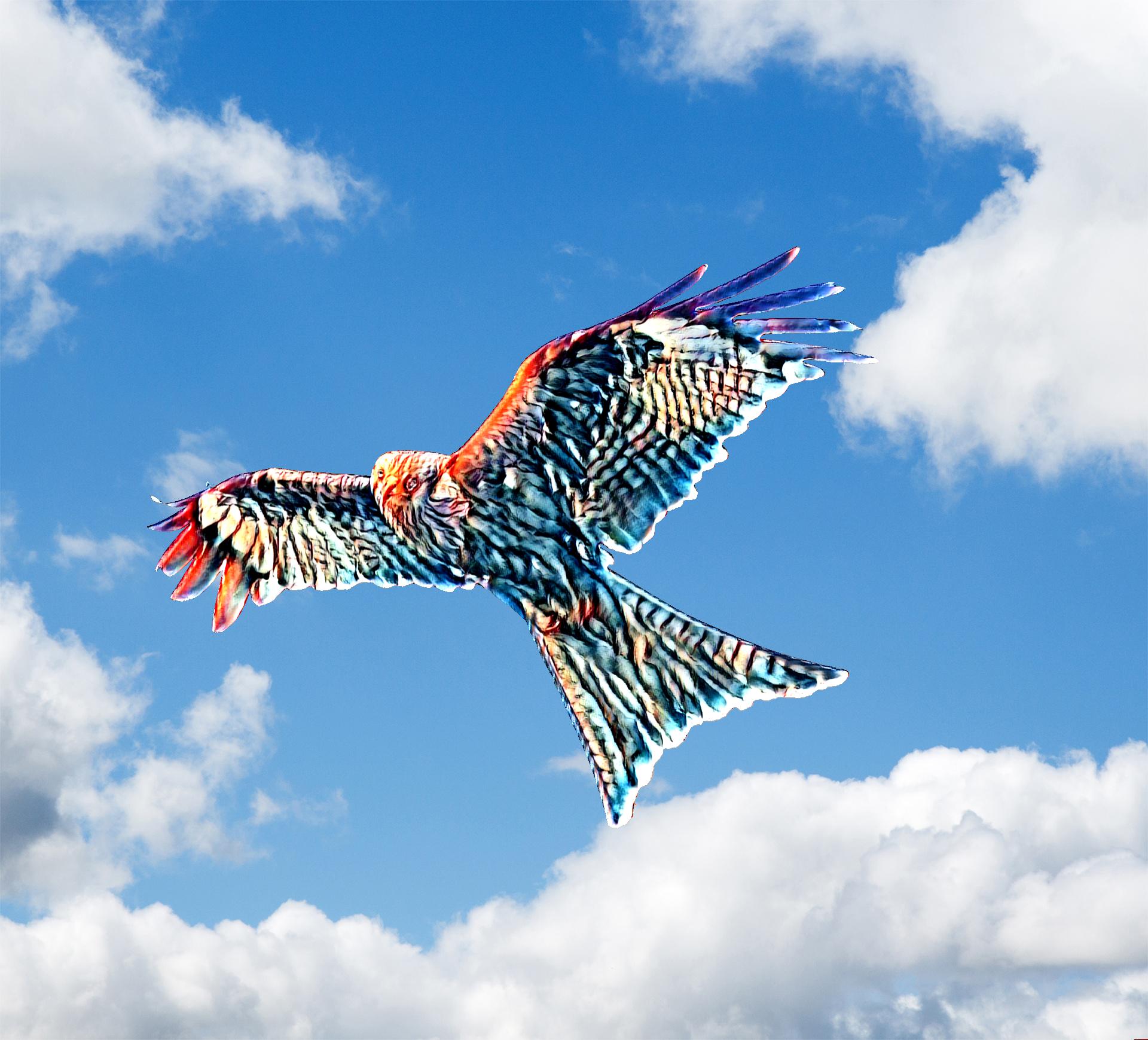} &
\hspace*{0.2in}
\includegraphics[width=\figwidth]{Figures/masked_output_style6.jpg}
\\

e) Composition of (a) and (d) &
\hspace*{0.2in}
f) Masked stylization
\end{tabular}
\end{center}
   \caption{
   A content image, a masked region of interest, and a given style image~(a). 
   To stylize the masked region with a traditional CNN, the entire image can be stylized~(b) or the image can be masked before stylization~(d) and then the masked result can be applied back to the original(c,e). In both cases, outside statistics influence the stylization inside the region of interest (making (c) darker than expected and (e) brighter than expected).
   In comparison, our method (f) can generate a graph just for the masked region, which more closely matches the style image statistics in the region of interest.
   }
\label{fig:masked1}
\end{figure}

\noindent
We provide an expanded version of Figure 9 from the paper here for easier viewing~(Fig.~\ref{fig:masked1}). 

~

\begin{figure}
\newcommand{\figwidth}{0.30\linewidth}
\begin{center}
\begin{tabular}{ccc}

\includegraphics[width=0.27\linewidth]{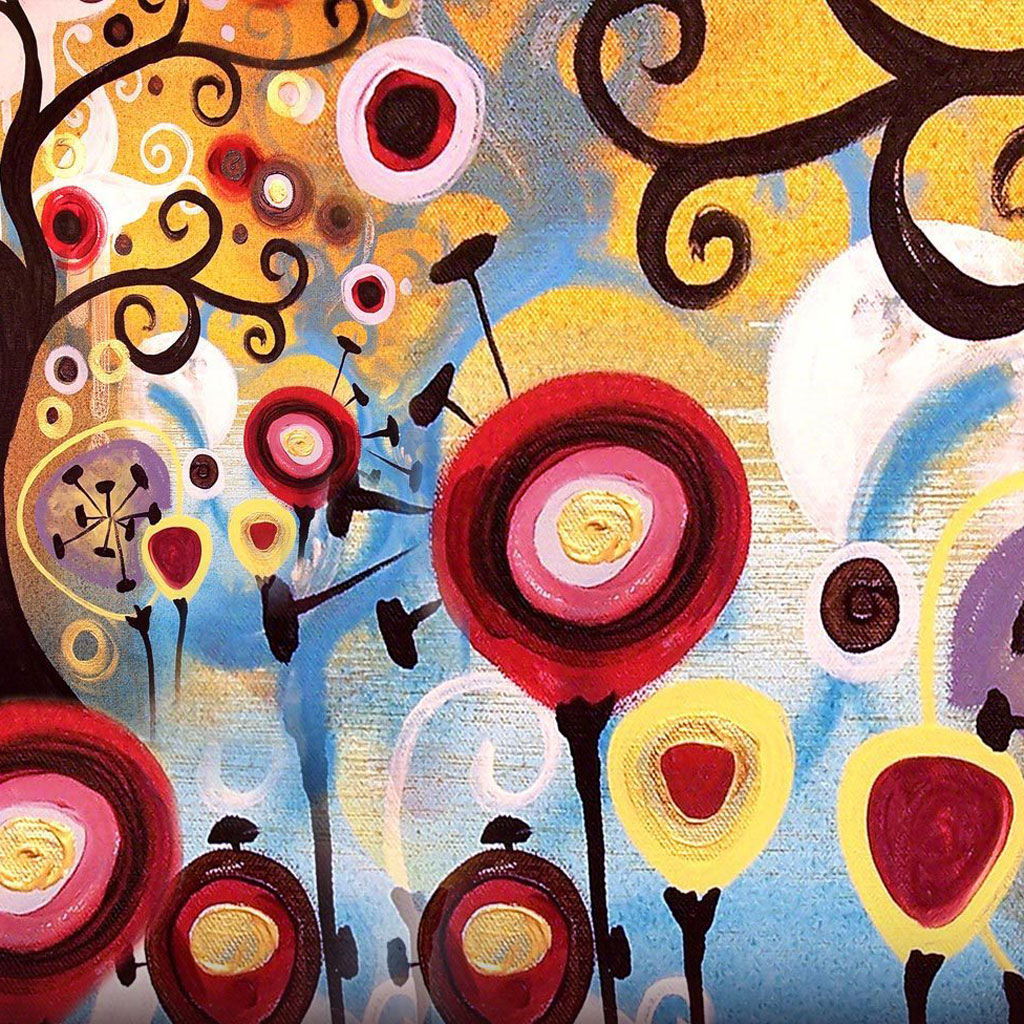} &
\includegraphics[width=\figwidth]{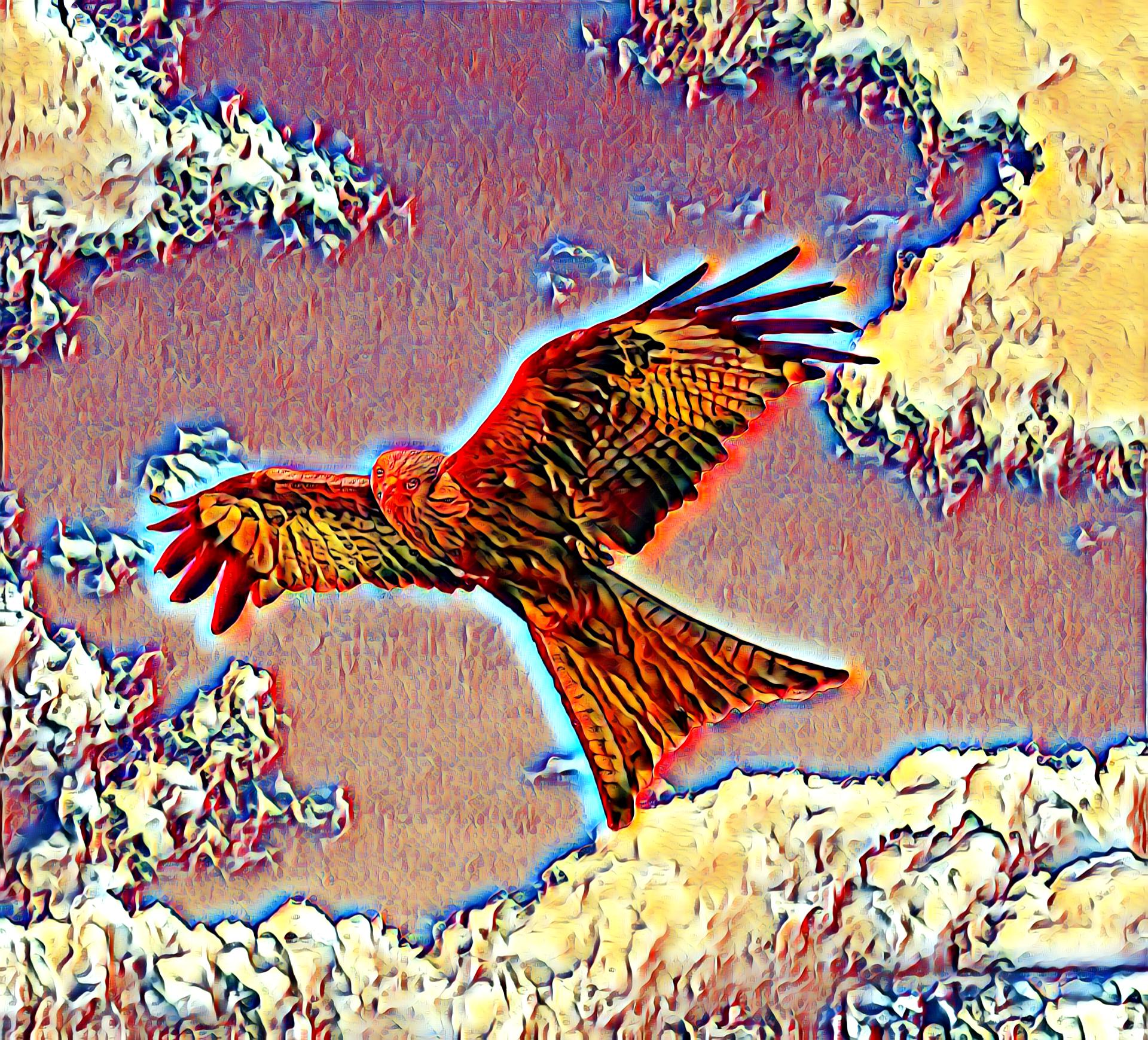} &
\includegraphics[width=\figwidth]{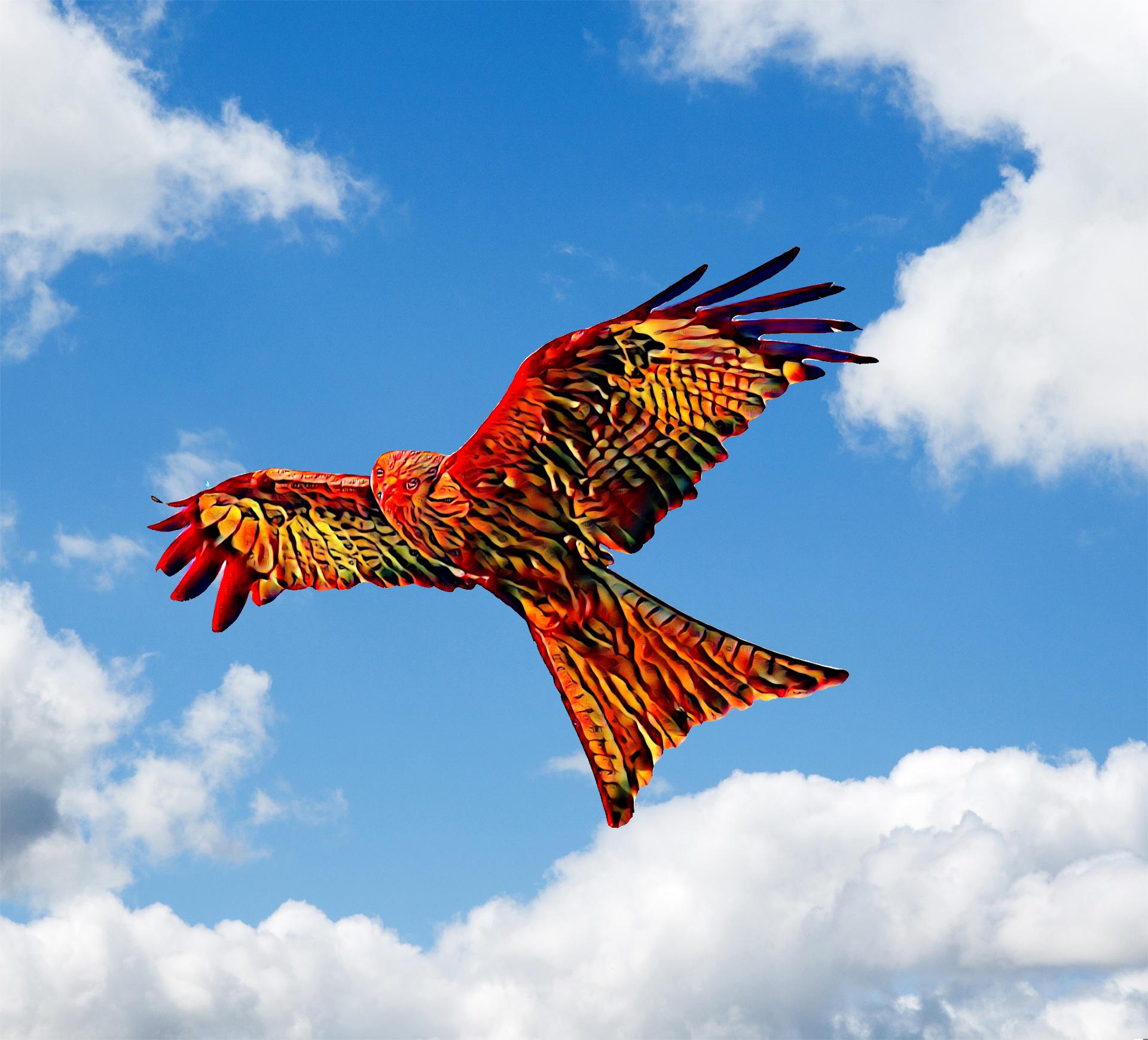} 
\\
a) Style Image &
b) Entire stylized image &
c) Composition of (b)
\\
\\
\includegraphics[width=\figwidth]{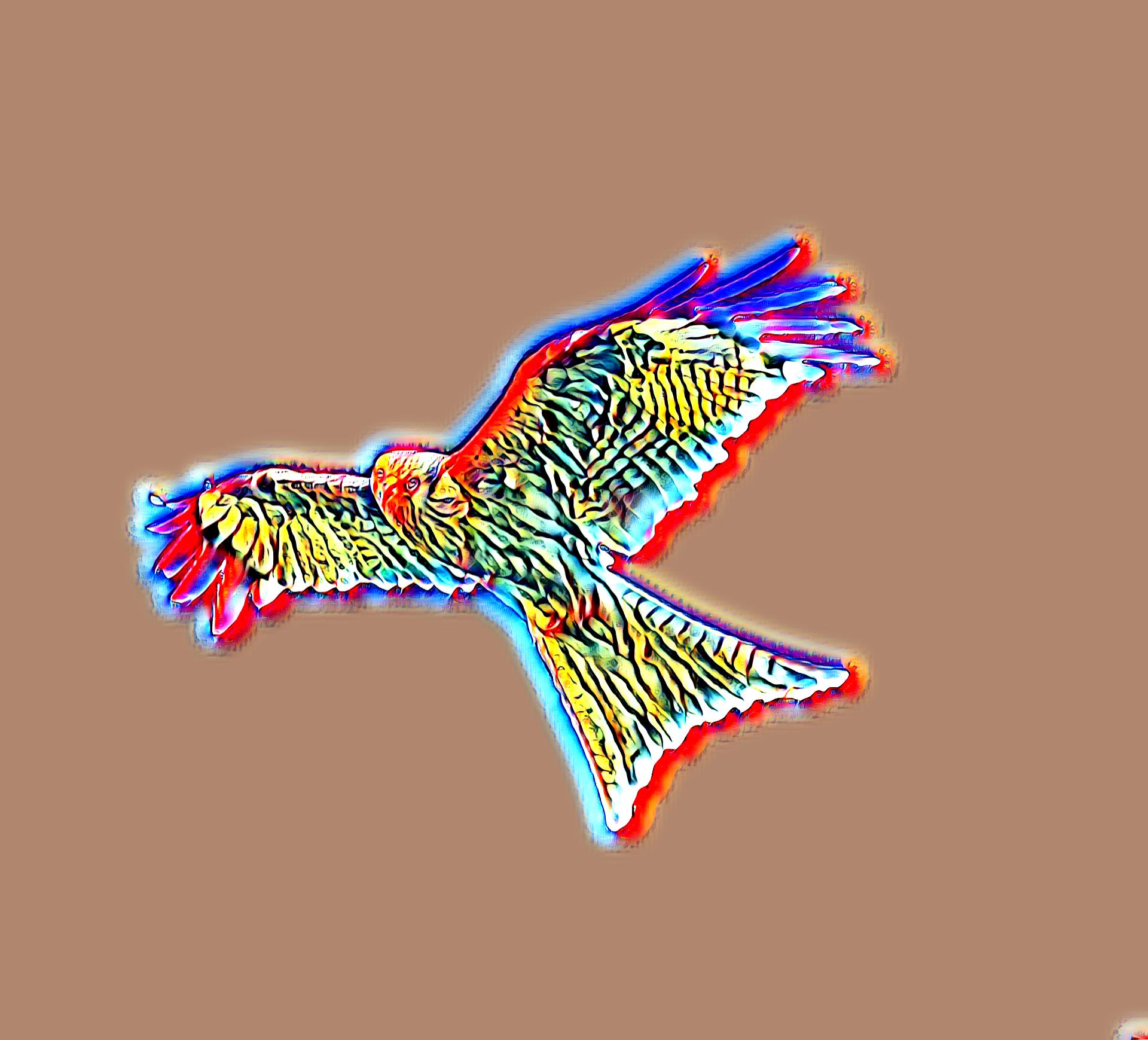} &
\includegraphics[width=\figwidth]{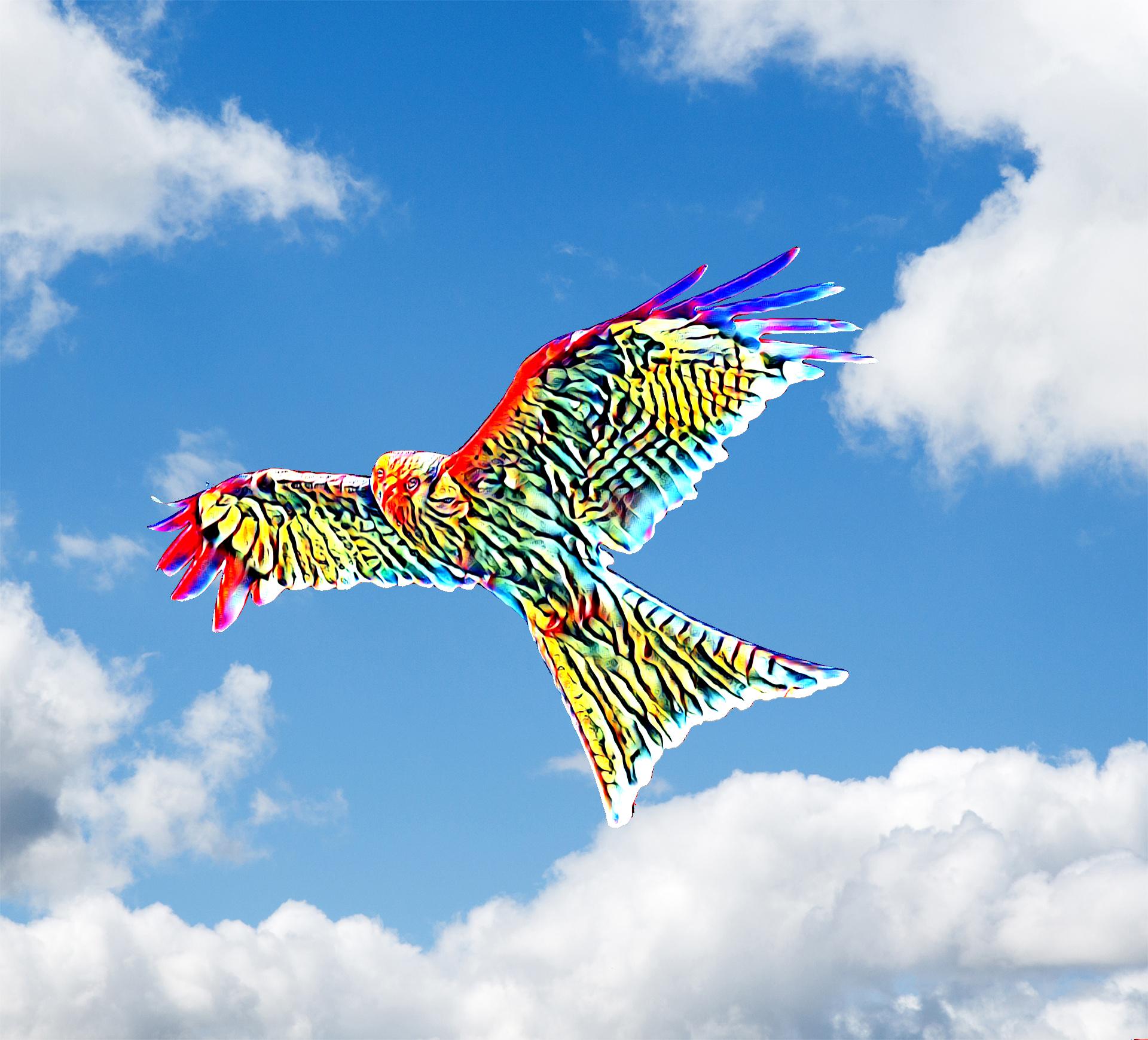} &
\includegraphics[width=\figwidth]{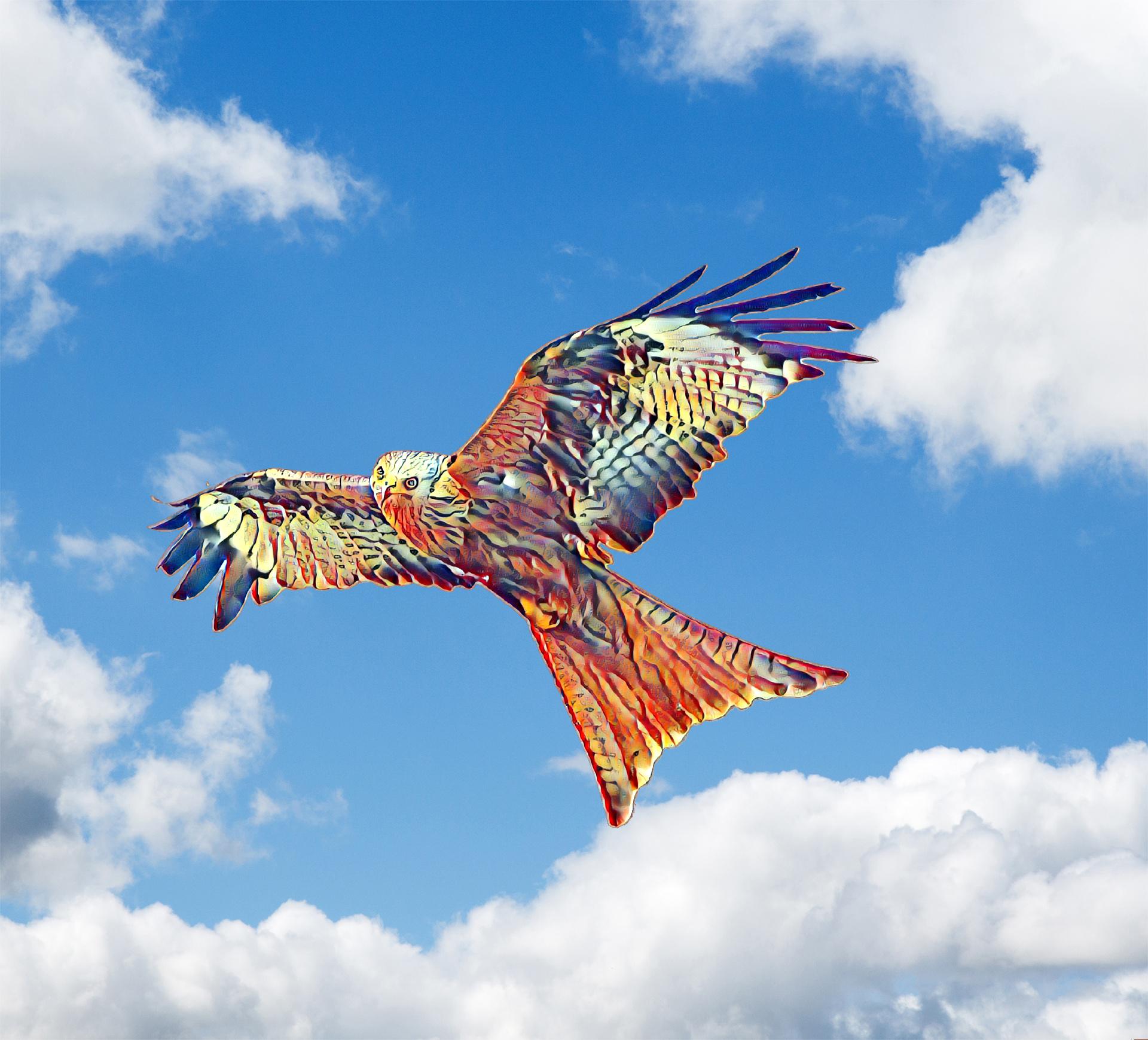}
\\
d) Masked then stylized &
e) Composition of (d) &
f) Masked stylization
\end{tabular}
\end{center}
   \caption{Another example, similar to Fig.~\ref{fig:masked1}, using a different style image.
   }
\label{fig:masked2a}
\end{figure}

\begin{figure}
\newcommand{\figwidth}{0.30\linewidth}
\begin{center}
\begin{tabular}{ccc}

\includegraphics[width=0.27\linewidth]{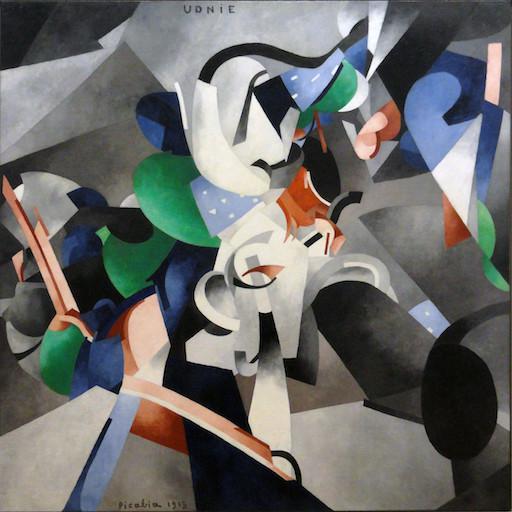} &
\includegraphics[width=\figwidth]{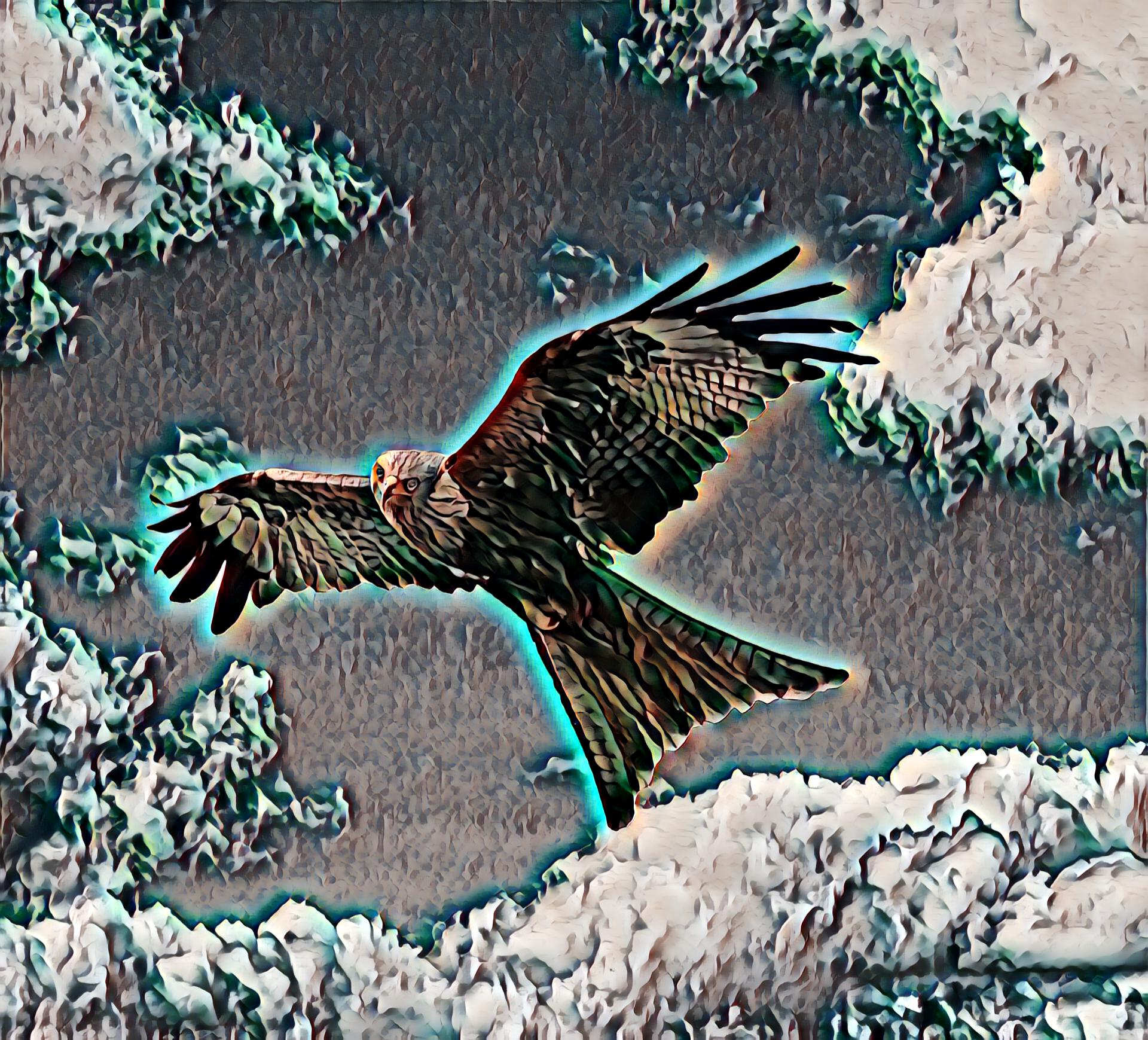} &
\includegraphics[width=\figwidth]{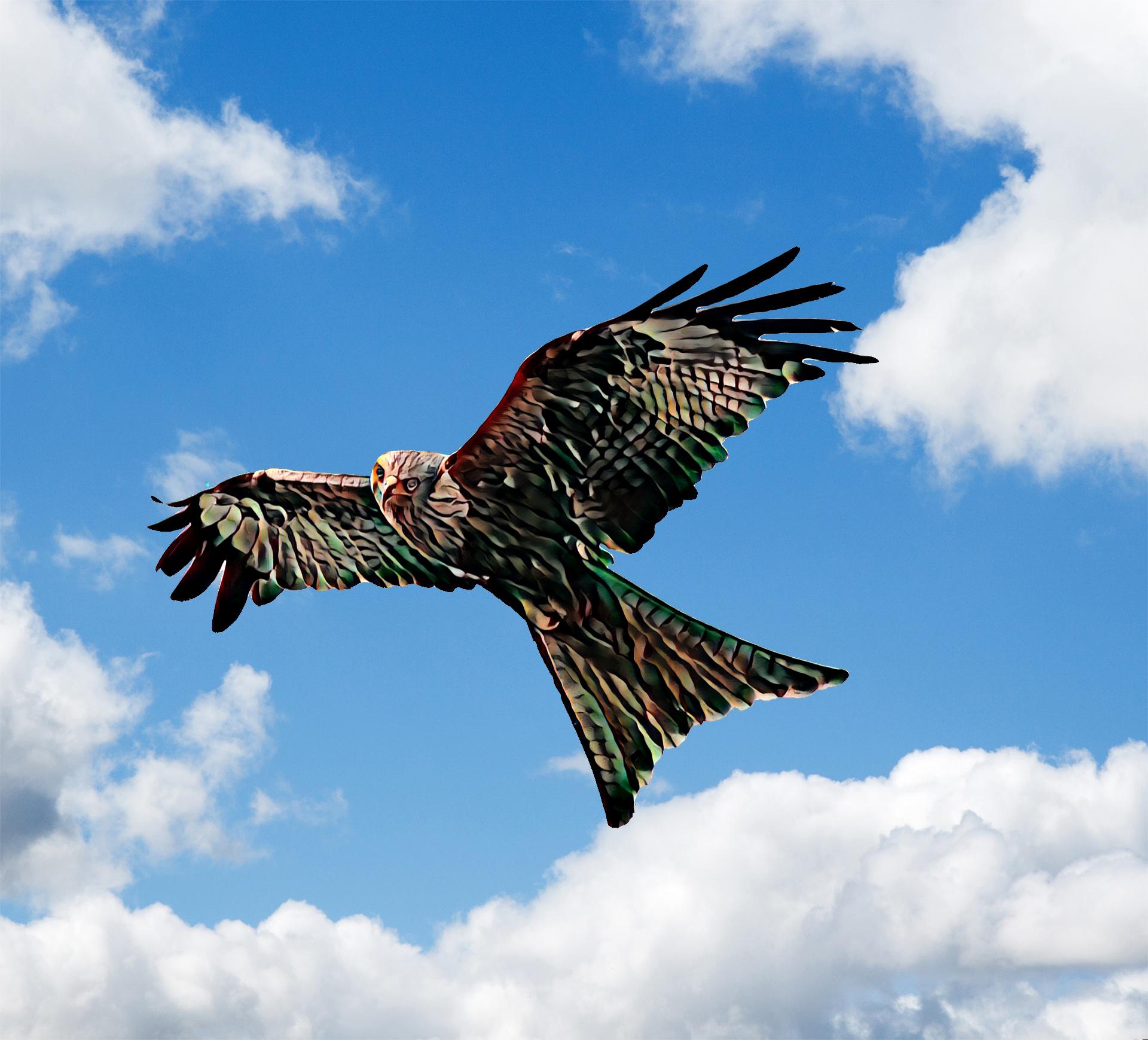}
\\
a) Style Image &
b) Entire stylized image &
c) Composition of (b)
\\
\\
\includegraphics[width=\figwidth]{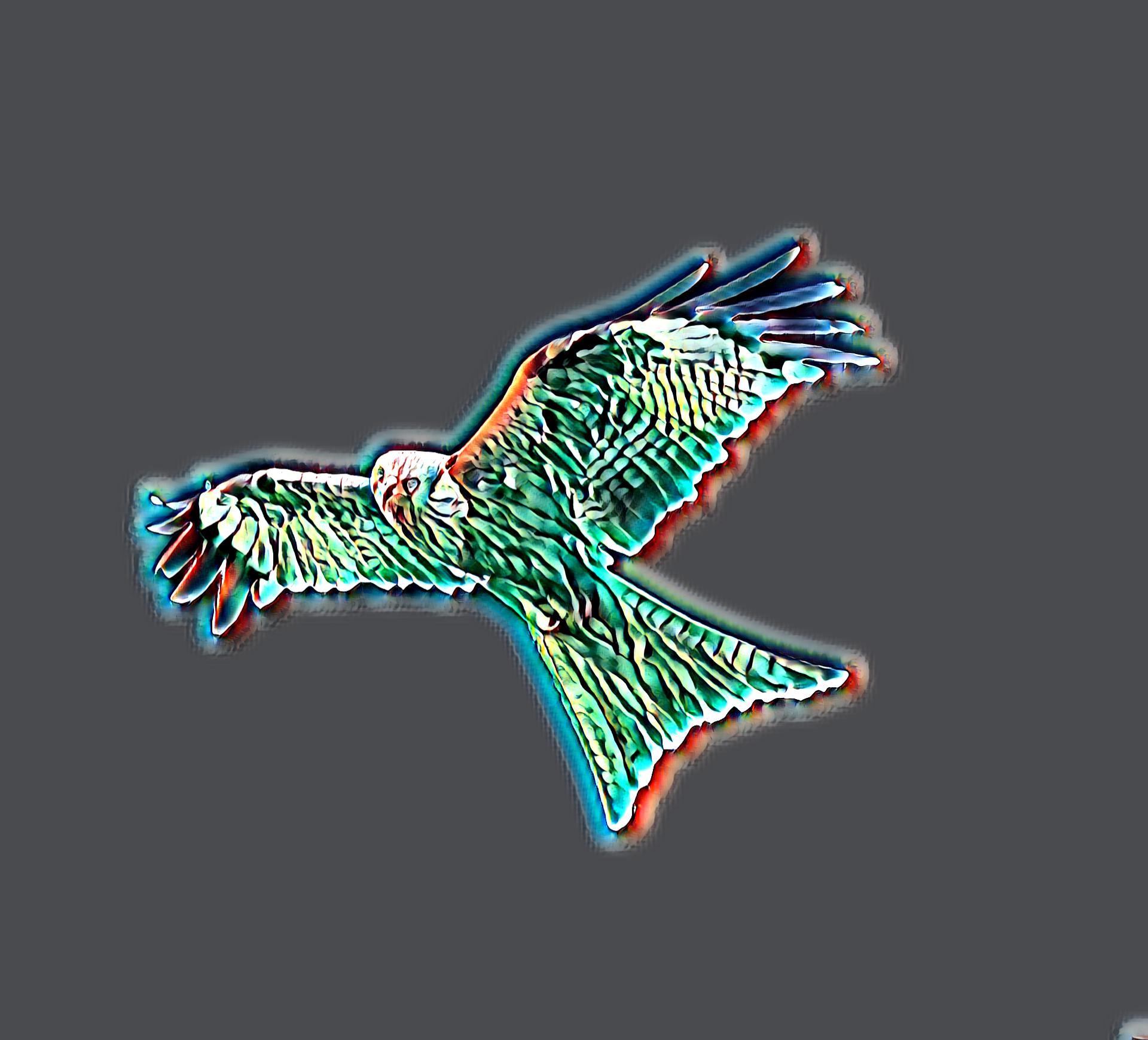} &
\includegraphics[width=\figwidth]{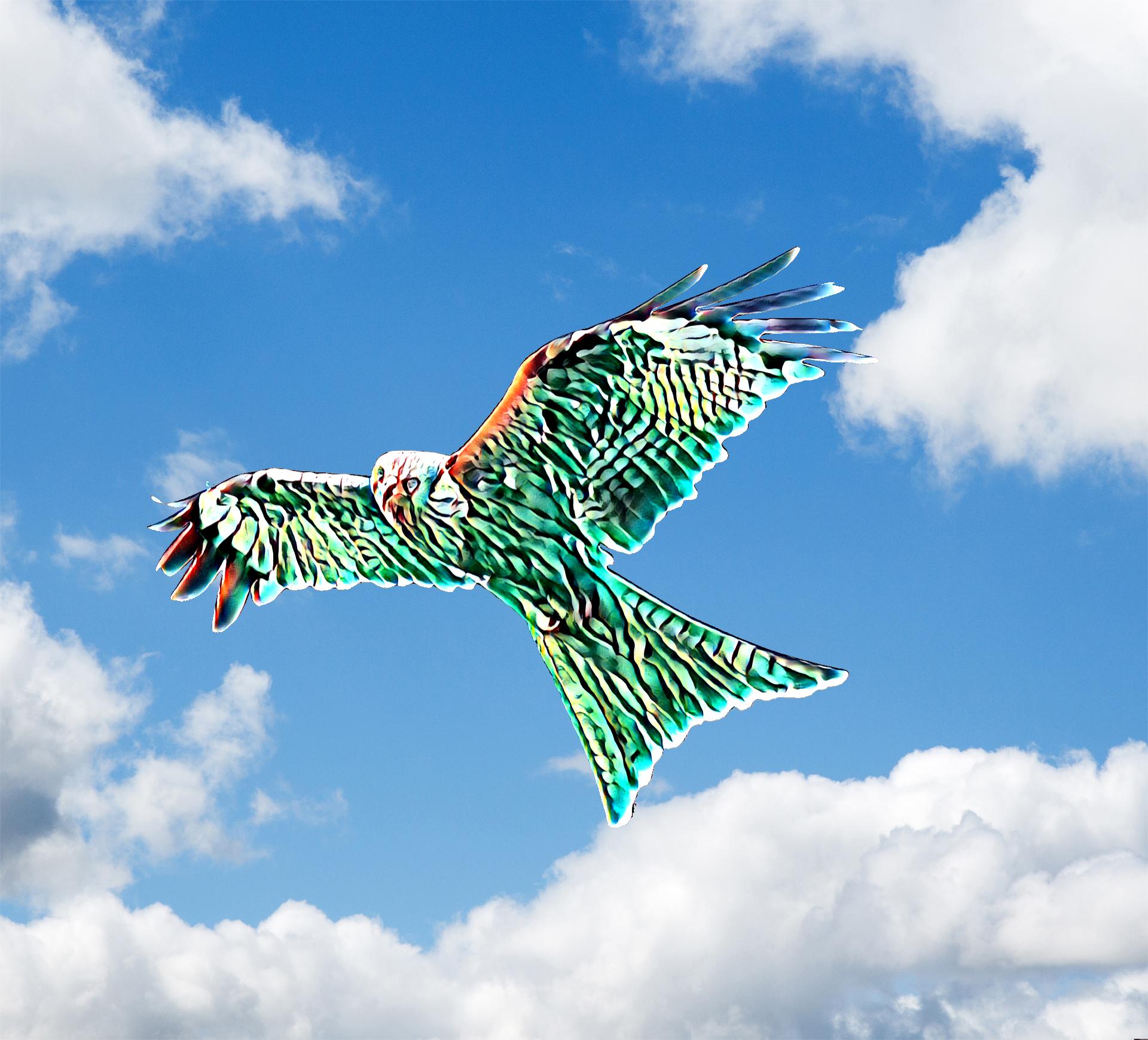} &
\includegraphics[width=\figwidth]{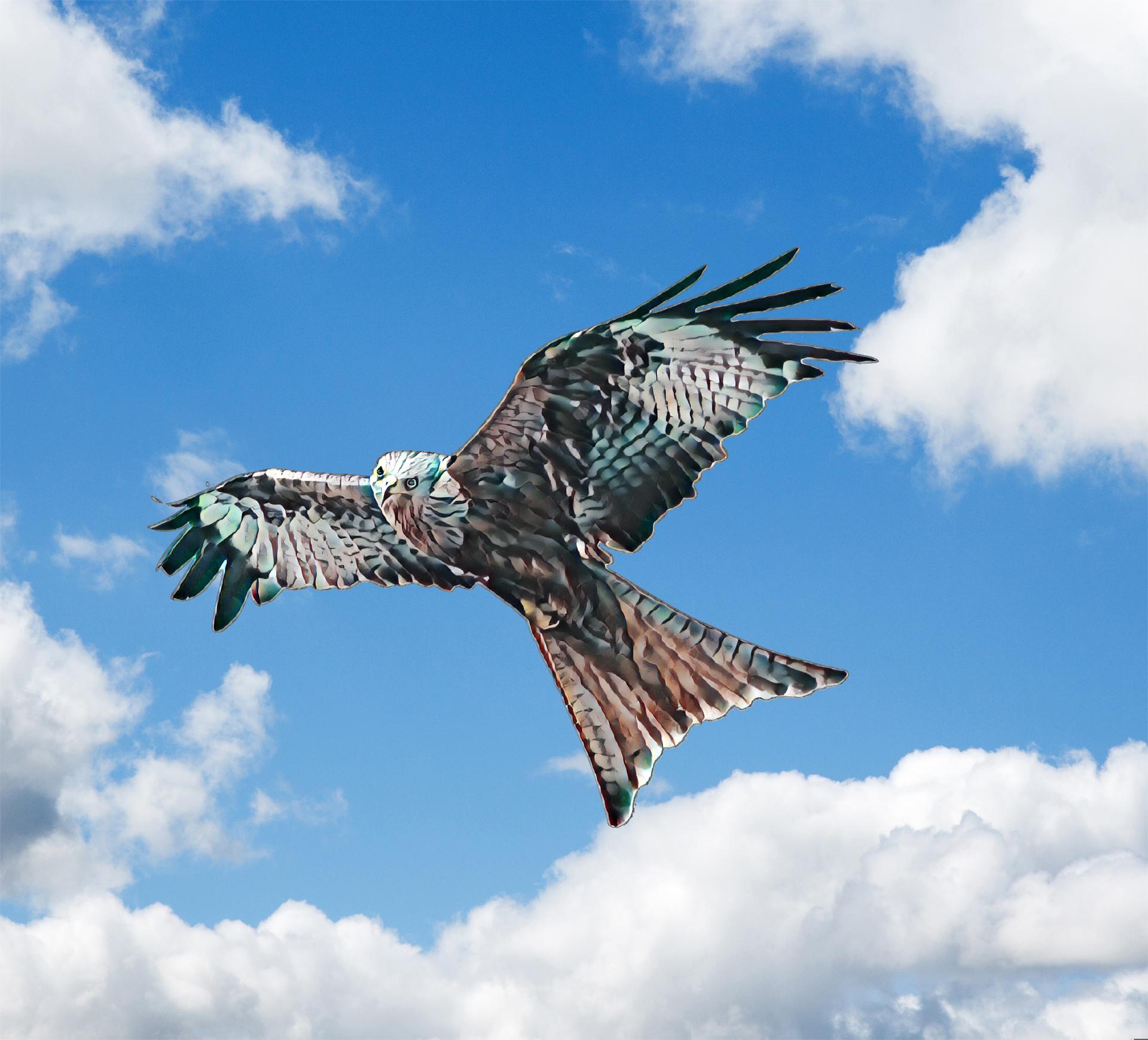}
\\
d) Masked then stylized &
e) Composition of (d) &
f) Masked stylization
\end{tabular}
\end{center}
   \caption{Another example, similar to Fig.~\ref{fig:masked1}, using a different style image.
   }
\label{fig:masked2b}
\end{figure}

\begin{figure}
\newcommand{\figwidth}{0.30\linewidth}
\begin{center}
\begin{tabular}{ccc}
\includegraphics[width=0.27\linewidth]{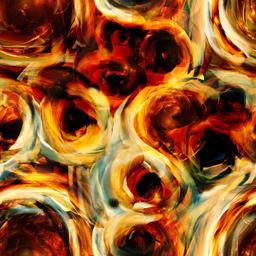} &
\includegraphics[width=\figwidth]{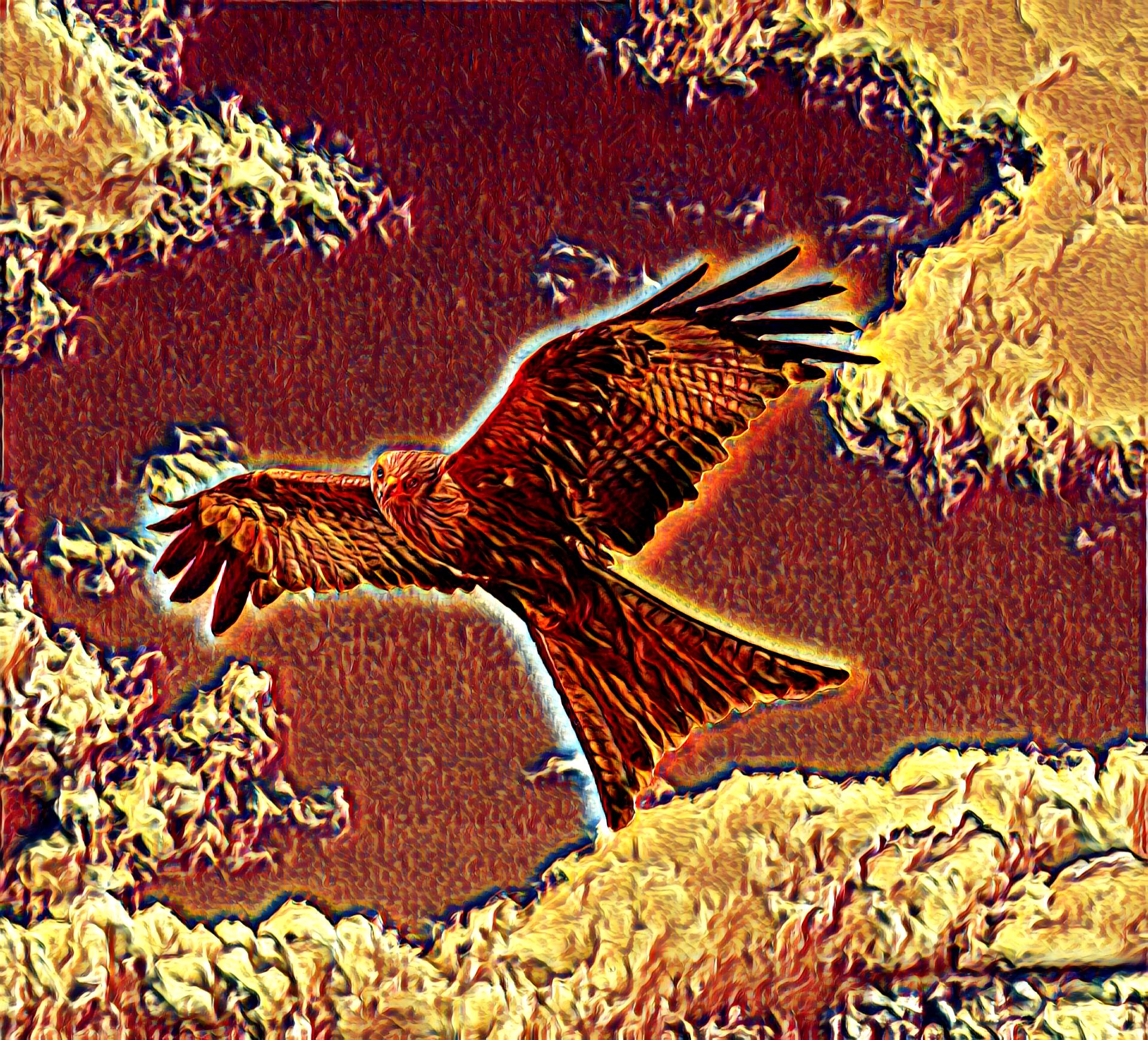} &
\includegraphics[width=\figwidth]{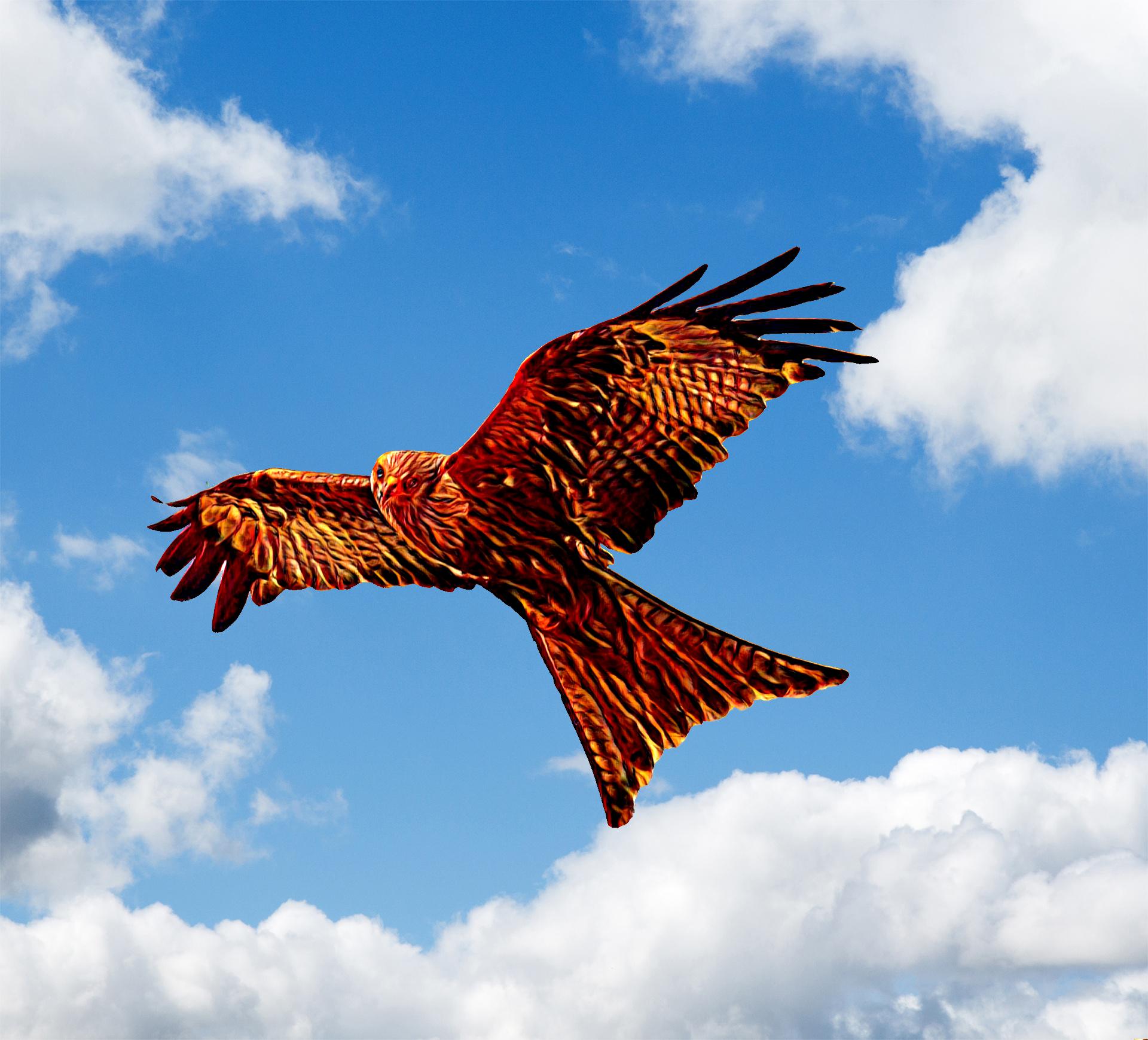} 
\\
a) Style Image &
b) Entire stylized image &
c) Composition of (b)
\\
\\
\includegraphics[width=\figwidth]{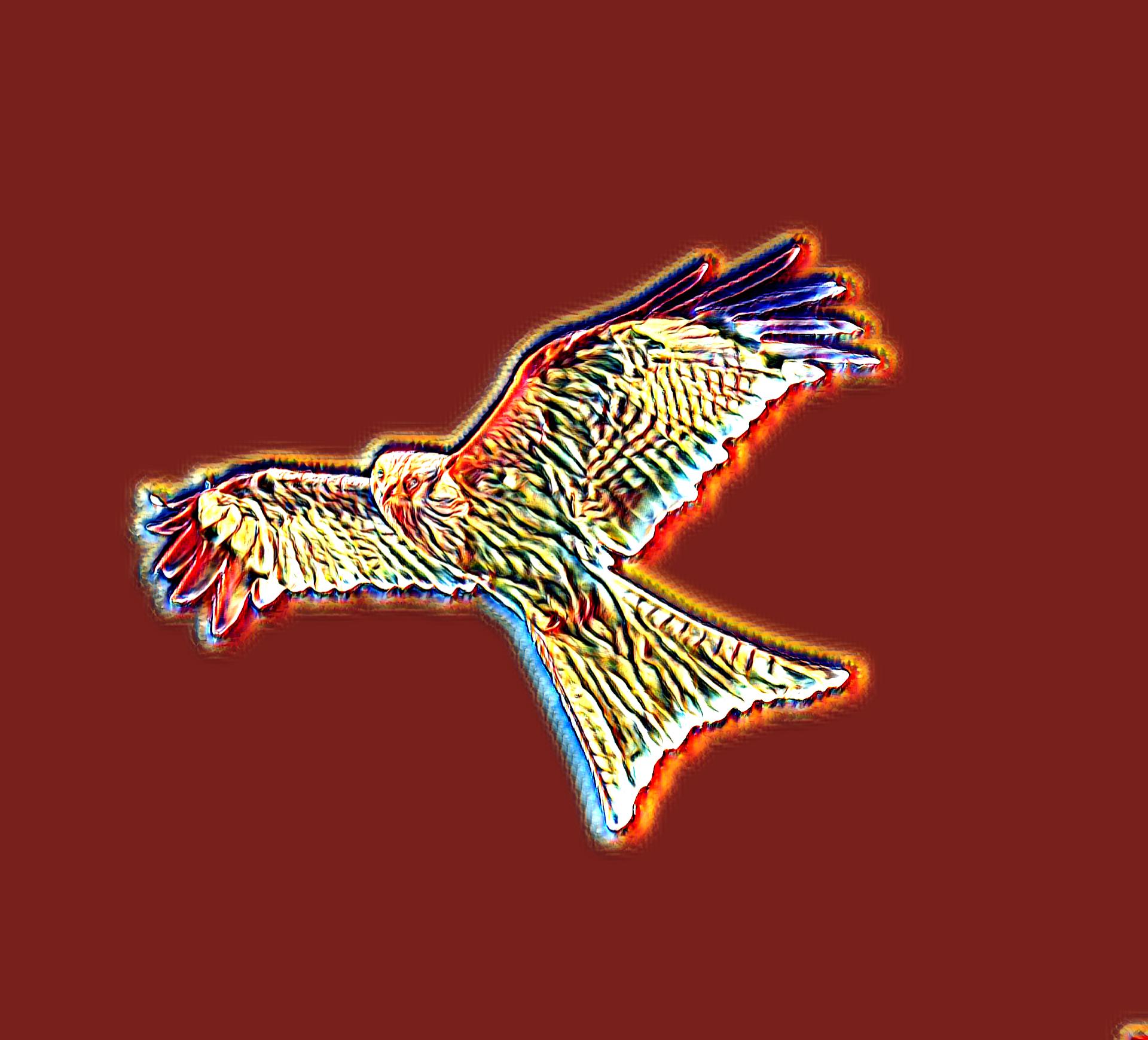} &
\includegraphics[width=\figwidth]{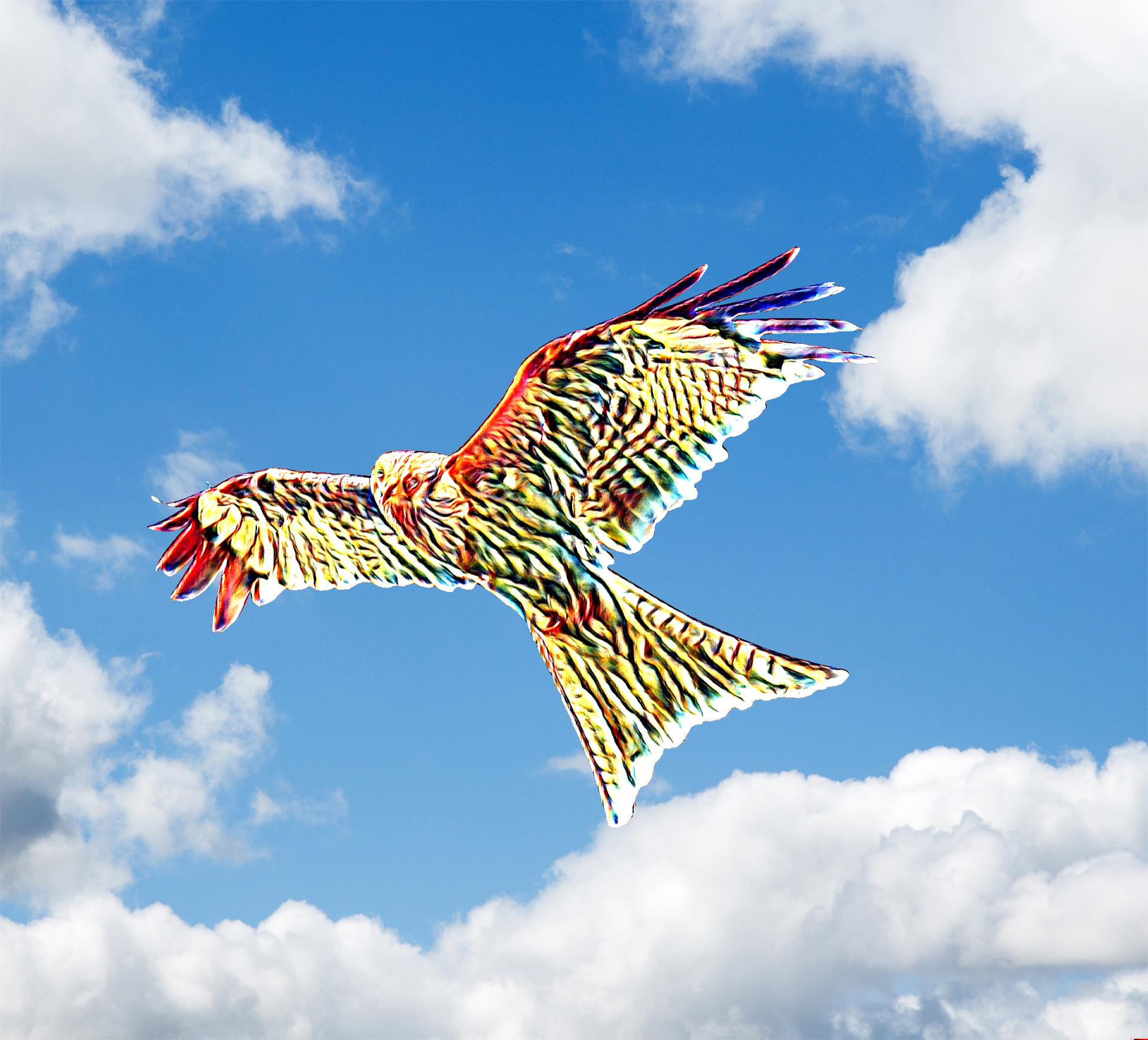} &
\includegraphics[width=\figwidth]{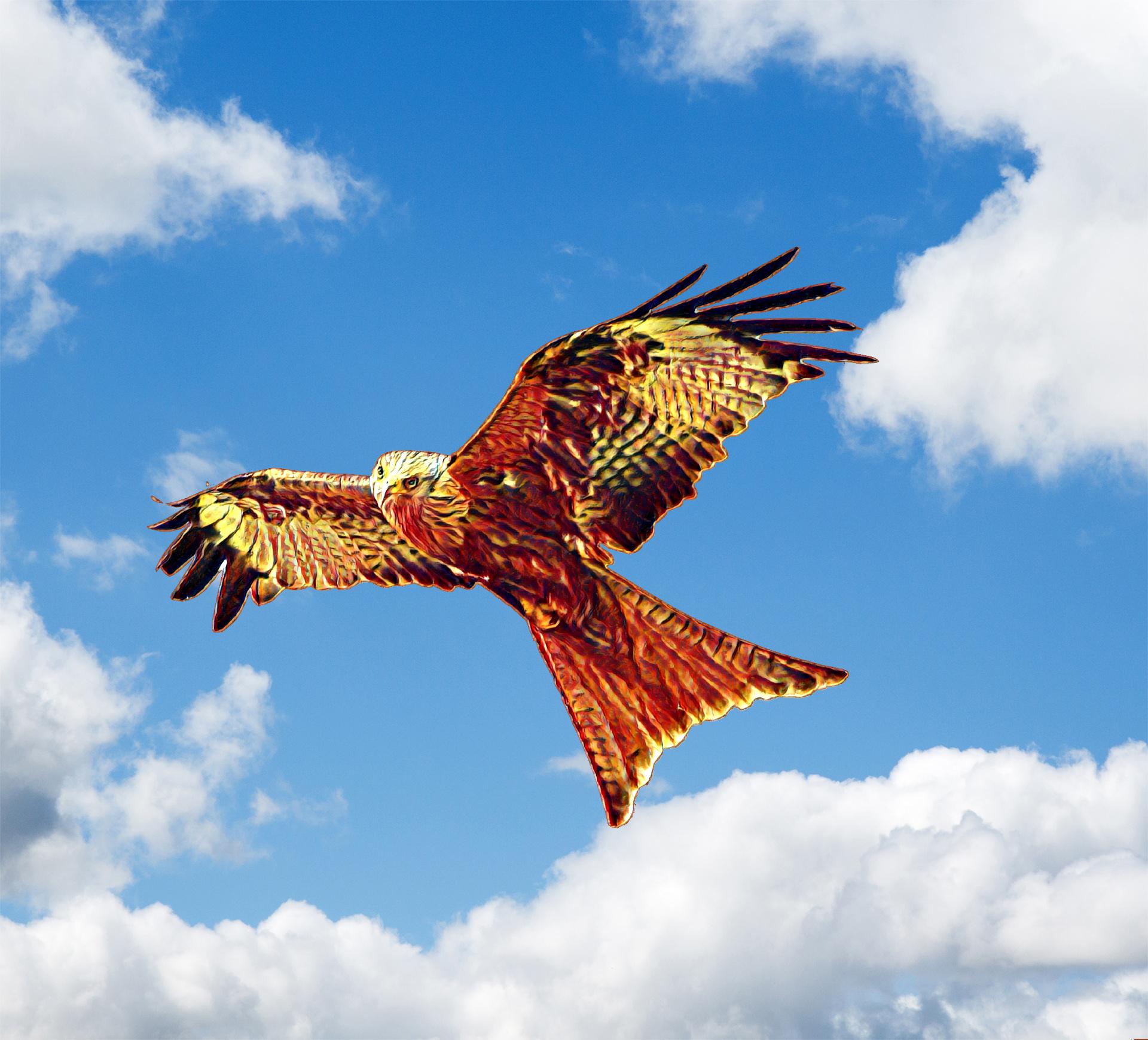}
\\
d) Masked then stylized &
e) Composition of (d) &
f) Masked stylization
\end{tabular}
\end{center}
   \caption{Another example, similar to Fig.~\ref{fig:masked1}, using a different style image.
   }
\label{fig:masked3a}
\end{figure}

\begin{figure}
\newcommand{\figwidth}{0.30\linewidth}
\begin{center}
\begin{tabular}{ccc}
\includegraphics[width=.18\linewidth]{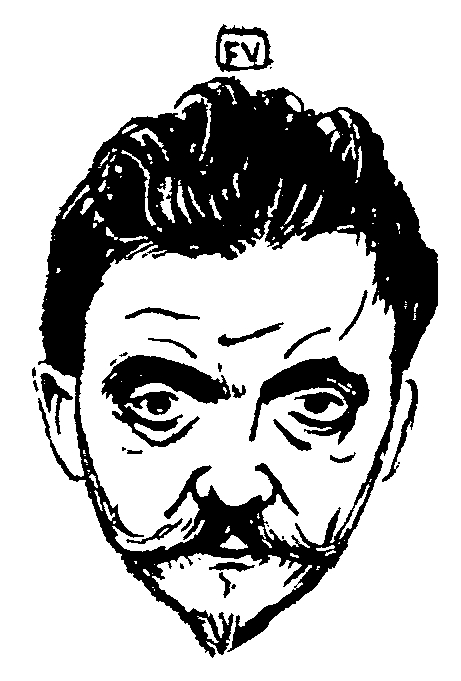} &
\includegraphics[width=\figwidth]{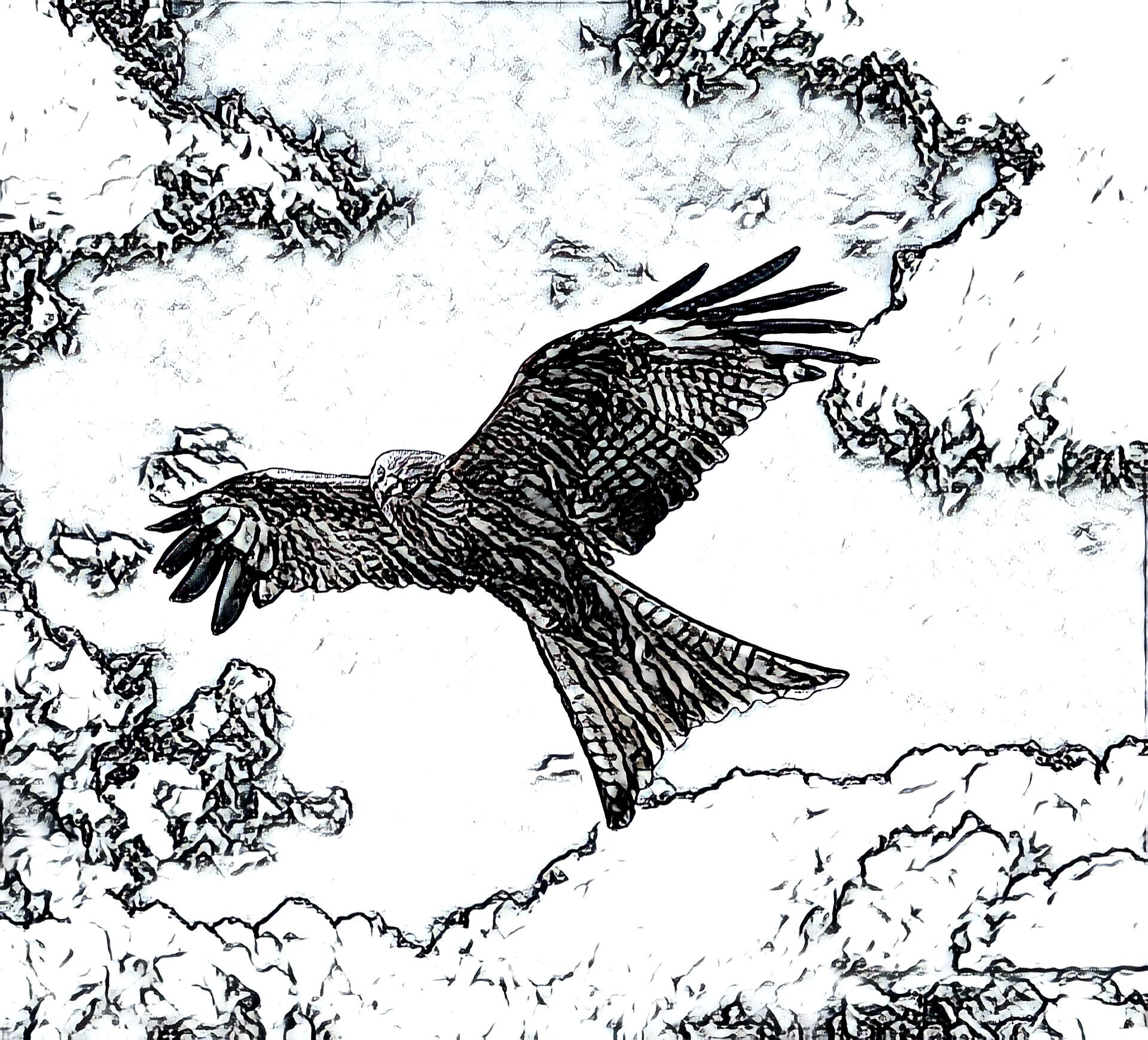} &
\includegraphics[width=\figwidth]{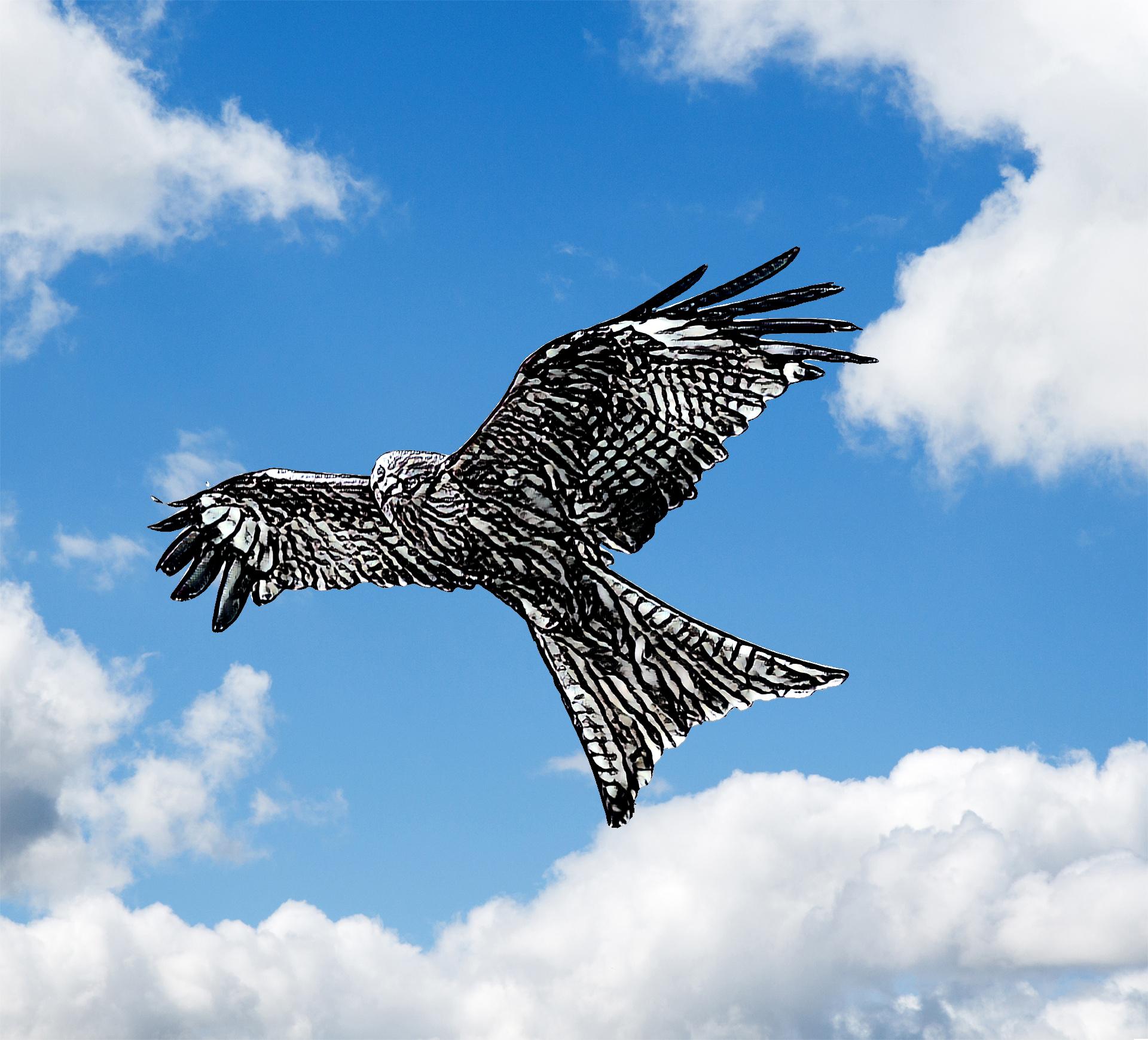} 
\\   
a) Style Image &
b) Entire stylized image &
c) Composition of (b) 
\\
\\
\includegraphics[width=\figwidth]{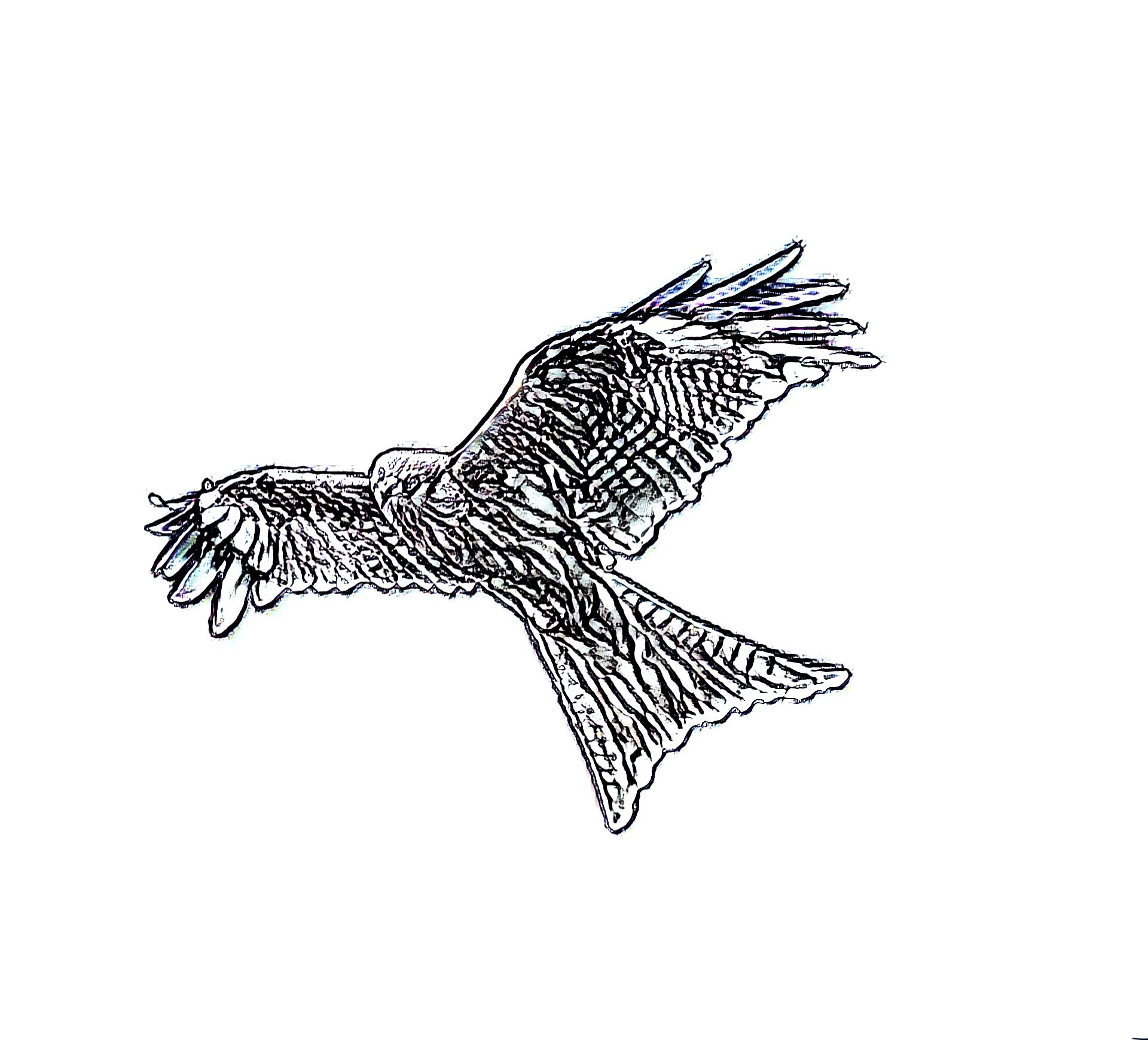} &
\includegraphics[width=\figwidth]{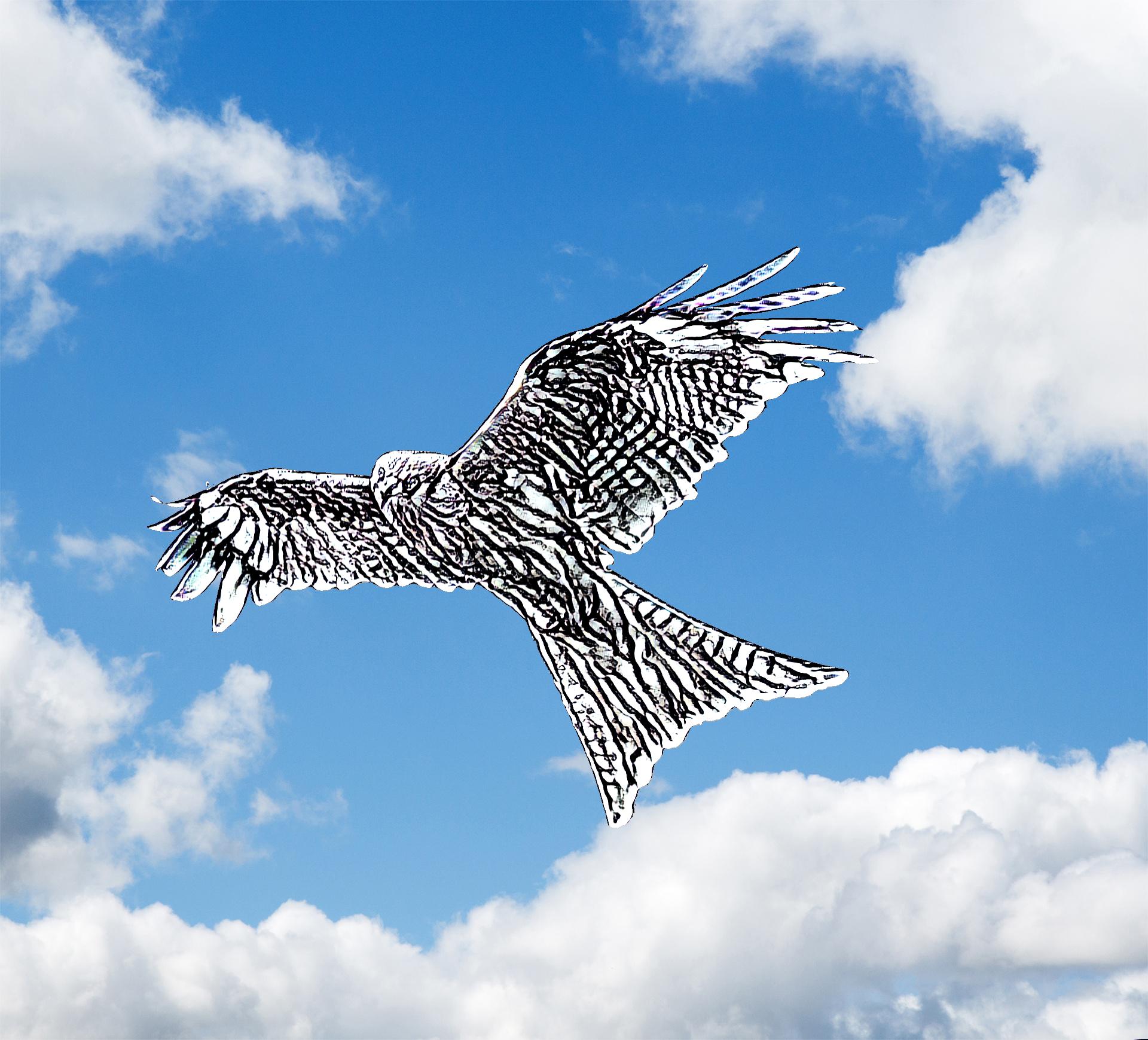} &
\includegraphics[width=\figwidth]{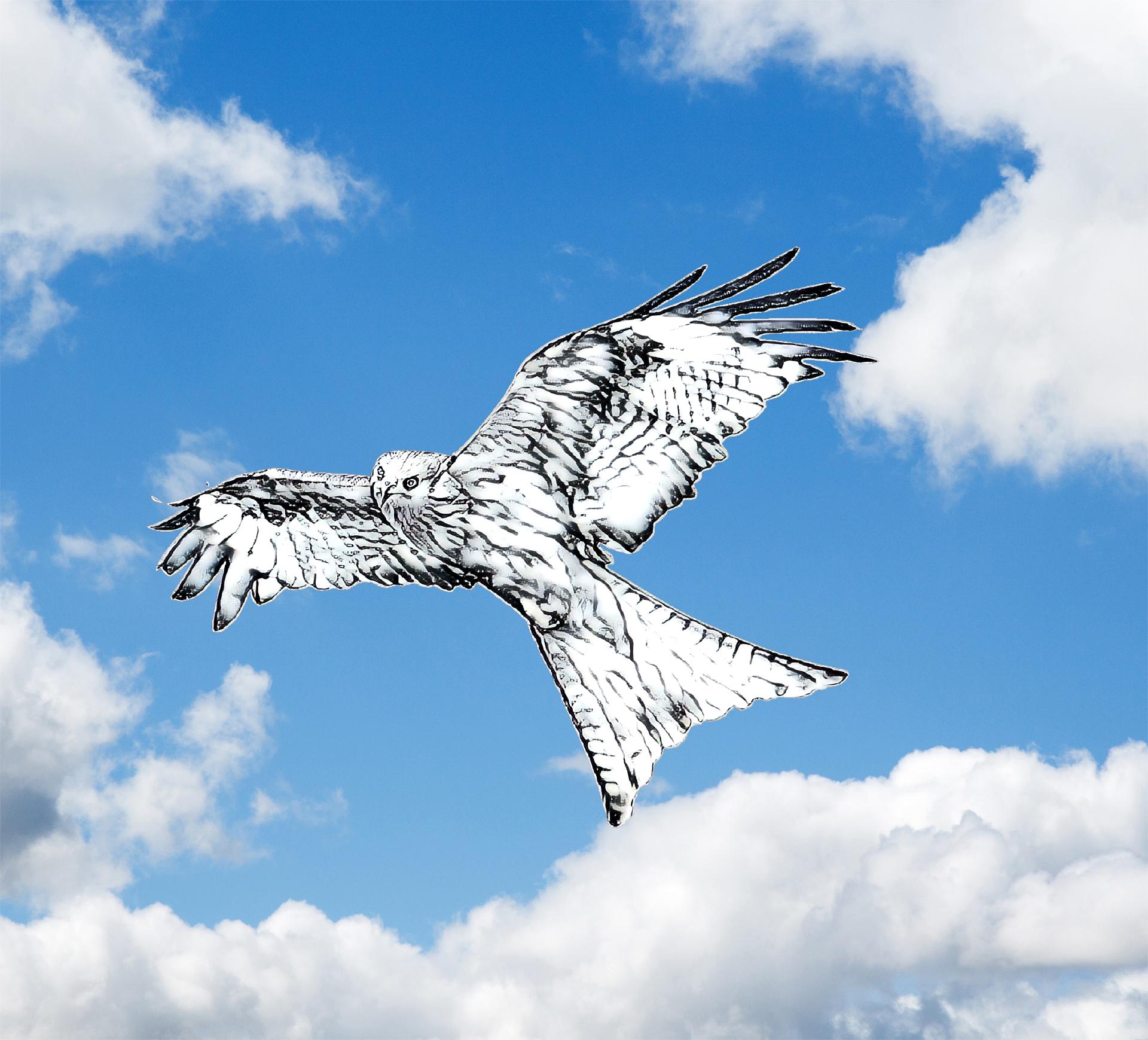}
\\   
d) Masked then stylized &
e) Composition of (d) &
f) Masked stylization
\end{tabular}
\end{center}
   \caption{Another example, similar to Fig.~\ref{fig:masked1}, using a different style image.
   }
\label{fig:masked3}
\end{figure}

\begin{figure}
\begin{center}
\begin{tabular}{c}
\begin{tabular}{cc}
     \includegraphics[width=.30\linewidth]{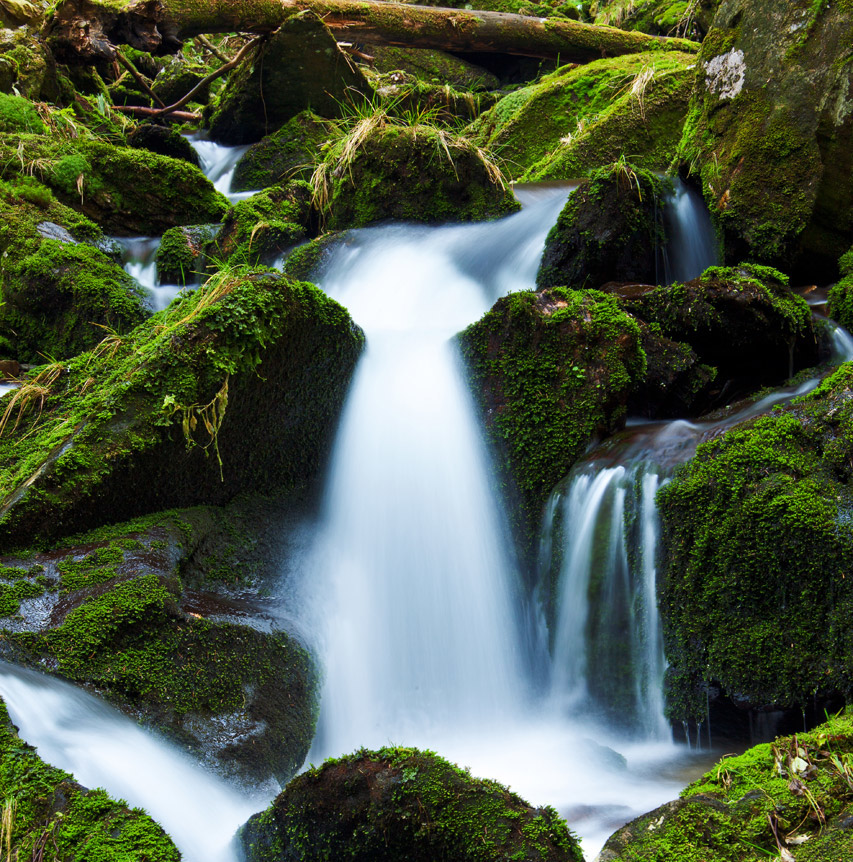} &
     \includegraphics[width=.30\linewidth]{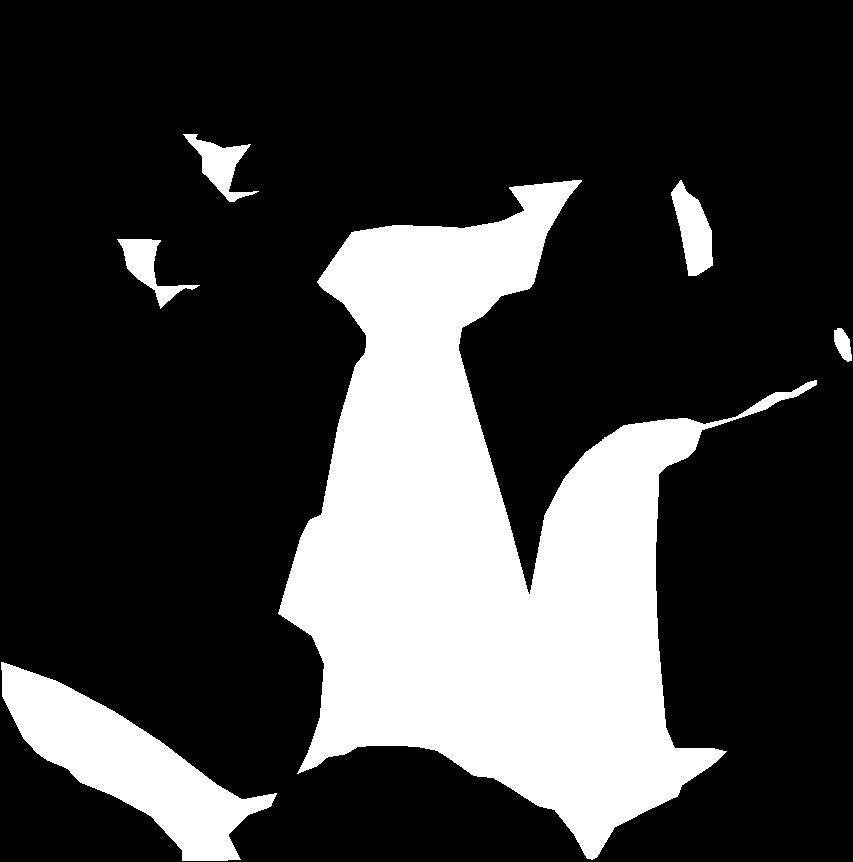} 
     \\ 
     Original Image & Mask
\end{tabular}
\\ \\
\begin{tabular}{ccc}
\includegraphics[width=.30\linewidth]{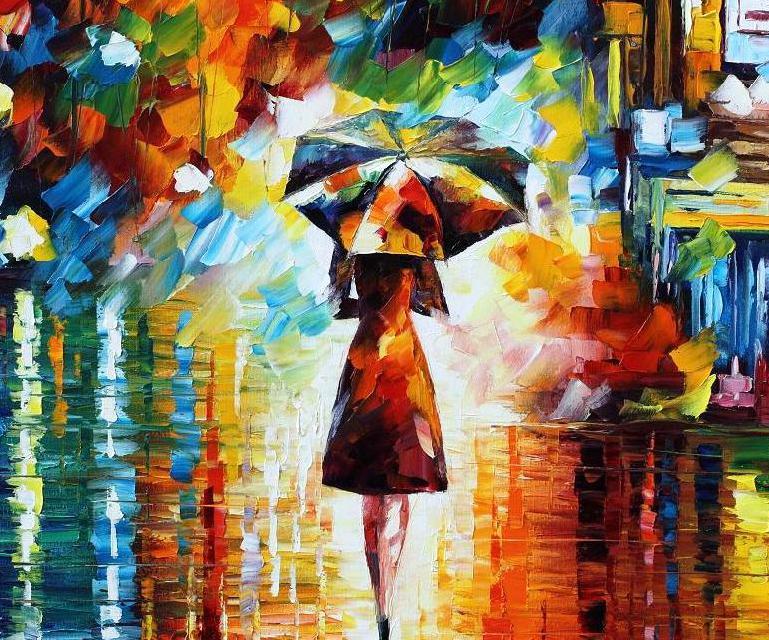} &
\includegraphics[width=.30\linewidth]{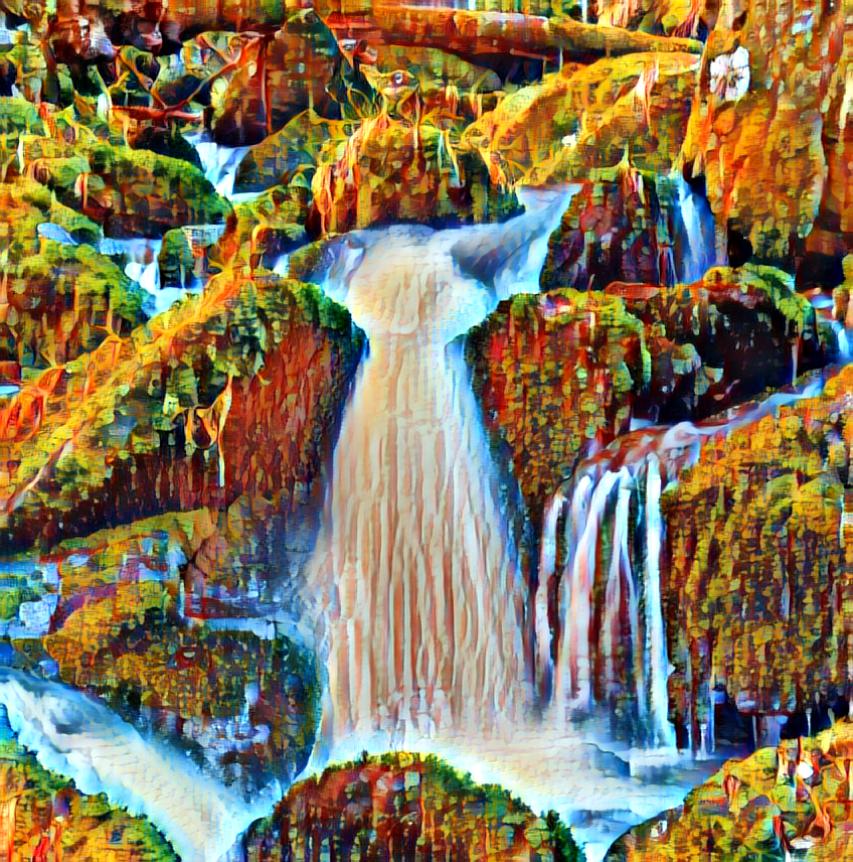} &
\includegraphics[width=.30\linewidth]{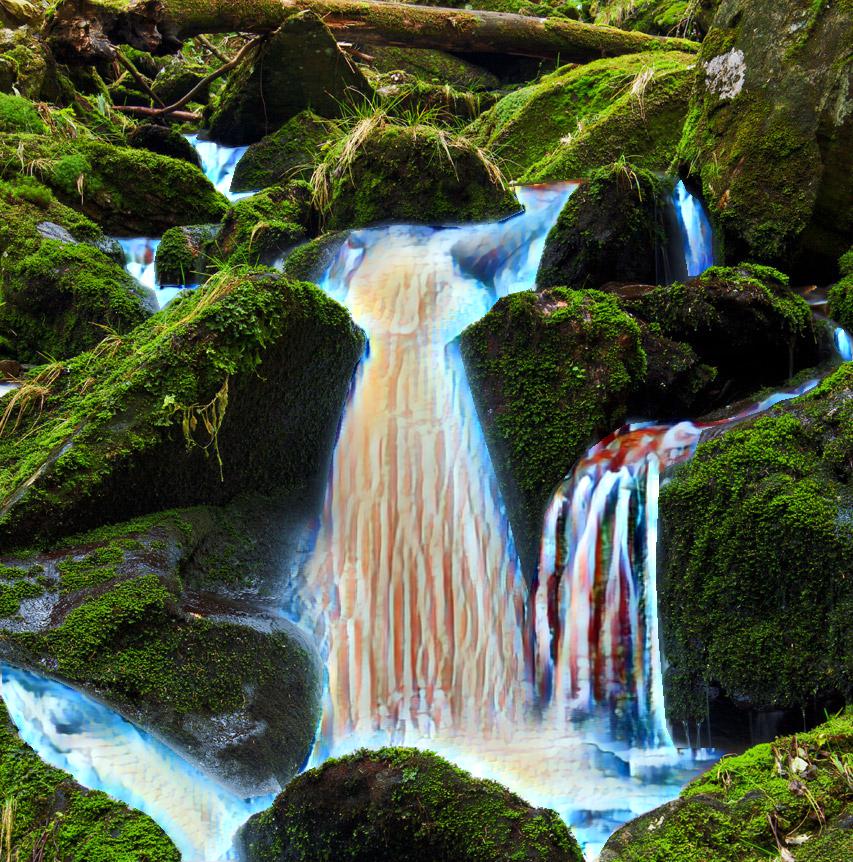} 
\\
a) Style Image &
b) Entire stylized image &
c) Composition of (b)
\\
\\
\includegraphics[width=.30\linewidth]{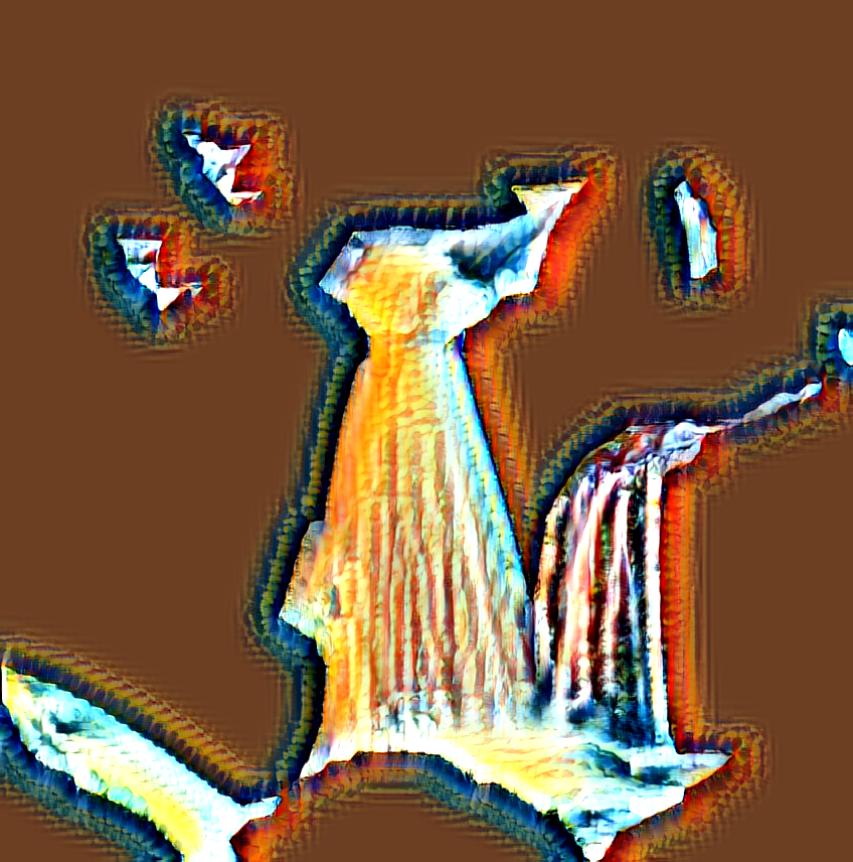} &
\includegraphics[width=.30\linewidth]{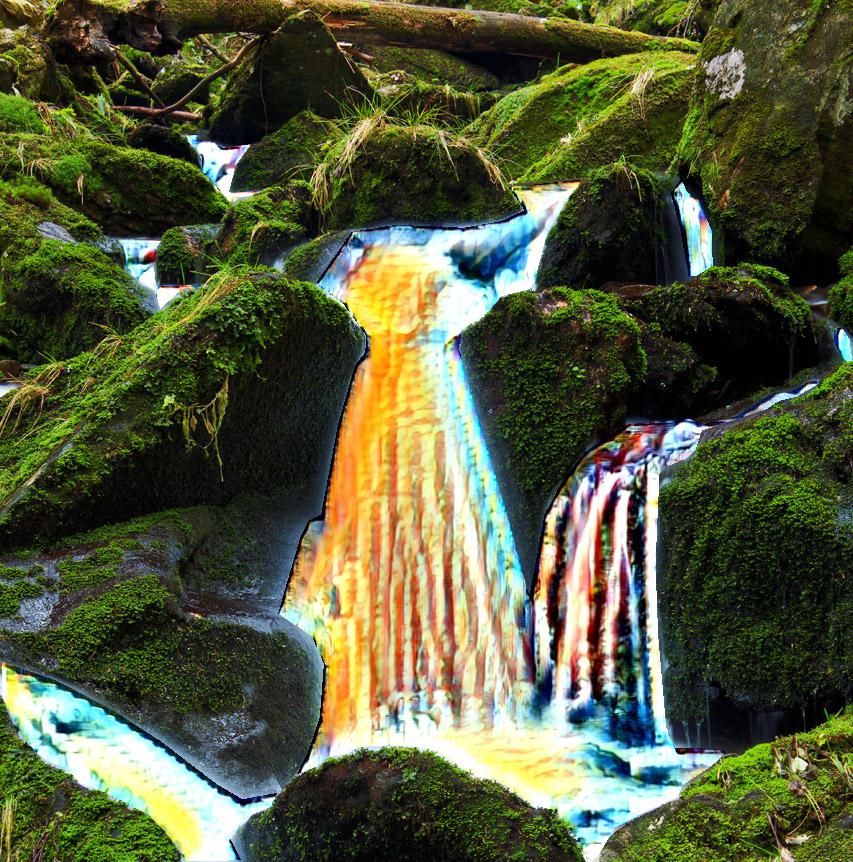} &
\includegraphics[width=.30\linewidth]{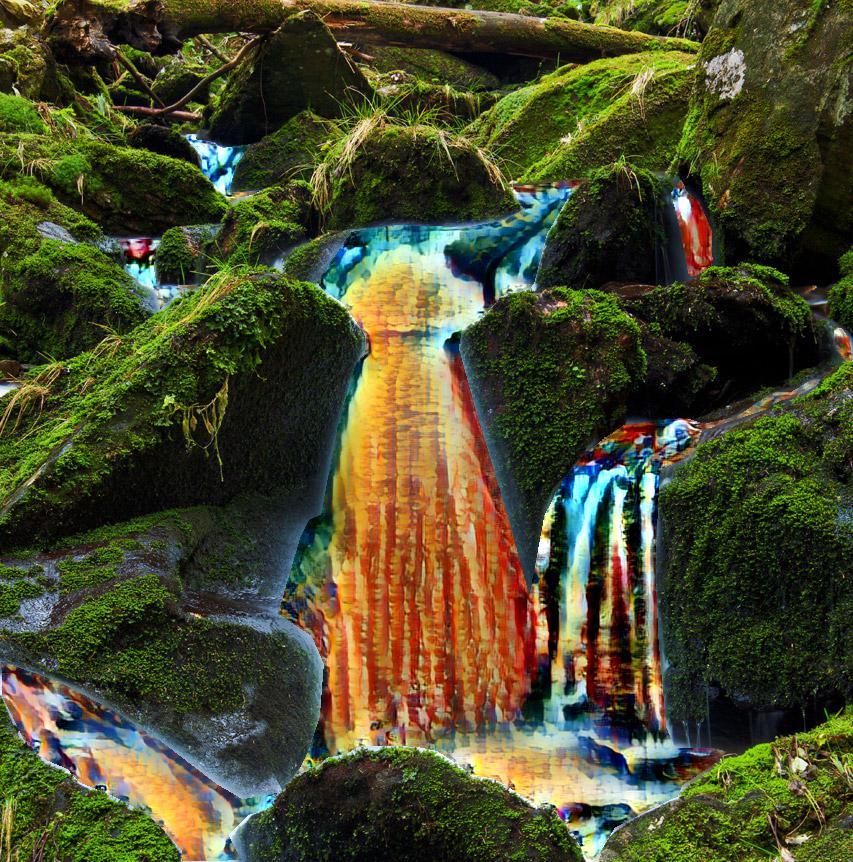}
\\   
d) Masked then stylized &
e) Composition of (d) &
f) Masked stylization
\end{tabular}
\end{tabular}
\end{center}
   \caption{Another example, similar to Fig.~\ref{fig:masked1}, using both a different content image and a different style image.
   }
\label{fig:masked4a}
\end{figure}

\begin{figure}
\begin{center}
\begin{tabular}{c}
\begin{tabular}{cc}
     \includegraphics[width=.30\linewidth]{Figures/stream.jpg} &
     \includegraphics[width=.30\linewidth]{Figures/stream_mask2.jpg} 
     \\ 
     Original Image & Mask
\end{tabular}
\\ \\
\begin{tabular}{ccc}
\includegraphics[width=.30\linewidth]{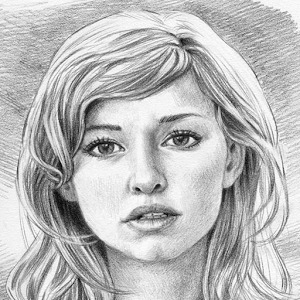} &
\includegraphics[width=.30\linewidth]{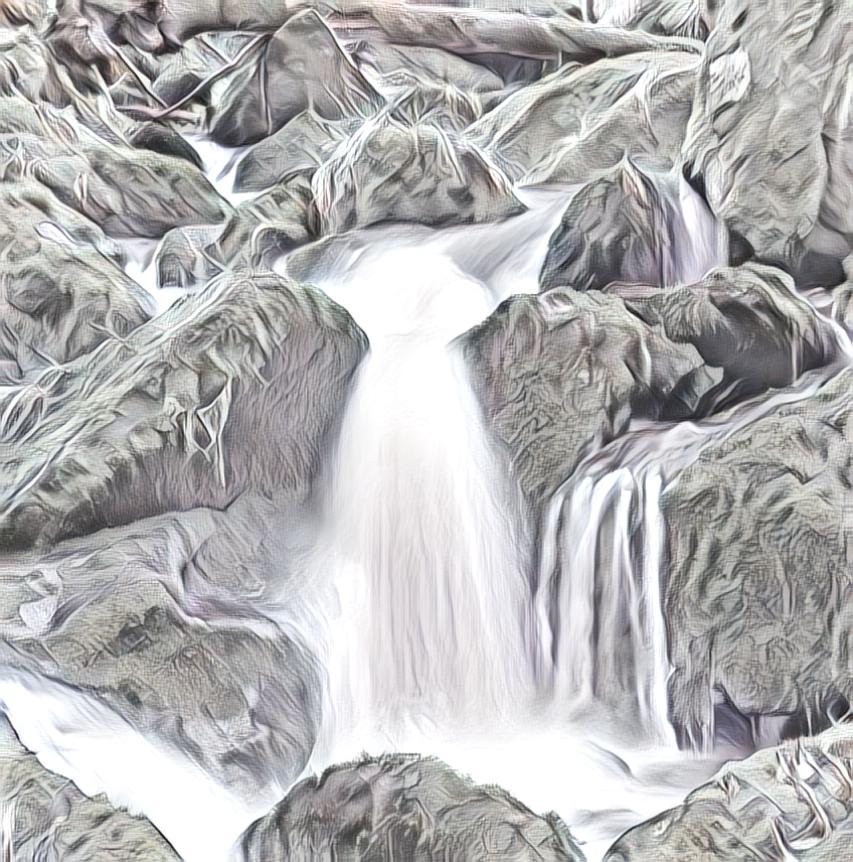} &
\includegraphics[width=.30\linewidth]{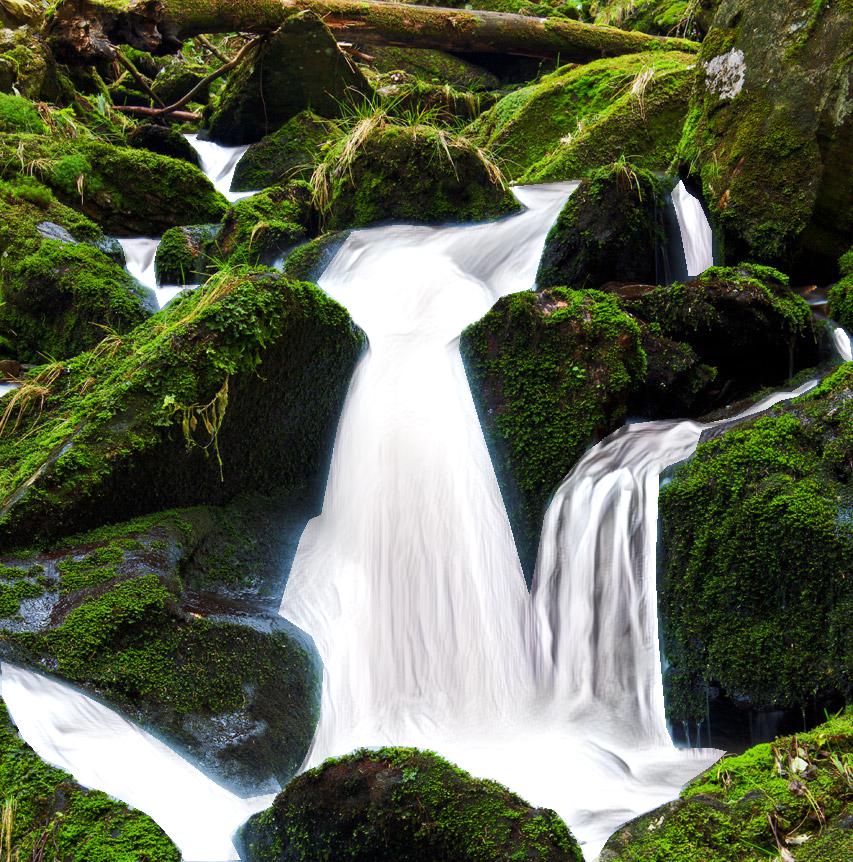} 
\\
a) Style Image &
b) Entire stylized image &
c) Composition of (b)
\\
\\
\includegraphics[width=.30\linewidth]{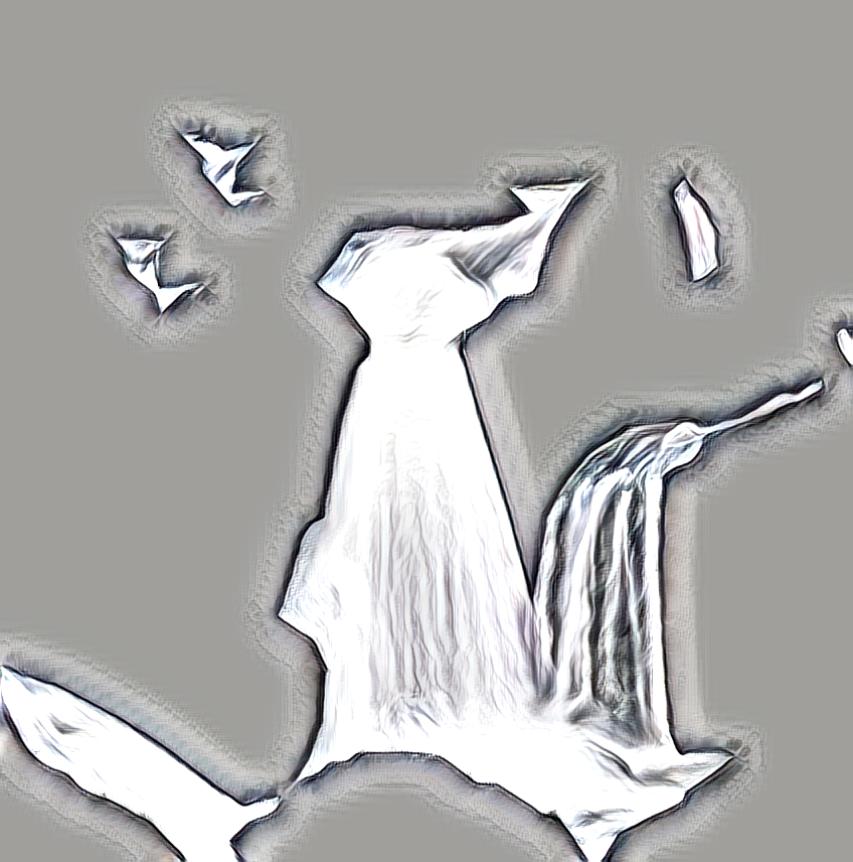} &
\includegraphics[width=.30\linewidth]{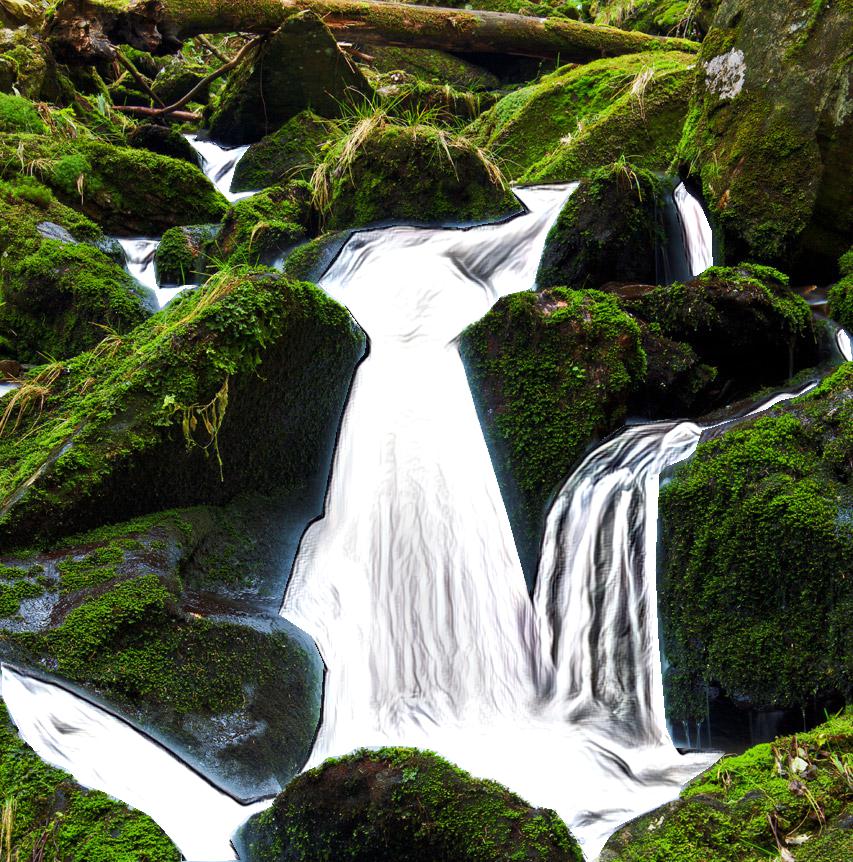} &
\includegraphics[width=.30\linewidth]{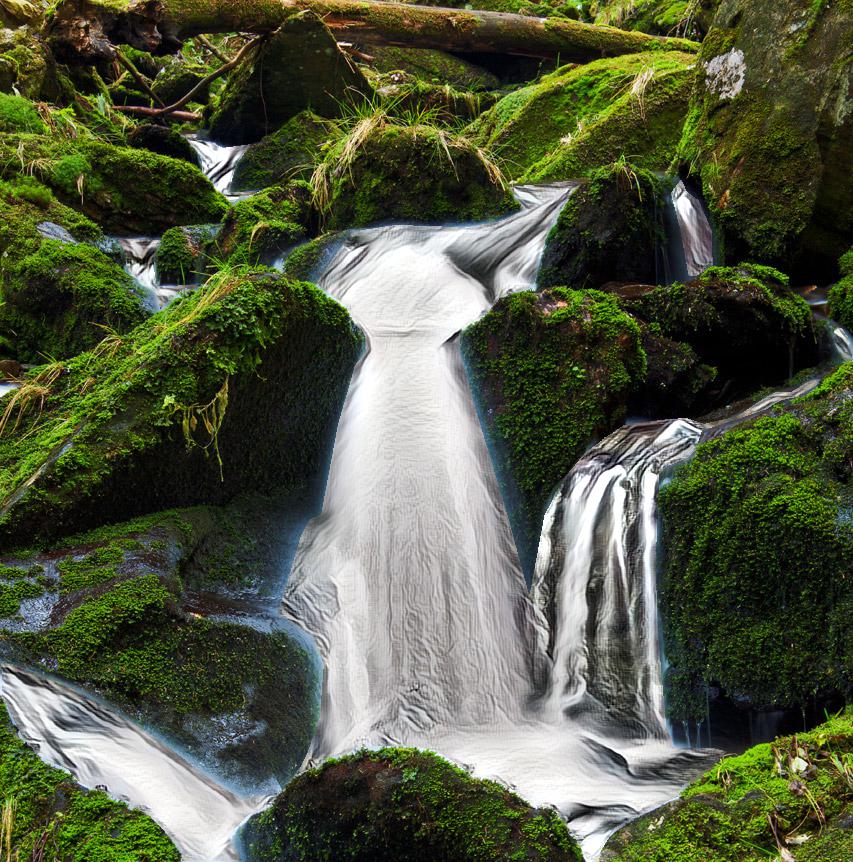}
\\   
d) Masked then stylized &
e) Composition of (d) &
f) Masked stylization
\end{tabular}
\end{tabular}
\end{center}
   \caption{Another example, similar to Fig.~\ref{fig:masked4a}, using a different style image.
   }
\label{fig:masked4}
\end{figure}

\noindent
We provide additional examples in Figs.~\ref{fig:masked2a}--\ref{fig:masked4}. 

~

\begin{figure}
\begin{center}
\begin{tabular}{c}
\begin{tabular}{c}
    \includegraphics[width=.30\linewidth]{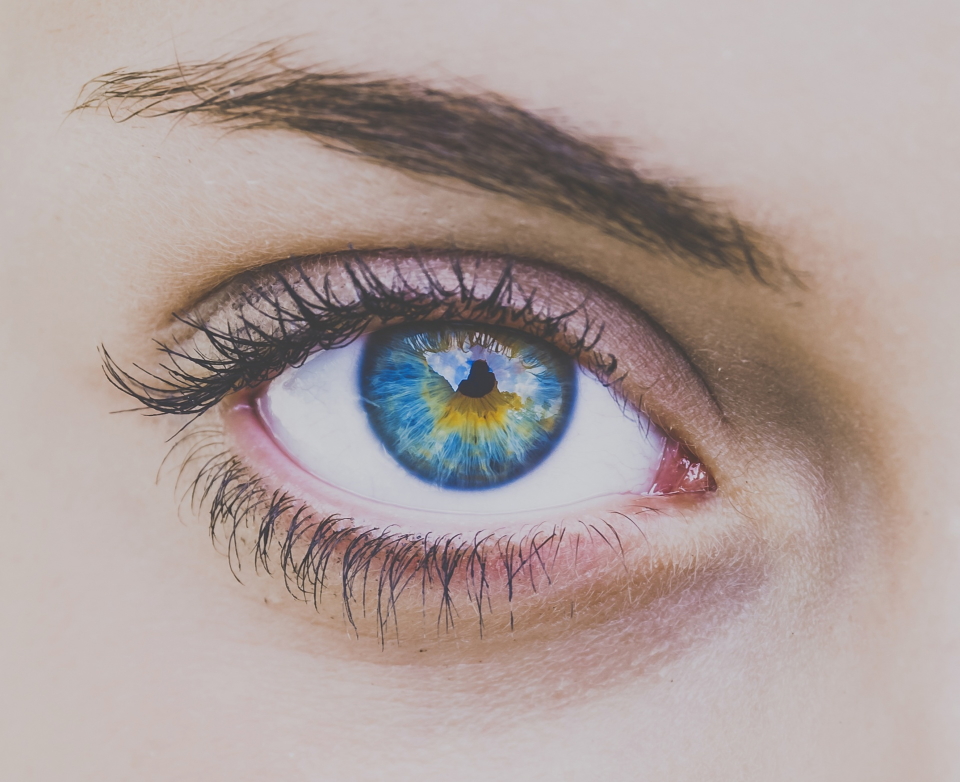} \\
    Original Image
\end{tabular}
\\ \\
\begin{tabular}{c}
\begin{tabular}{ccc}
    \includegraphics[width=.22\linewidth]{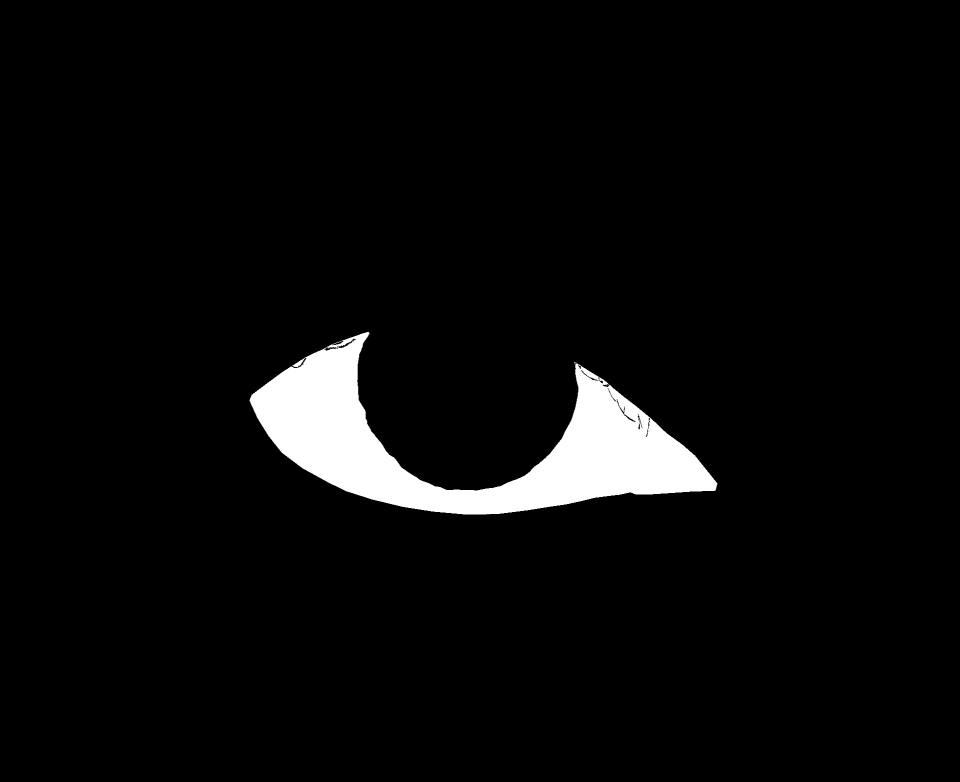} &
    \includegraphics[width=.22\linewidth]{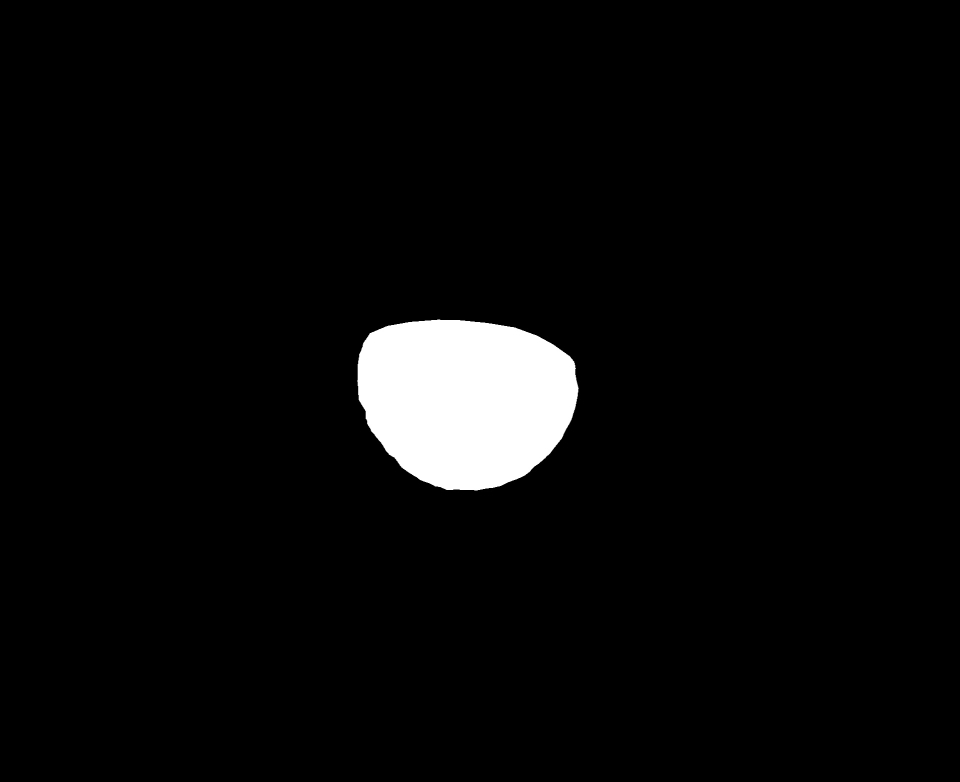} &
    \includegraphics[width=.22\linewidth]{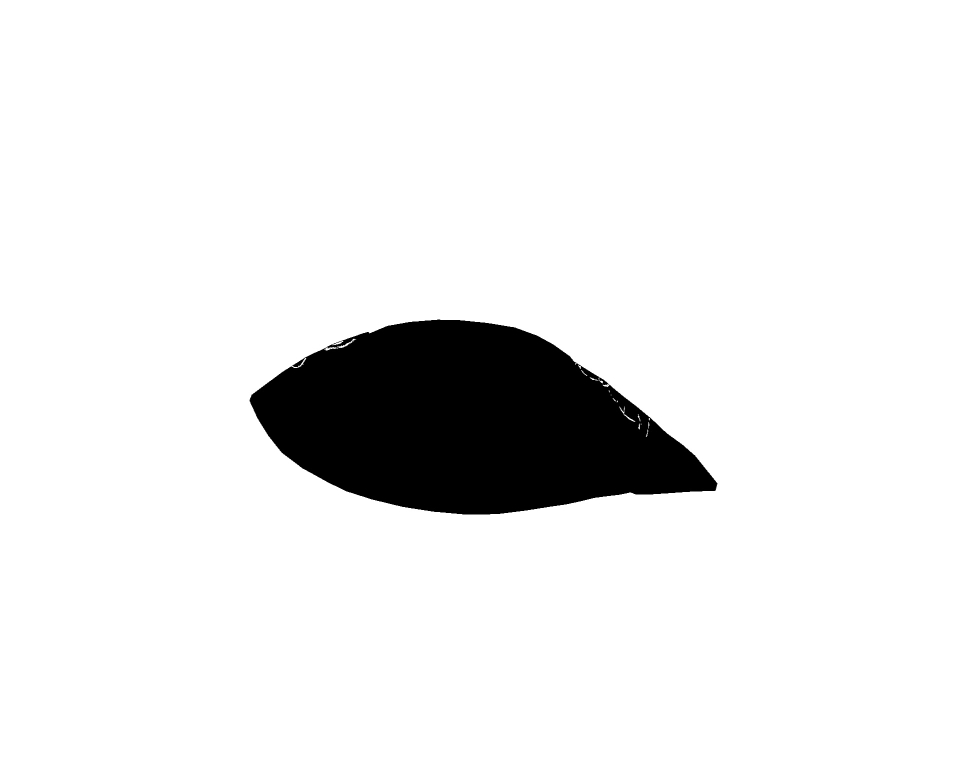} 
\\
    \includegraphics[width=.22\linewidth]{Figures/style9.jpg} &
    \includegraphics[width=.22\linewidth]{Figures/style3.jpg} &
    \includegraphics[width=.22\linewidth]{Figures/style0.jpg}
\end{tabular}
\\ 
Masked regions and respective style images to be applied
\end{tabular}
\\ \\
\begin{tabular}{ccc}
\includegraphics[width=.30\linewidth]{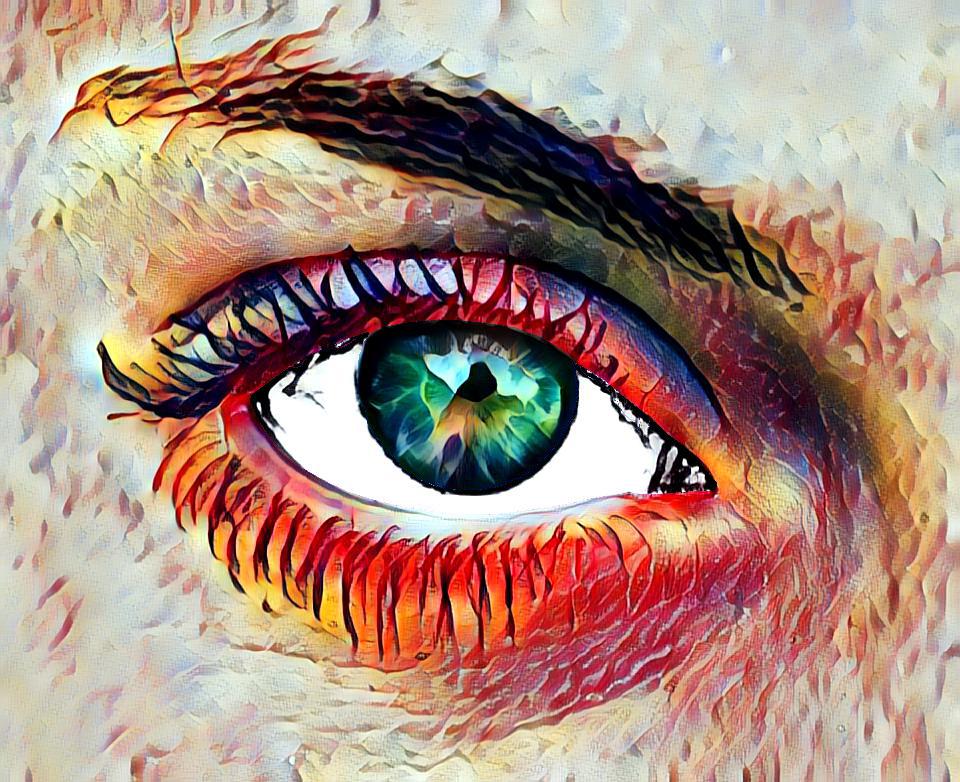} &
\includegraphics[width=.30\linewidth]{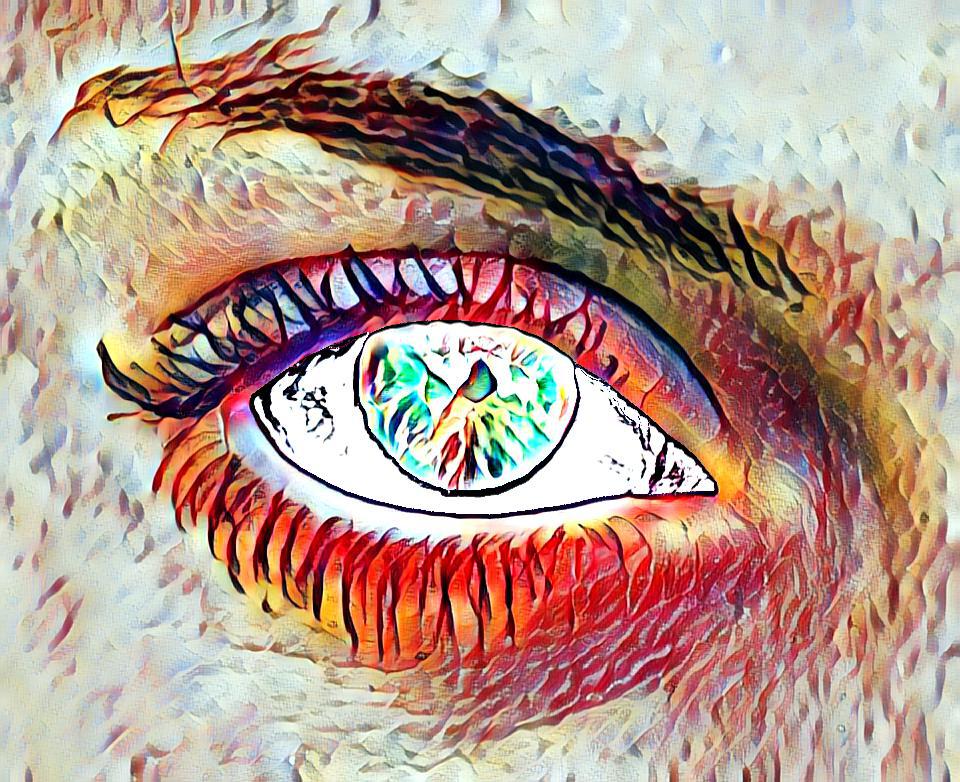} &
\includegraphics[width=.30\linewidth]{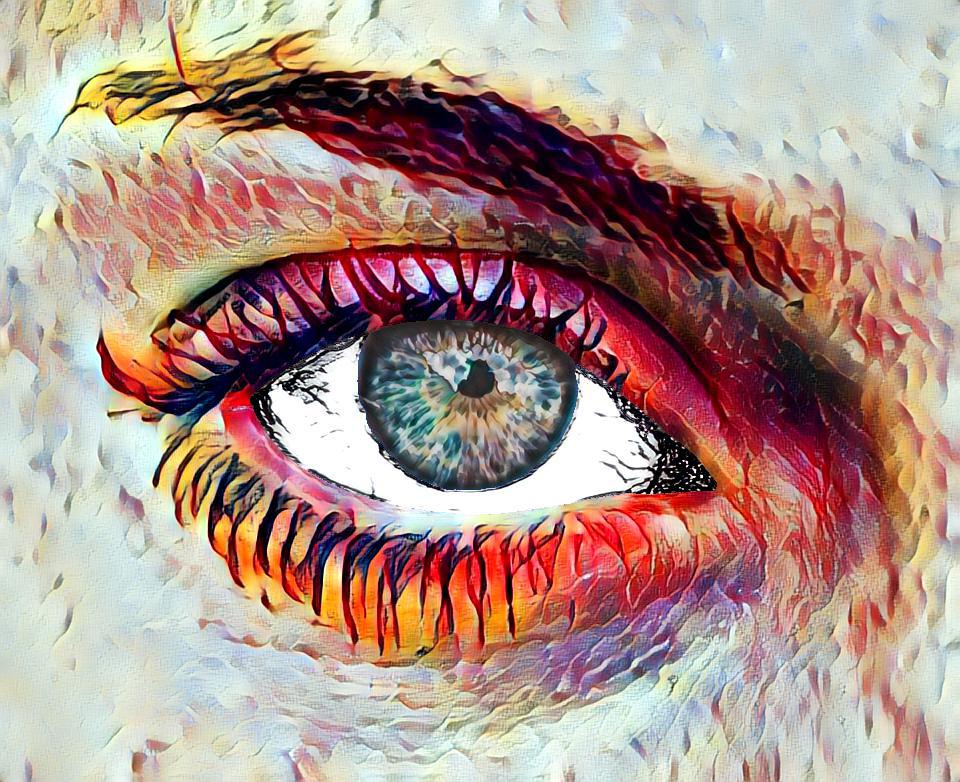}
\\ 
Stylization of entire images & Stylization of pre-masked images & Masked stylization \\
followed by composition & followed by composition & then composition
\end{tabular}
\end{tabular}
\end{center}
   \caption{ Examples of using multiple styles and masks.
   The original image (top) is separated into three masked regions, each with their respective style to be applied (middle).
   Stylizing the entire image using each of the three styles distributes characteristics of the respective styles images across the entire image before masking, resulting in a composition where each region captures only a portion of its respective style (bottom left).
   Pre-masking each part of the image and applying the respective styles likewise distributes style characteristics across the entire image since the stylization cannot ignore the pre-masked regions, again resulting in composited regions that capture only a portion of their respective styles (bottom center).
   Using our masked stylization approach, the stylization is applied to each region without regard to the rest of the image, resulting in a composition where each region reflects more of its respective source style (bottom right).
   %
   }
\label{fig:masked5a}
\end{figure}

\begin{figure}
\begin{center}
\begin{tabular}{c}
\begin{tabular}{c}
    \includegraphics[width=.30\linewidth]{Figures/eye.jpg} \\
    Original Image
\end{tabular}
\\ \\
\begin{tabular}{c}
\begin{tabular}{ccc}
    \includegraphics[width=.22\linewidth]{Figures/eye_mask1.jpg} &
    \includegraphics[width=.22\linewidth]{Figures/eye_mask2.jpg} &
    \includegraphics[width=.22\linewidth]{Figures/eye_mask3.jpg} 
\\
    \includegraphics[width=.22\linewidth]{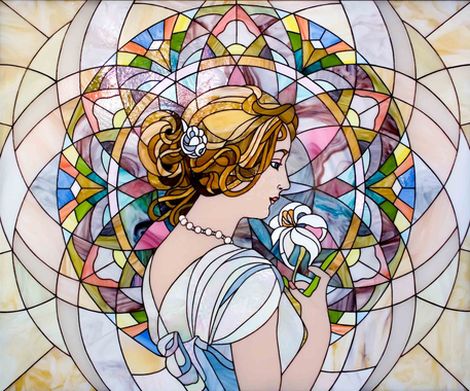} &
    \includegraphics[width=.144\linewidth]{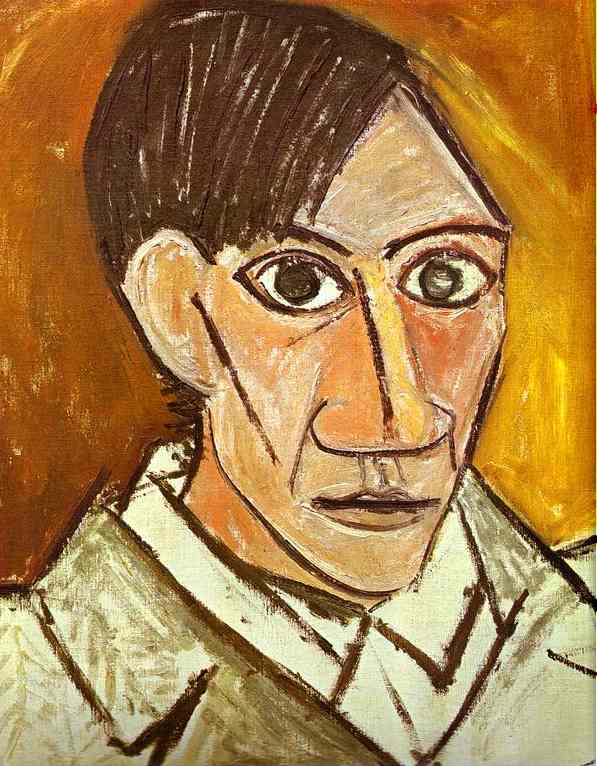} &
    \includegraphics[width=.181\linewidth]{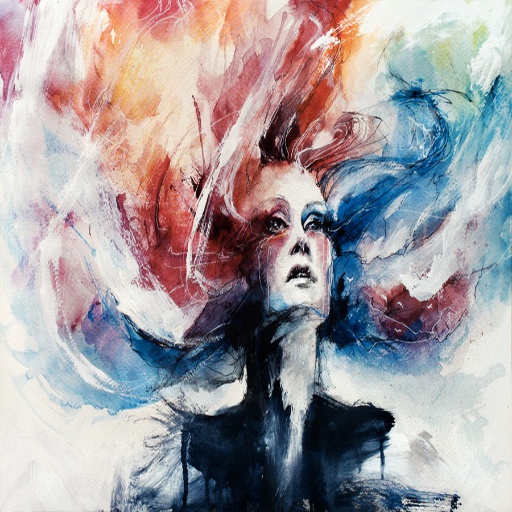}
\end{tabular}
\\ 
Masked regions and respective style images to be applied
\end{tabular}
\\ \\
\begin{tabular}{ccc}
\includegraphics[width=.30\linewidth]{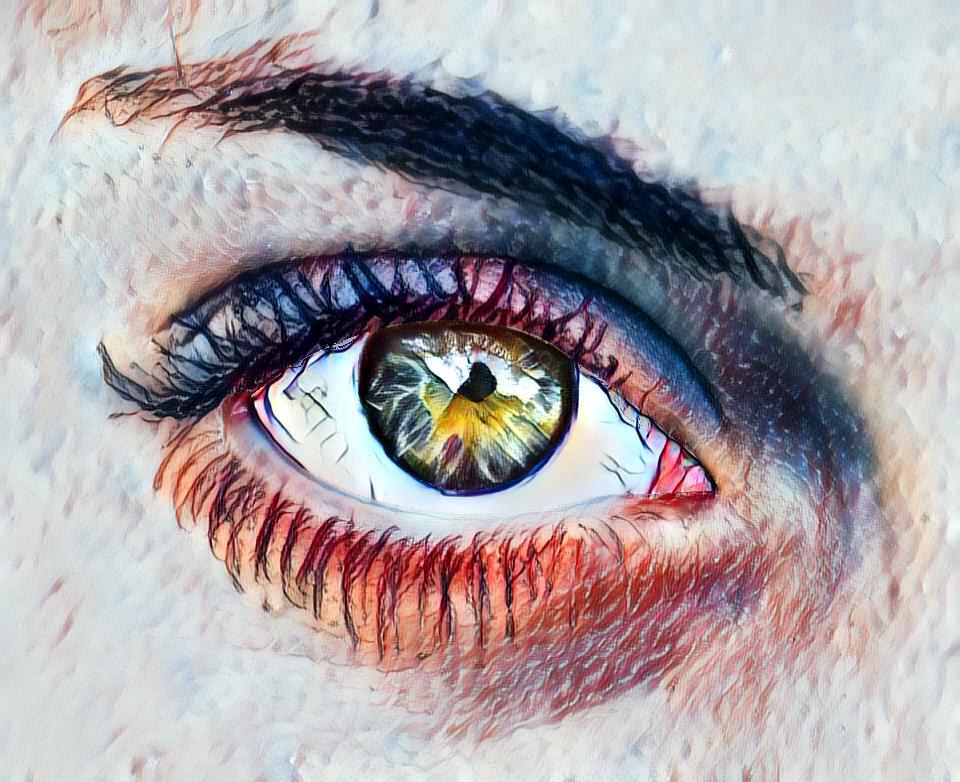} &
\includegraphics[width=.30\linewidth]{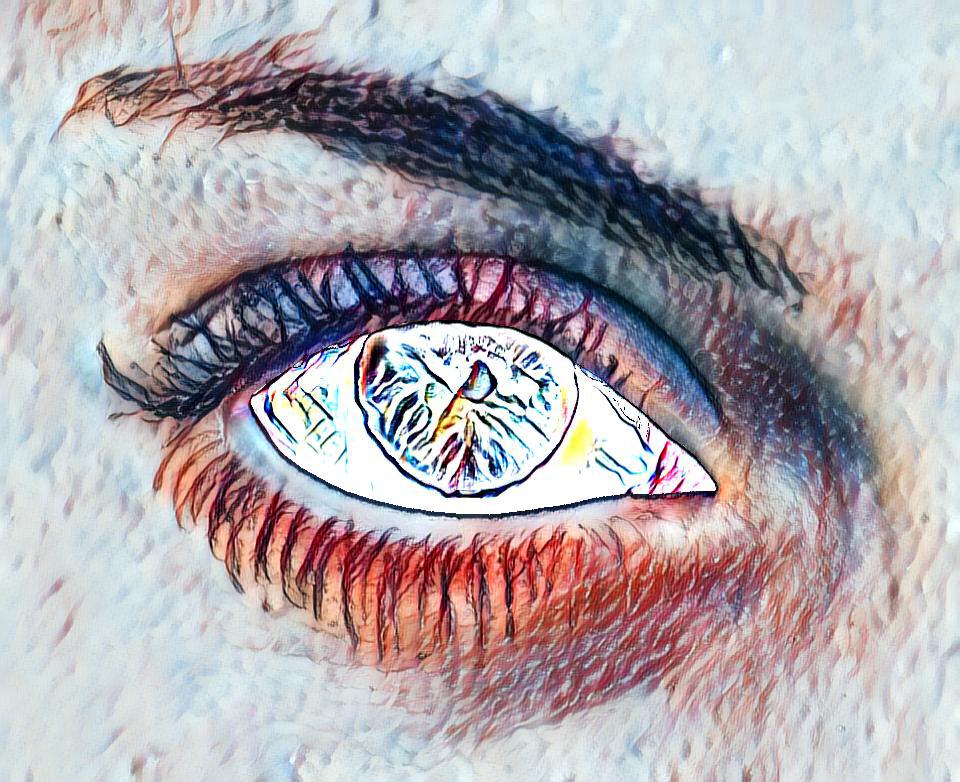} &
\includegraphics[width=.30\linewidth]{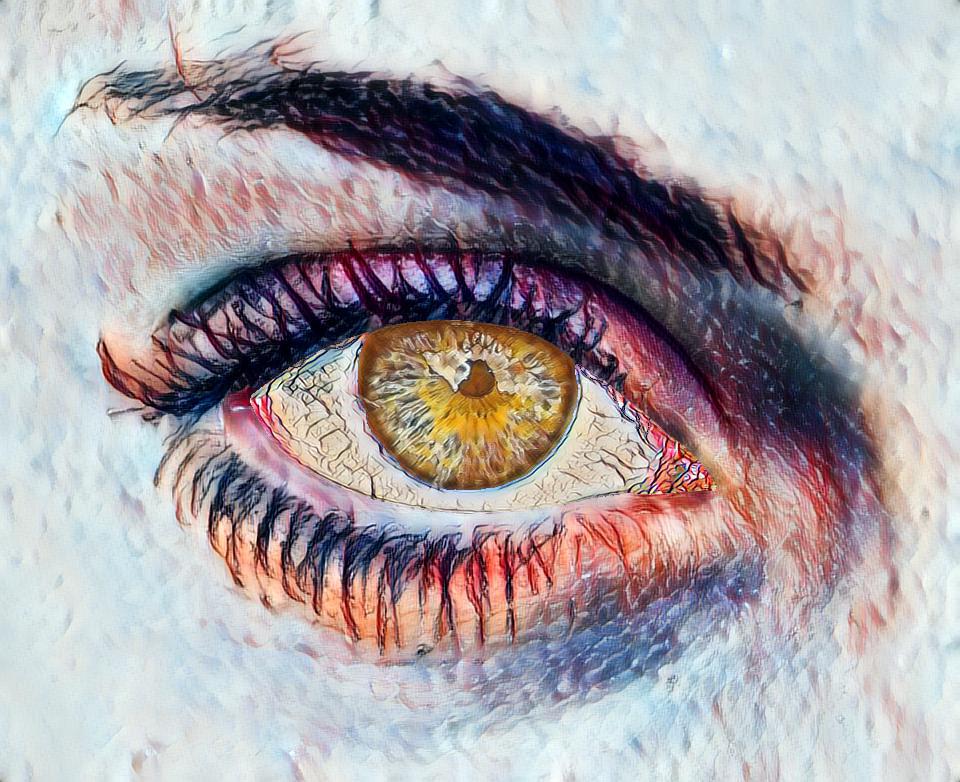}
\\ 
Stylization of entire images & Stylization of pre-masked images & Masked stylization \\
followed by composition & followed by composition & then composition
\end{tabular}
\end{tabular}
\end{center}
   \caption{ Another example using multiple styles and masks.
   The original content image and masked regions are the same as in in Fig.~\ref{fig:masked5a}, with a different set of styles applied to the regions.
   Again, masked stylization results in regions that each better represent the full content of the respective source style.
   }
\label{fig:masked5b}
\end{figure}

\begin{figure}
\begin{center}
\begin{tabular}{c}
\begin{tabular}{c}
    \includegraphics[width=.30\linewidth]{Figures/eye.jpg} \\
    Original Image
\end{tabular}
\\ \\
\begin{tabular}{c}
\begin{tabular}{ccc}
    \includegraphics[width=.22\linewidth]{Figures/eye_mask1.jpg} &
    \includegraphics[width=.22\linewidth]{Figures/eye_mask2.jpg} &
    \includegraphics[width=.22\linewidth]{Figures/eye_mask3.jpg} 
\\
    \includegraphics[width=.22\linewidth]{Figures/style3.jpg} &
    \includegraphics[width=.22\linewidth]{Figures/style4.jpg} &
    \includegraphics[width=.22\linewidth]{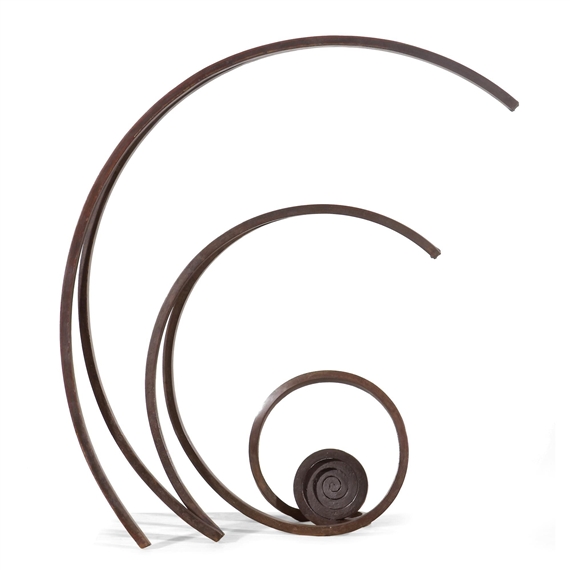}
\end{tabular}
\\ 
Masked regions and respective style images to be applied
\end{tabular}
\\ \\
\begin{tabular}{ccc}
\includegraphics[width=.30\linewidth]{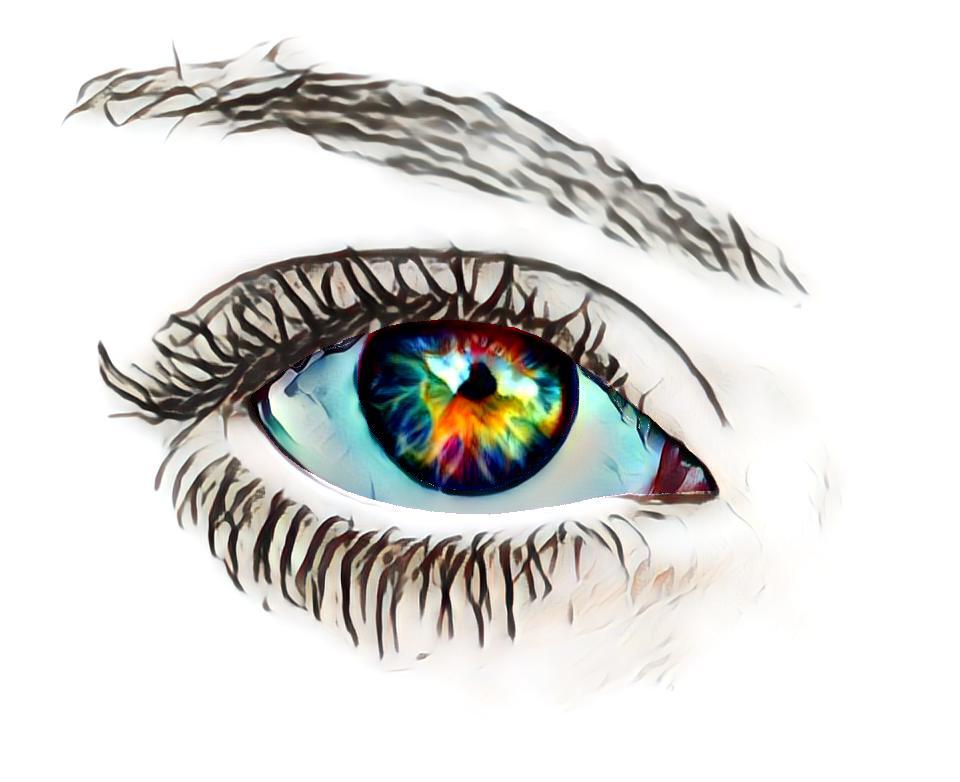} &
\includegraphics[width=.30\linewidth]{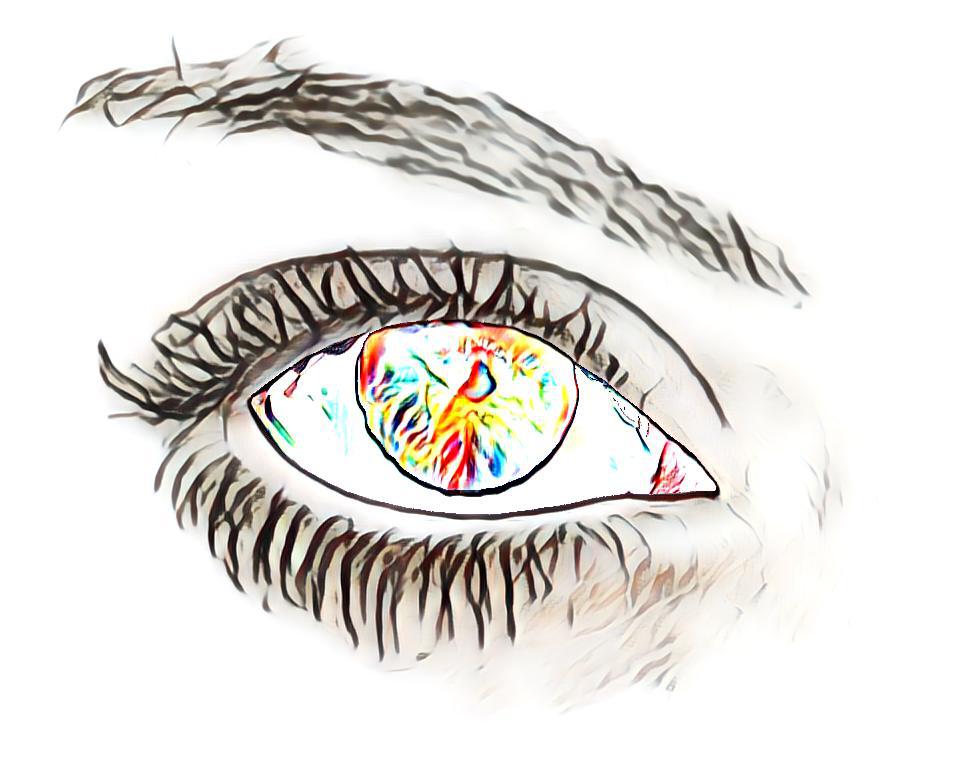} &
\includegraphics[width=.30\linewidth]{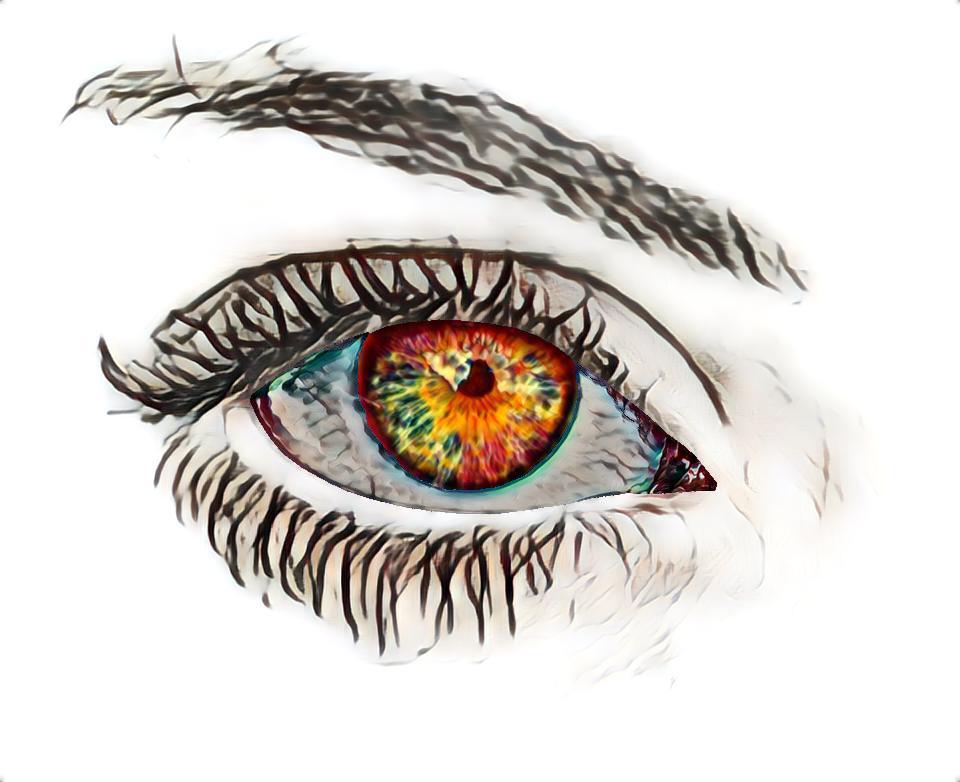}
\\ 
Stylization of entire images & Stylization of pre-masked images & Masked stylization \\
followed by composition & followed by composition & then composition
\end{tabular}
\end{tabular}
\end{center}
   \caption{ Another example using multiple styles and masks.
   The original content image and masked regions are the same as in Fig.~\ref{fig:masked5a}, with a different set of styles applied to the regions.
   Again, masked stylization results in regions that each better represent the full content of the respective source style.
   }
\label{fig:masked5c}
\end{figure}

\noindent
Additionally, in Figs.~\ref{fig:masked5a}--\ref{fig:masked5c}, we provide examples of combining multiple styles by using multiple masks. 
In each of these examples, notice how stylizing the entire original image or pre-masking the background results in the style statistics being applied across the entire image even though the stylization is only intended to be applied to a portion of it.
With our masked stylization, each region more completely reflects the characteristics of the respective source style.

~

\section{Spherical Segmentation}

\begin{figure}
\newcommand{\figwidth}{0.42\linewidth}
\begin{center}
\vspace{0.1in}
\hspace*{-0.1in}
\begin{tabular}{rcc}
\raisebox{0.17\linewidth}{a)}\! 
&
\includegraphics[width=\figwidth]{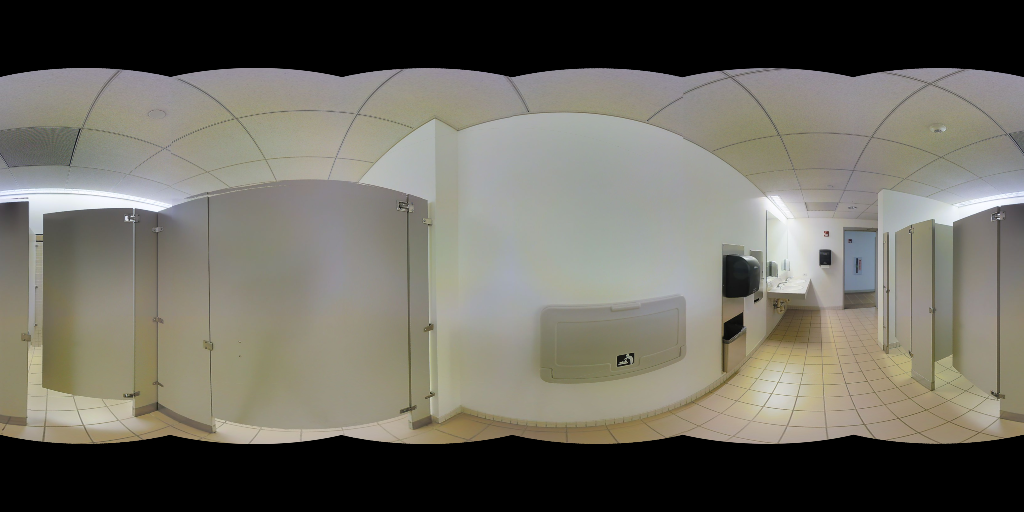} 
&
\includegraphics[width=\figwidth]{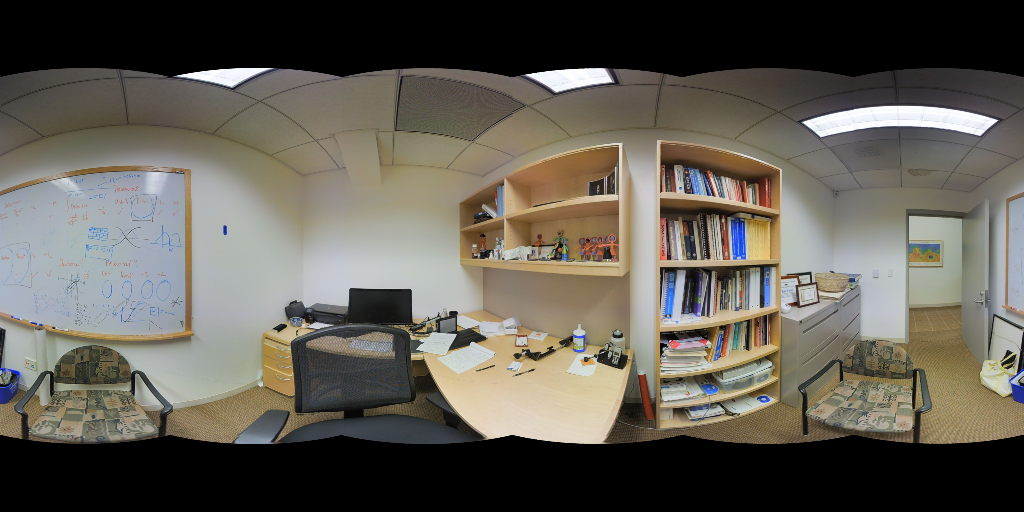} 
\\
\raisebox{0.17\linewidth}{b)}\! 
&
\includegraphics[width=\figwidth]{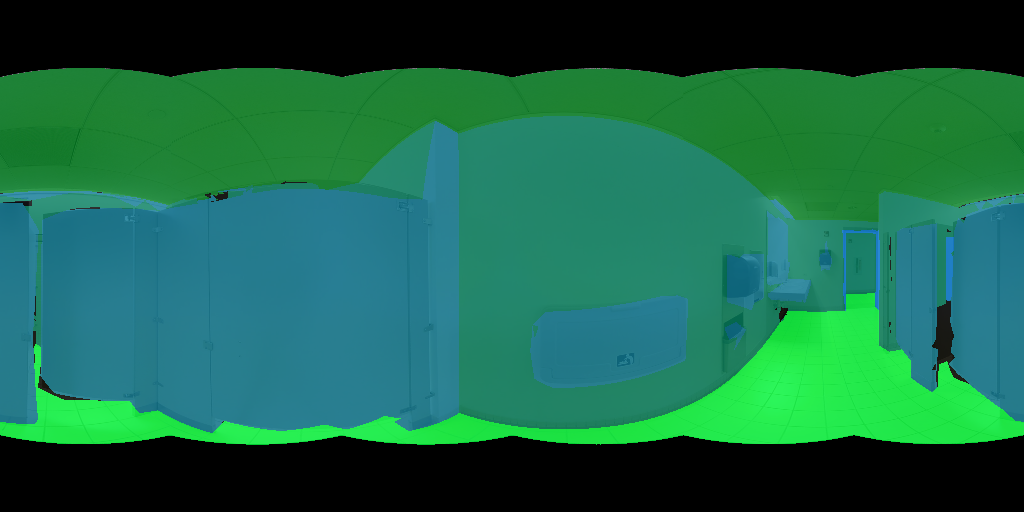} 
&
\includegraphics[width=\figwidth]{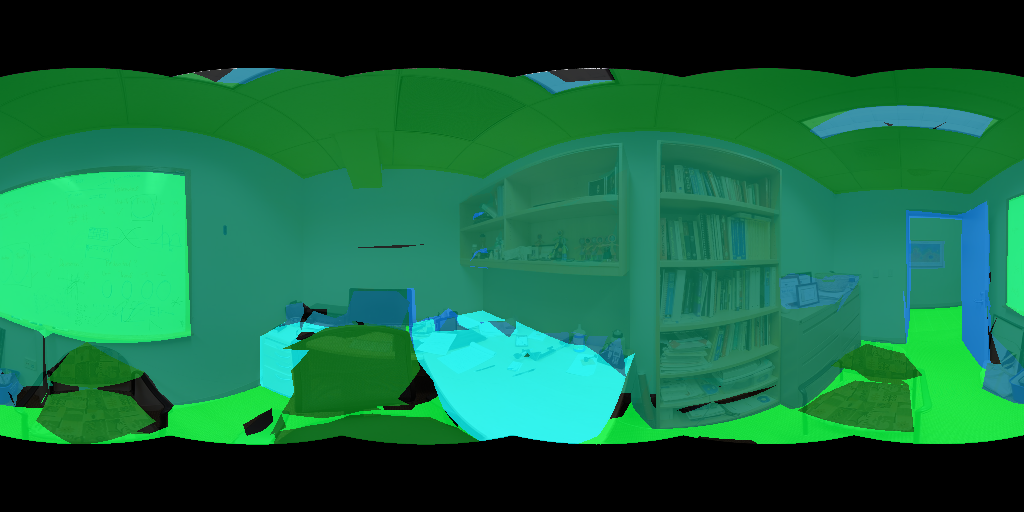} 
\\
\raisebox{0.17\linewidth}{c)}\! 
&
\includegraphics[width=\figwidth]{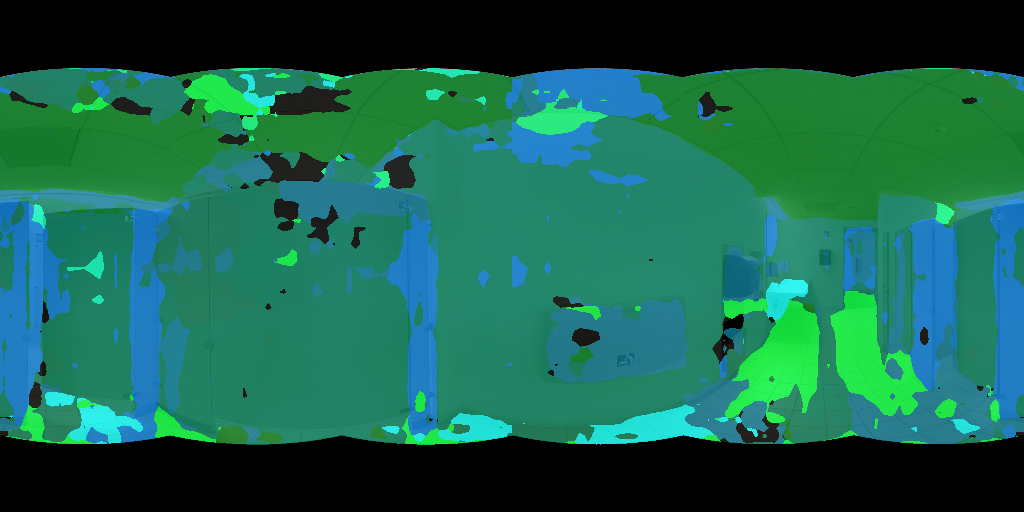} 
&
\includegraphics[width=\figwidth]{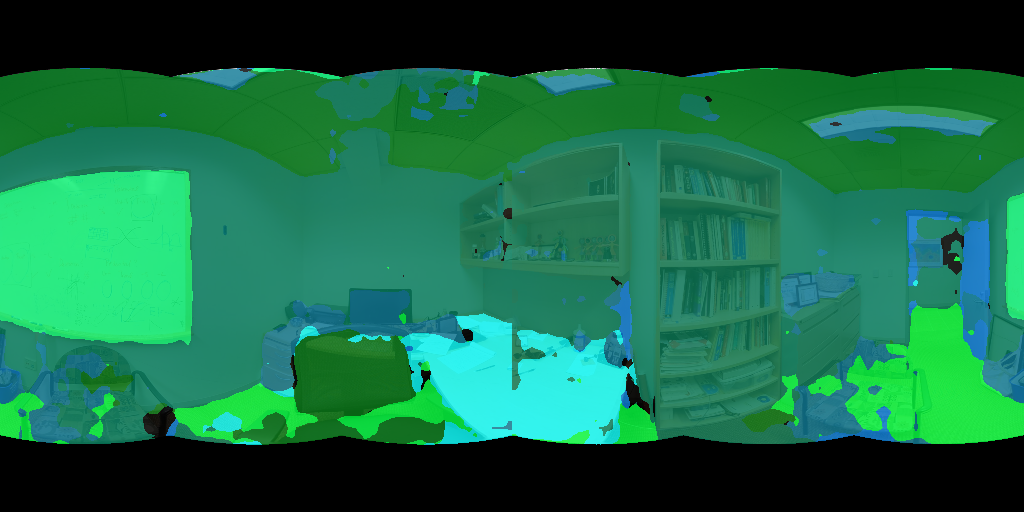} 
\\
\raisebox{0.17\linewidth}{d)}\! 
&
\includegraphics[width=\figwidth]{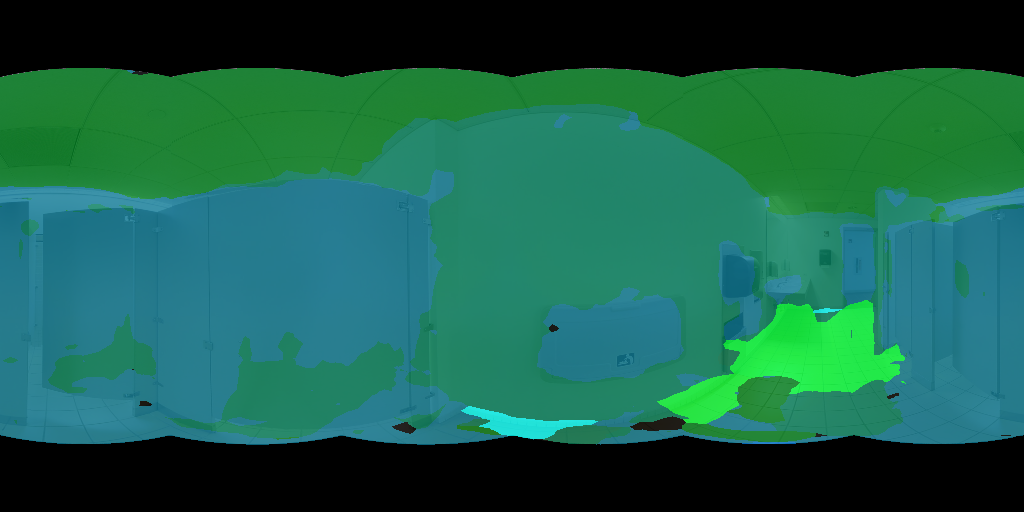} 
&
\includegraphics[width=\figwidth]{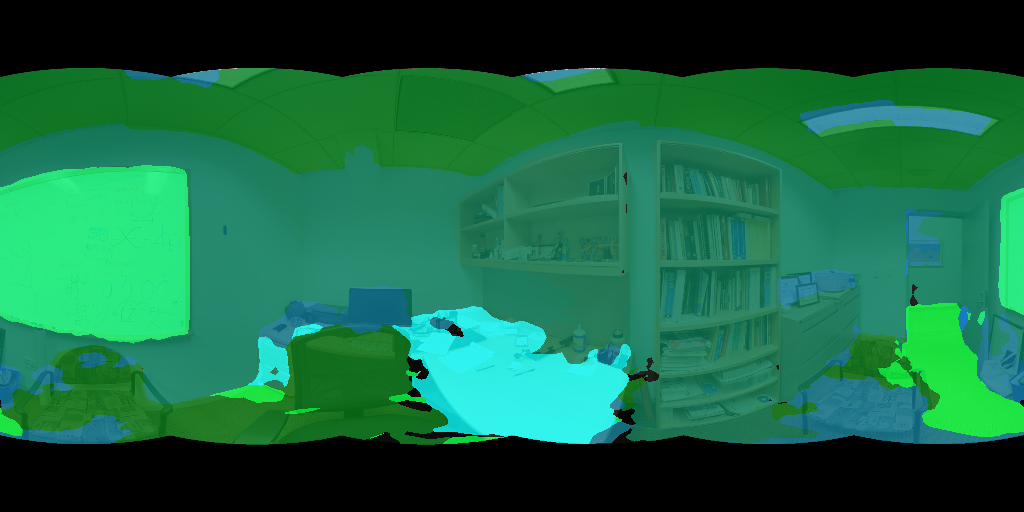} 
\\
\end{tabular}
\vspace*{-0.15in}
\end{center}
   \caption{A visual comparison of semantic segmentation of images from the Stanford 2D-3D-S dataset (a) with ground truth (b) between an FCN with a ResNet-50 backbone using standard convolutions (c) versus our SelectionConv operations (d).
   Note that the use of SelectionConv gives cleaner segmentation results along the poles of and seam of the image (located in the center of this representation).}
   
   
   
\label{fig:segmentation_2d_3ds}
\end{figure}

\begin{figure}
\begin{center}
\includegraphics[width=.80\linewidth]{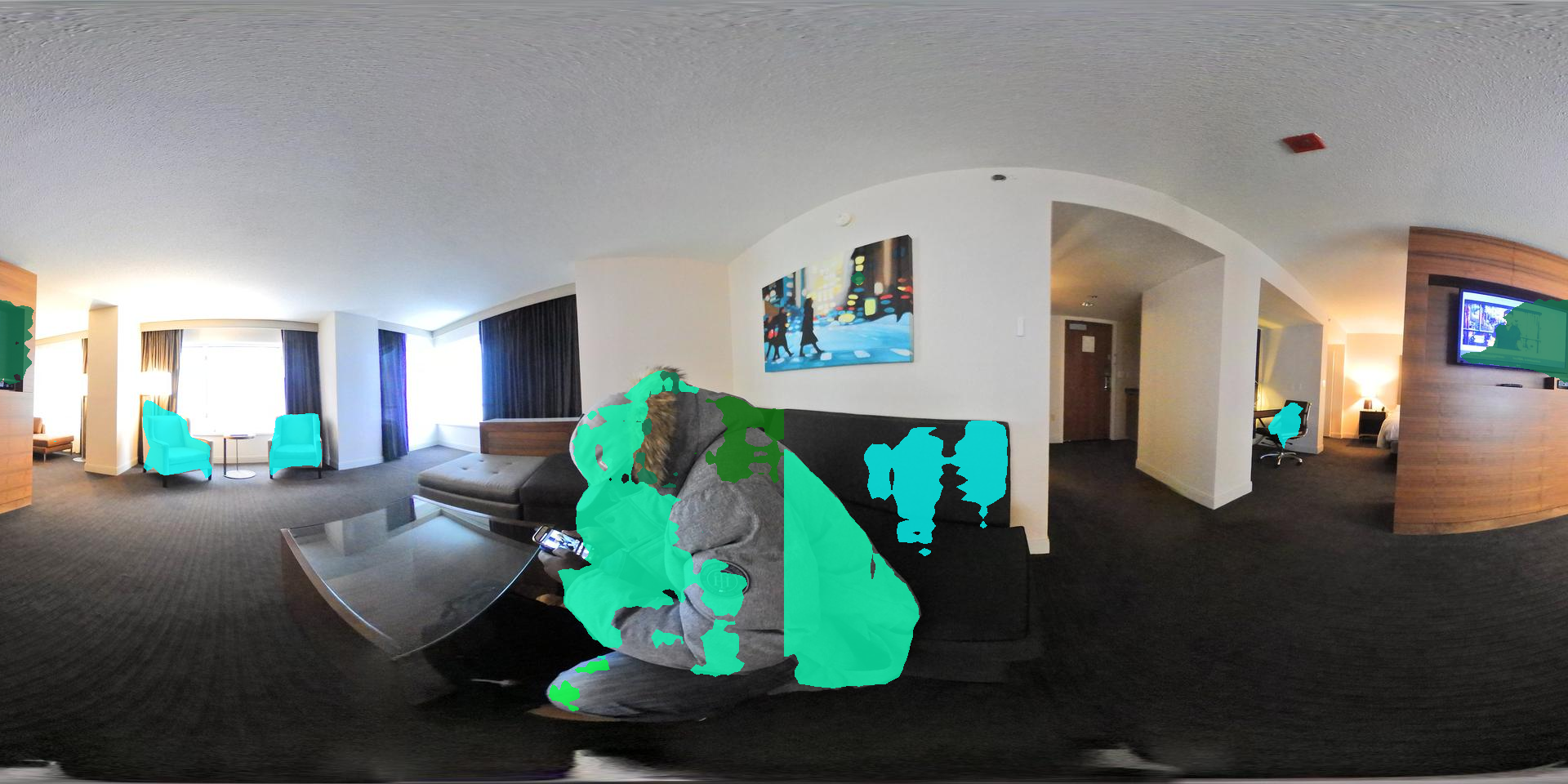}
\\
a) Naive
\\
\vspace{.3in}
\includegraphics[width=.80\linewidth]{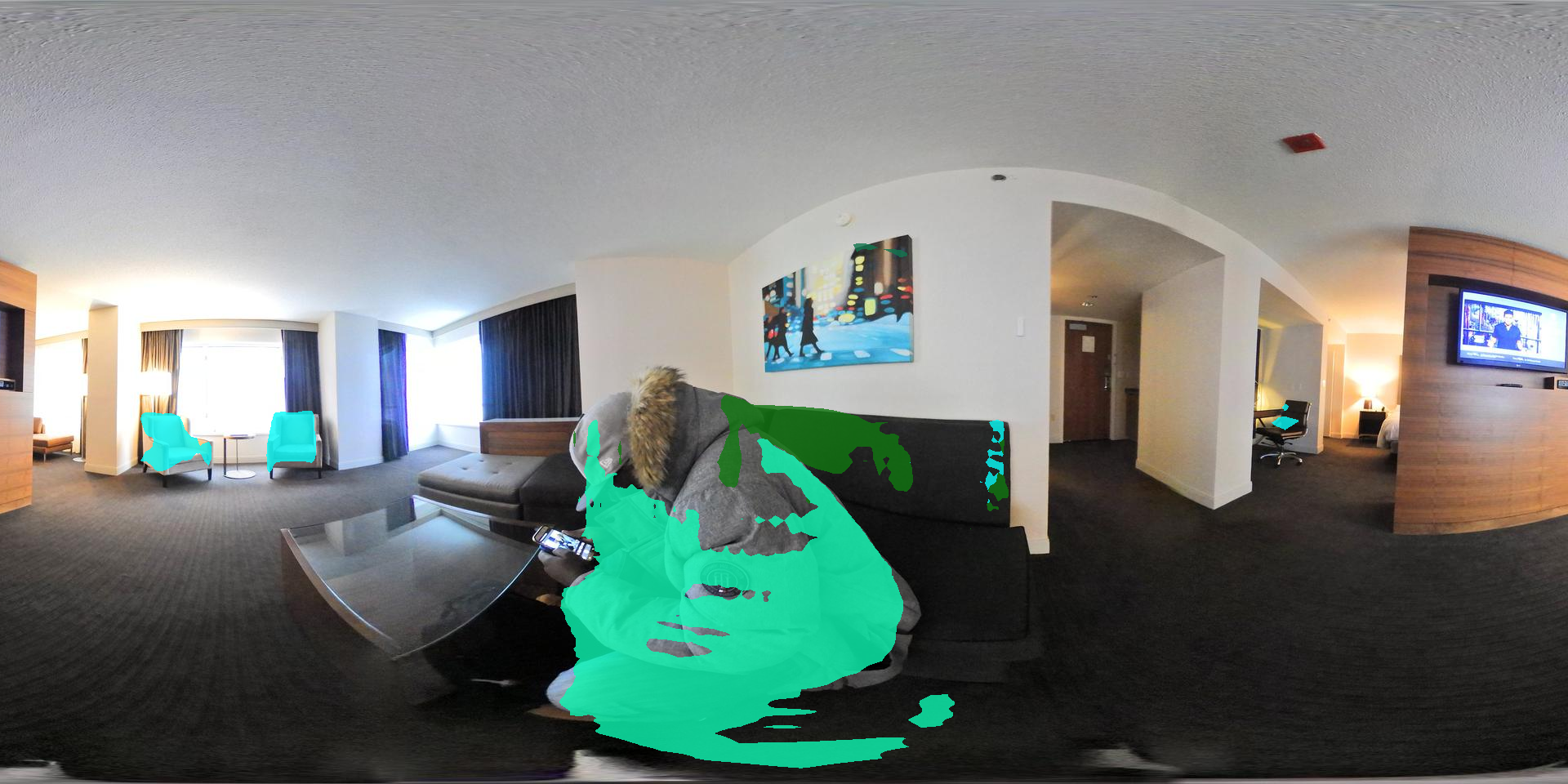} 
\\
b) Ours
\\

\end{center}
   \caption{Segmentation comparison using a ResNet-50 backbone~(a) and our method with transferred weights~(b).  
   Both are circularly rotated by 180 degrees to place the original vertical seam location in the center.
   Note that naive CNN-based segmentation results in a disjoint region for the foreground person while the proposed SelectionConv method allows for more complete selection.
   }
\label{fig:segmentation}
\end{figure}

\noindent
We provide an enlarged version of Figure 7 from the paper here for easier viewing (Fig.~\ref{fig:segmentation_2d_3ds}). We also provide an additional example of using standard FCN pretrained segmentation weights from Pytorch on a spherical image in Fig.~\ref{fig:segmentation}. Note the discontinuous nature of the segmentation along the seam in the naive result compared to our method.


~

\section{Superpixel Depth Prediction}

\begin{figure}[t]
\newcommand{\figwidth}{0.45\linewidth}
\begin{center}
\begin{tabular}{cc}
\includegraphics[width=\figwidth]{Figures/sup_original.png} &
\includegraphics[width=\figwidth]{Figures/sup_reference.png} 
\\
a) Original &
b) Prediction

\\
\includegraphics[width=\figwidth]{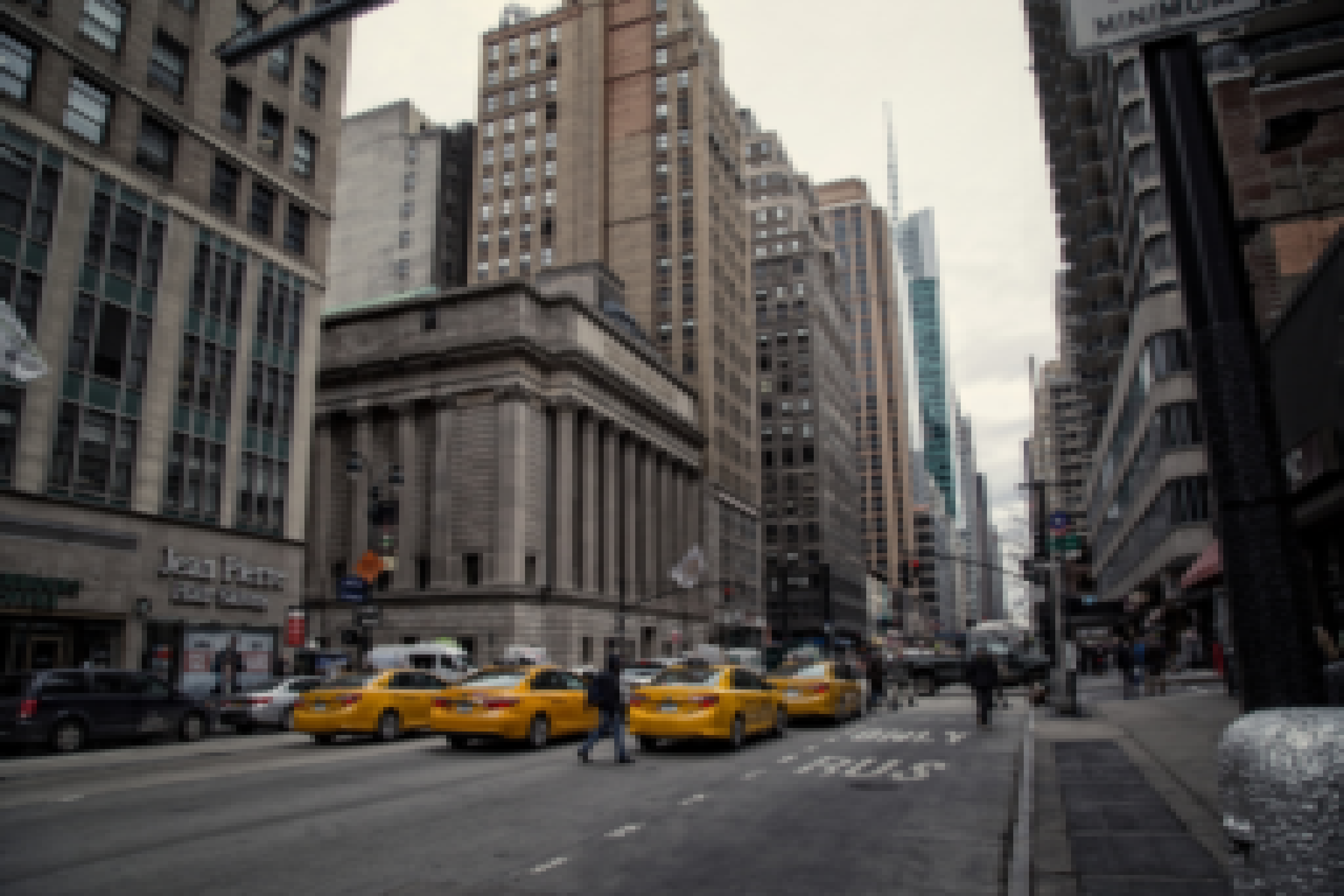} &
\includegraphics[width=\figwidth]{Figures/sup_upscaled.png} 
\\
c) Low-Res &
d) Low-Res Prediction

\\
\includegraphics[width=\figwidth]{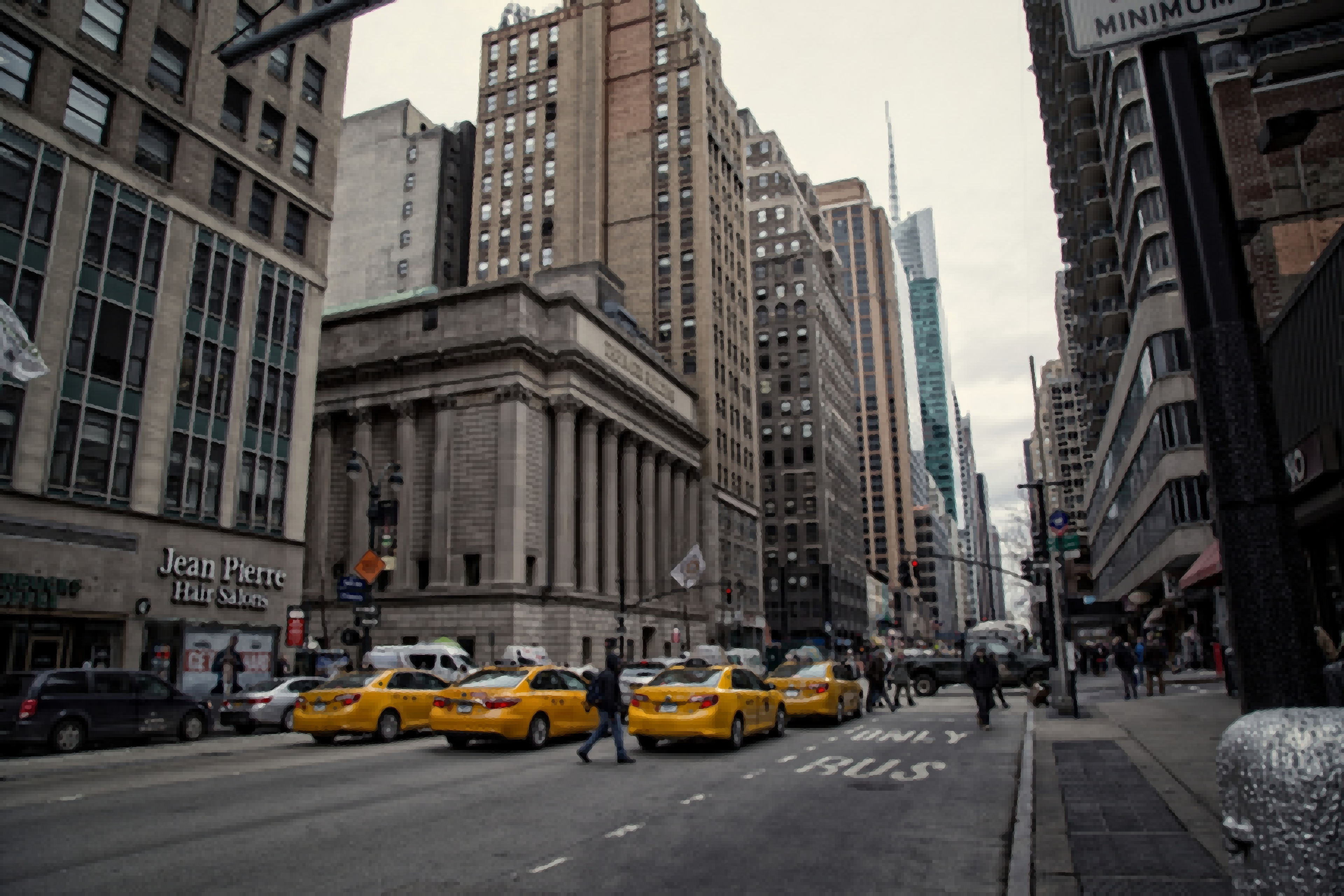} &
\includegraphics[width=\figwidth]{Figures/sup_ours.png} 
\\
e) Superpixel &
f) Superpixel Prediction

\\

\end{tabular}
\end{center}
   \caption{ A high-resolution image (a) requires $\thicksim$25.9 seconds on a CPU to create a predicted depth map (b). A lower-resolution $256 \times 256$ version can be processed (c) by a network in $\thicksim$0.8 seconds on a GPU, but with low-fidelity results when upscaled to the same resolution~(d). Generating approximately the same number of superpixels as the low-resolution image (e) then using our graph-based network requires only $\thicksim$5.1 seconds on a GPU with higher-fidelity results (f).
   \figcheat
   }
\label{fig:superpixel}
\end{figure}

\noindent
We provide an expanded version of Figure 8 from the paper here for easier viewing (Fig.~\ref{fig:superpixel}).

~

\section{Panoramic Stylization}

Though we showed an example of removing seams when stylizing a spherical image in the paper, an even simpler problem is to attempt stylization on a 360$^\circ$ panoramic image. To construct the graph for such a panoramic, edges simply need to be added from the left side of the image to the right side of the image, giving one continuous loop of nodes.

\begin{figure}
\begin{center}
\vspace{0.1in}
\hspace*{-0.2in}
\begin{tabular}{rcc}
\raisebox{0.60in}{a)}\! &
\includegraphics[width=.65\linewidth]{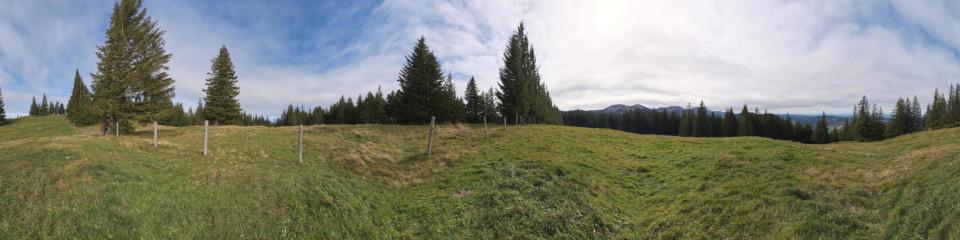} & \includegraphics[width=.10\linewidth]{Figures/style5.jpg}

\\
\raisebox{0.60in}{b)}\! &
\includegraphics[width=.65\linewidth]{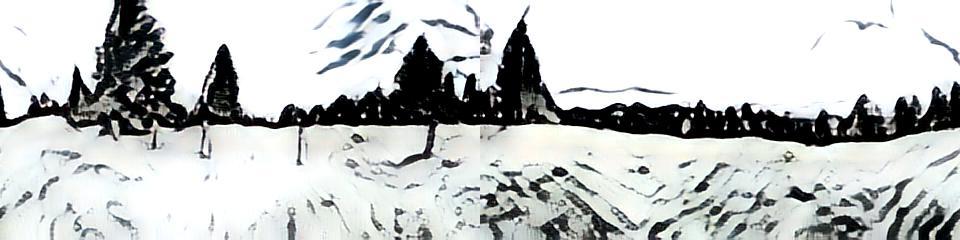} &
\cropim{.15}{Figures/naive_view_style5.jpg}{450}{10}{450}{175}

\\

\raisebox{0.60in}{c)}\! &
\includegraphics[width=.65\linewidth]{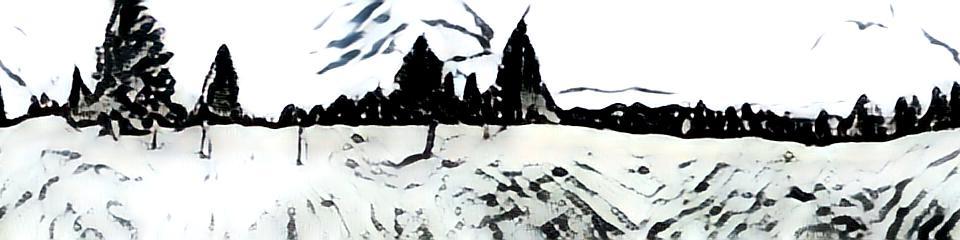} &
\cropim{.15}{Figures/pan_view_style5.jpg}{450}{10}{450}{175}

\end{tabular}
\end{center}
\vspace{-0.1in}
   \caption{A panoramic content image and a given style image~(a). The stylized panoramic image using a traditional CNN~(b) compared to our method~(c). 
   The images are rotated by 180 degrees to place the seam location at the center and a magnified portion is shown at the seam location. 
   }
\label{fig:panoramic1}
\end{figure}

\begin{figure}
\begin{center}
\vspace{0.1in}
\hspace*{-0.2in}
\begin{tabular}{rcc}

\raisebox{0.60in}{a)}\! &
\includegraphics[width=.65\linewidth]{Figures/test360_pan_small_view.jpg} & \includegraphics[width=.15\linewidth]{Figures/style2.jpg}

\\
\raisebox{0.60in}{b)}\! &
\includegraphics[width=.65\linewidth]{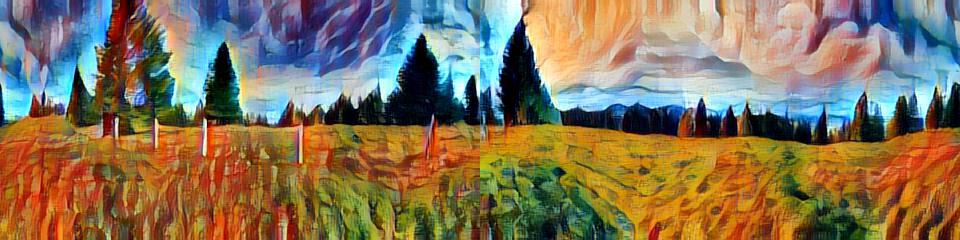} &
\cropim{.15}{Figures/naive_view_style2.jpg}{450}{50}{450}{135}
\\

\raisebox{0.60in}{c)}\! &
\includegraphics[width=.65\linewidth]{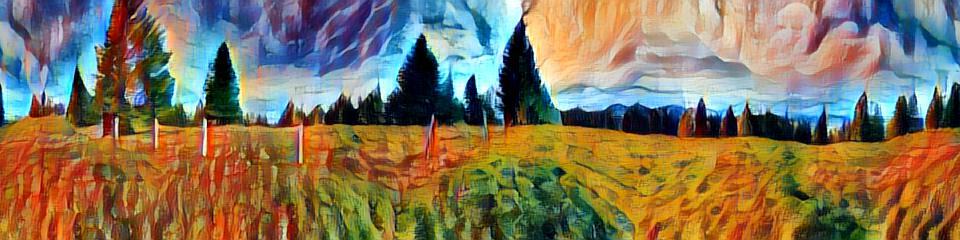} &
\cropim{.15}{Figures/pan_view_style2.jpg}{450}{50}{450}{135}

\\

\end{tabular}
\end{center}
\vspace{-0.1in}
   \caption{Another example, similar to Fig.~\ref{fig:panoramic1}, using a different style image.
   }
\label{fig:panoramic2}
\end{figure}

\begin{figure}
\begin{center}
\vspace{0.1in}
\hspace*{-0.2in}
\begin{tabular}{rcc}

\raisebox{0.50in}{a)}\! &
\includegraphics[width=.65\linewidth]{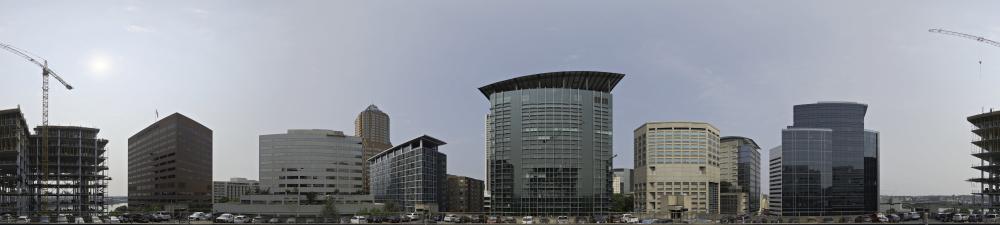} & \includegraphics[width=.15\linewidth]{Figures/style1.jpg}
\\
\raisebox{0.50in}{b)}\! &
\includegraphics[width=.65\linewidth]{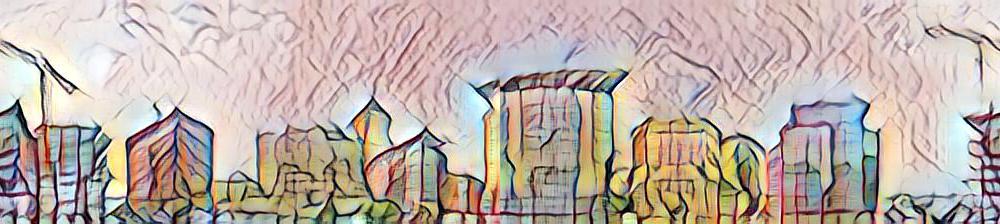} &
\cropim{.15}{Figures/naive_view_style1.jpg}{450}{90}{450}{35}

\\

\raisebox{0.50in}{c)}\! &
\includegraphics[width=.65\linewidth]{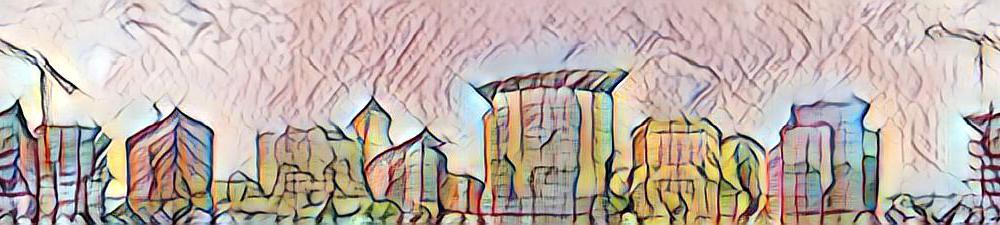} &
\cropim{.15}{Figures/pan_view_style1.jpg}{450}{90}{450}{35}

\end{tabular}
\end{center}
\vspace{-0.1in}
   \caption{Another example, similar to Fig.~\ref{fig:panoramic1}, using different content and style images.
   }
\label{fig:panoramic3}
\end{figure}

\begin{figure}
\begin{center}
\vspace{0.1in}
\hspace*{-0.2in}
\begin{tabular}{rcc}

\raisebox{0.50in}{a)}\! &
\includegraphics[width=.65\linewidth]{Figures/test360_panb_small_view.jpg} & \includegraphics[width=.13\linewidth]{Figures/style6.jpg}
\\
\raisebox{0.50in}{b)}\! &
\includegraphics[width=.65\linewidth]{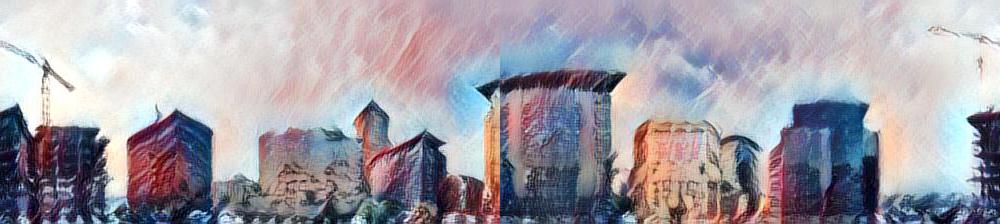} &
\cropim{.15}{Figures/naive_view_style6.jpg}{450}{90}{450}{35}

\\

\raisebox{0.50in}{c)}\! &
\includegraphics[width=.65\linewidth]{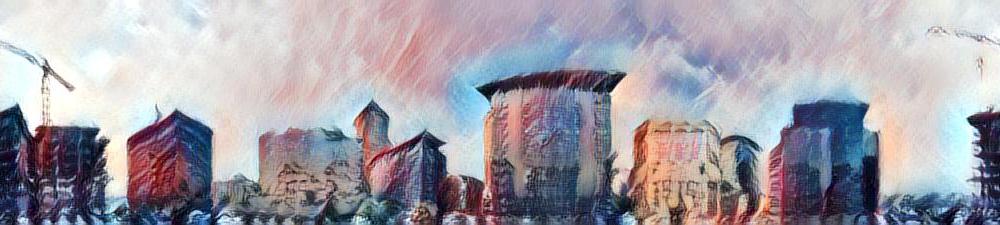} &
\cropim{.15}{Figures/pan_view_style6.jpg}{450}{90}{450}{35}

\end{tabular}
\end{center}
\vspace{-0.1in}
   \caption{Another example, similar to Fig.~\ref{fig:panoramic3}, using a different style image.
   }
\label{fig:panoramic4}
\end{figure}

The results of stylization on panoramic images is demonstrated in Fig.~\ref{fig:panoramic1} through Fig.~\ref{fig:panoramic4}.
Just like with spherical stylization, naive stylization of panoramic images results in a noticeable seam where the image wraps around horizontally.
Our graph-based approach, transferring from a pre-trained image-based network, avoids such issues.

~

\section{Spherical and Texture Map Stylization}

\begin{figure}[p]
\begin{center}
\begin{tabular}{c}
\begin{tabular}{cc}
\includegraphics[width=.35\linewidth]{Figures/test360b_withstyle.jpg} &
\hspace*{.2in}
\includegraphics[width=.35\linewidth]{Figures/cube_mapstyle0.jpg} 
\\ 
a) Spherical image map and style  & 
\hspace*{.2in}
b) Stylization with our method
\end{tabular}
\\ \\
\begin{tabular}{ccc}
\cropim{.28}{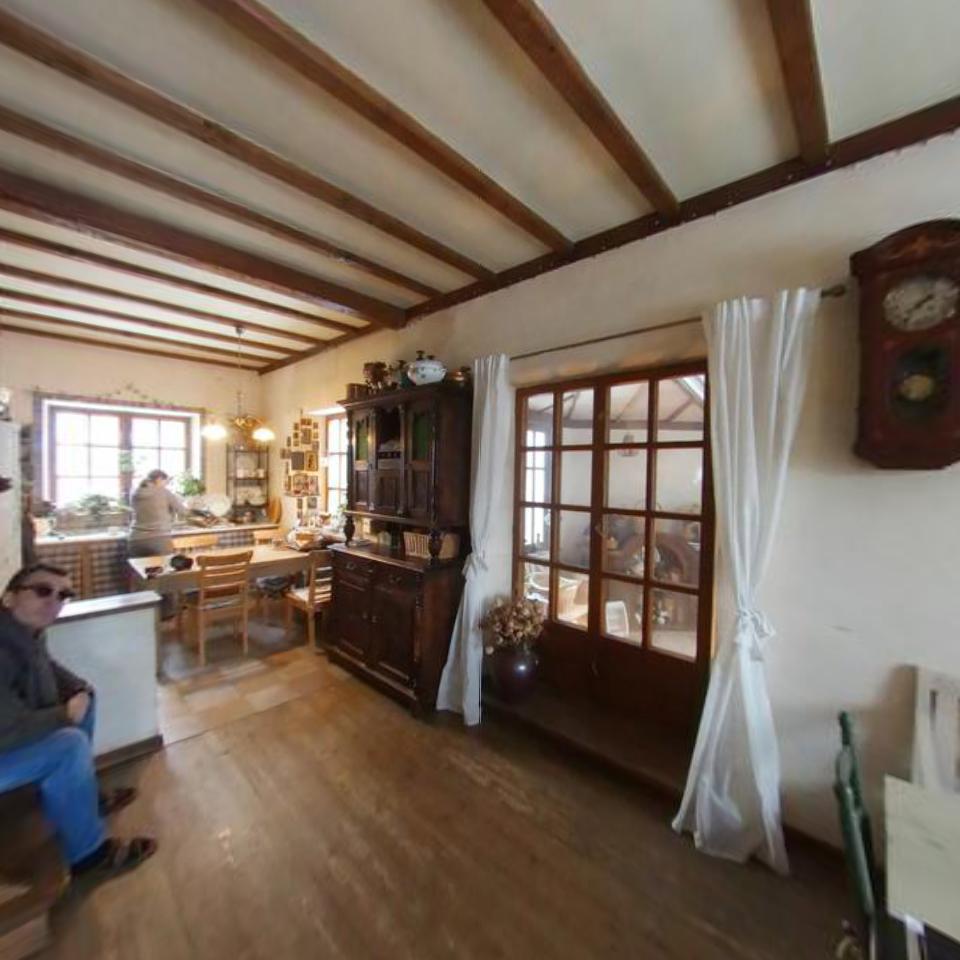}{150}{150}{150}{150} &
\cropim{.28}{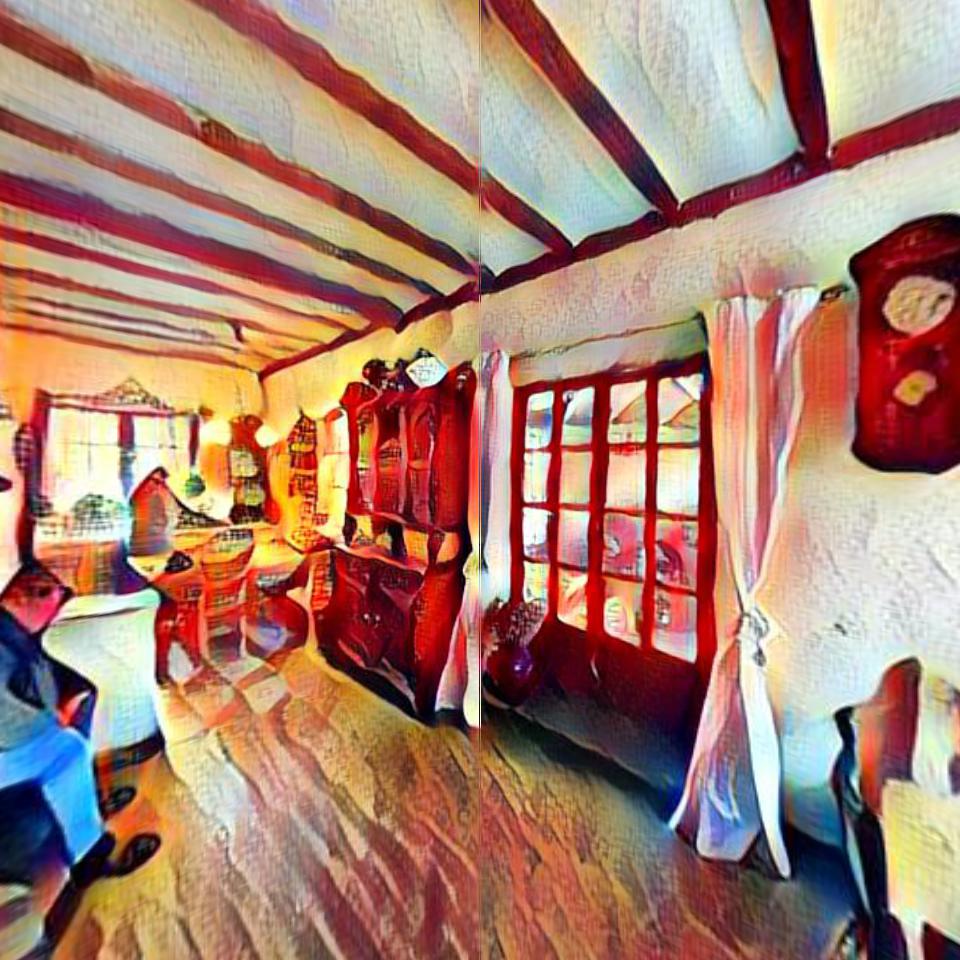}{150}{150}{150}{150} &
\cropim{.28}{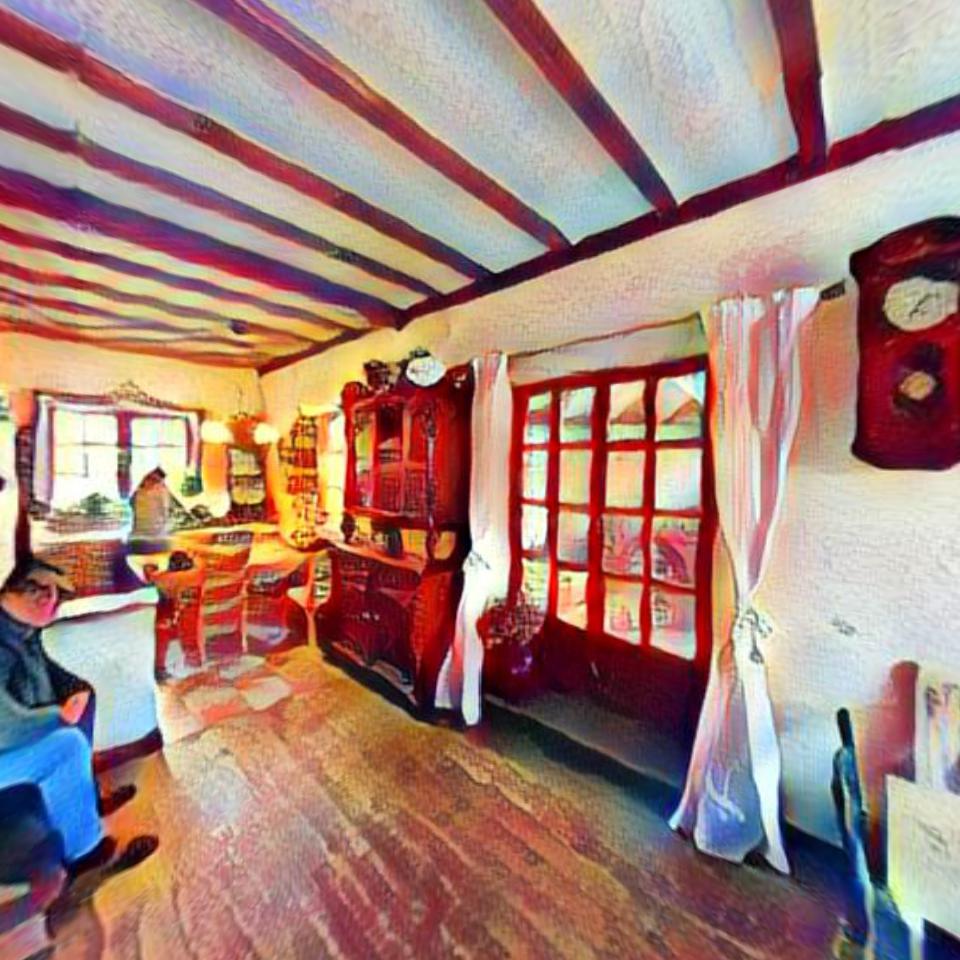}{150}{150}{150}{150}
\\
c) Horizontal view &
d) Naive stylization &
e) Our method
\\ \\
\cropim{.28}{Figures/test360b_groundview.jpg}{150}{150}{150}{150} &
\cropim{.28}{Figures/naive_groundview_style0.jpg}{150}{150}{150}{150} &
\cropim{.28}{Figures/cubemap_groundview_style0.jpg}{150}{150}{150}{150}
\\
f) Polar view (looking down) &
g) Naive stylization &
h) Our method
\end{tabular}
\end{tabular}
\end{center}
\vspace{-0.15in}
   \caption{A 360$^\circ$ image (a) and its stylization using our feed-forward method (b). Example views (c,f) have seams and distortion due to discontinuities when the original image map is stylized naively (d,g), but those seams and distortion are minimized with our method (e,h).
   A sequence panning around the sphere is shown in our supplementary video.
   }
\label{fig:spherestyle}
\end{figure}

\begin{figure}
\begin{center}
\begin{tabular}{c}
\includegraphics[width=.70\linewidth]{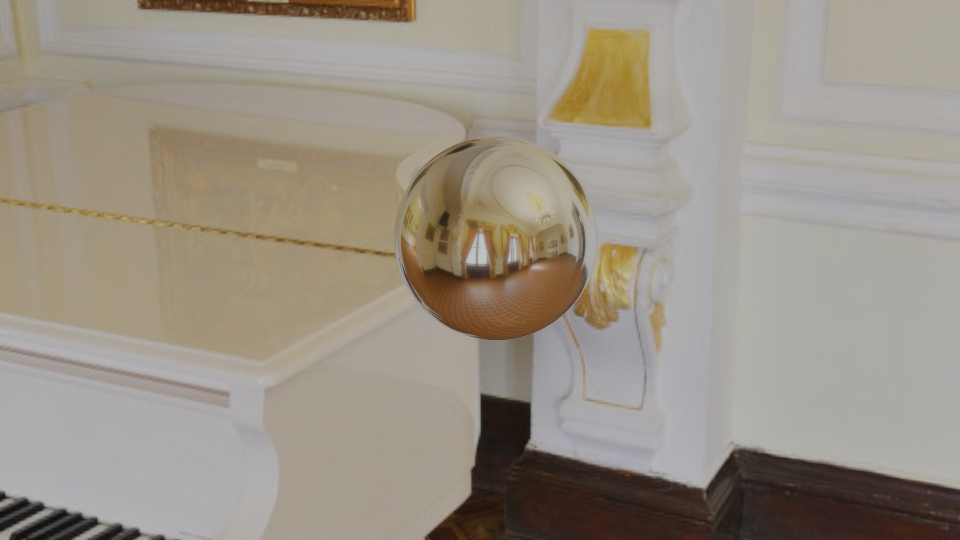} \\
a) Original rendering
\\
\\
\includegraphics[width=.70\linewidth]{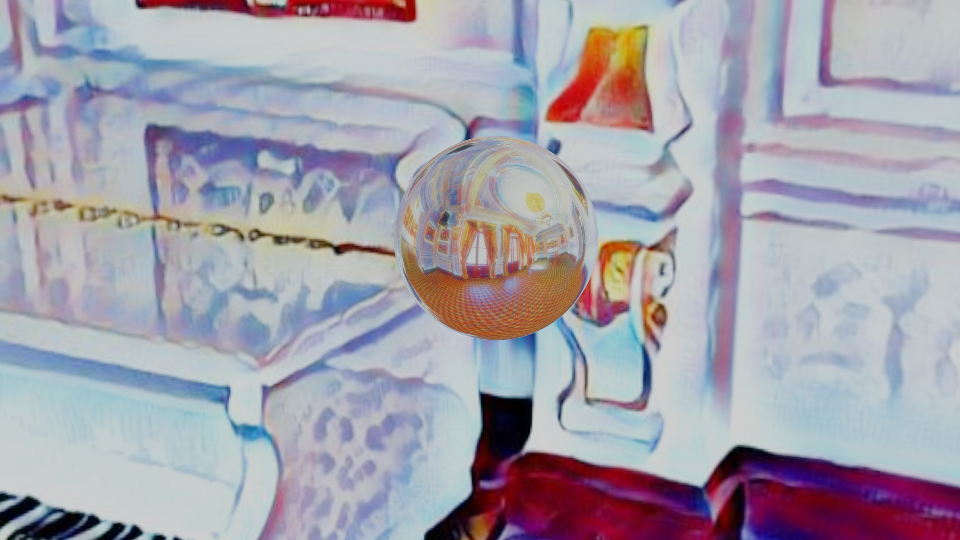} \\
b) Stylized environment map (with reflective sphere)
\\
\end{tabular}
\end{center}
   \caption{A rendering of a mirrored sphere with a high resolution environment map (a) and a rendering with a stylized environment map using our method (b). Note that there is no direct stylization of the reflective sphere. A visualization of this scene is shown in our supplementary video.
   }
\label{fig:environment}
\end{figure}

\begin{figure}
\begin{center}
\begin{tabular}{ccc}

\includegraphics[width=.25\linewidth]{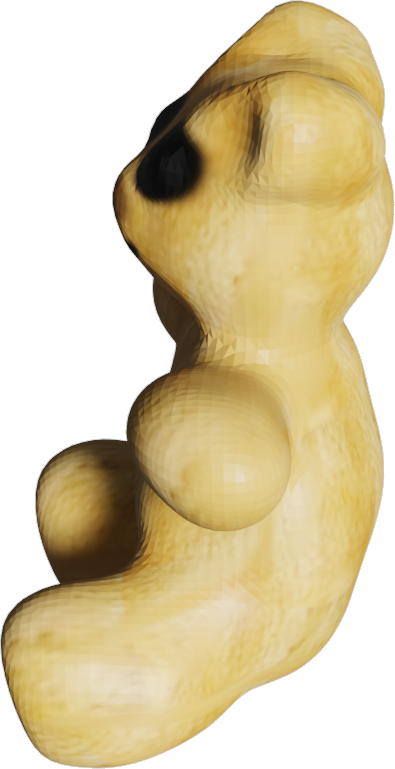} &
\includegraphics[width=.25\linewidth]{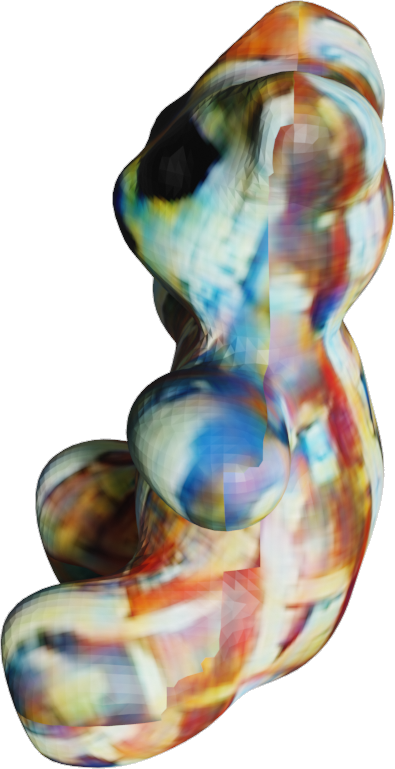} &
\includegraphics[width=.25\linewidth]{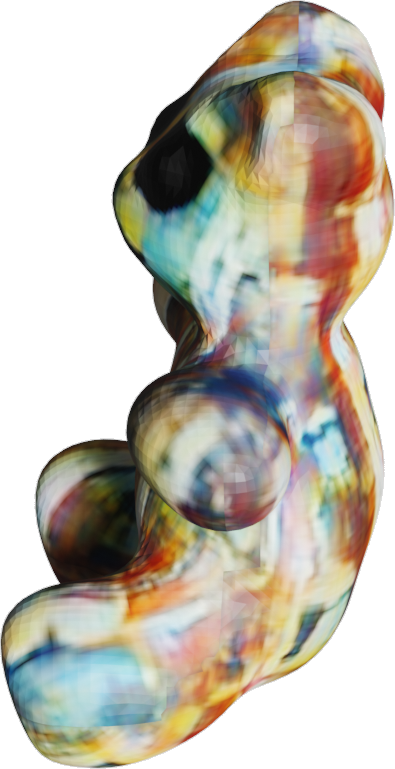} 
\\
a) 3D Mesh & 
b) Naive &
c) Our Method
\\
\\
\includegraphics[width=.25\linewidth]{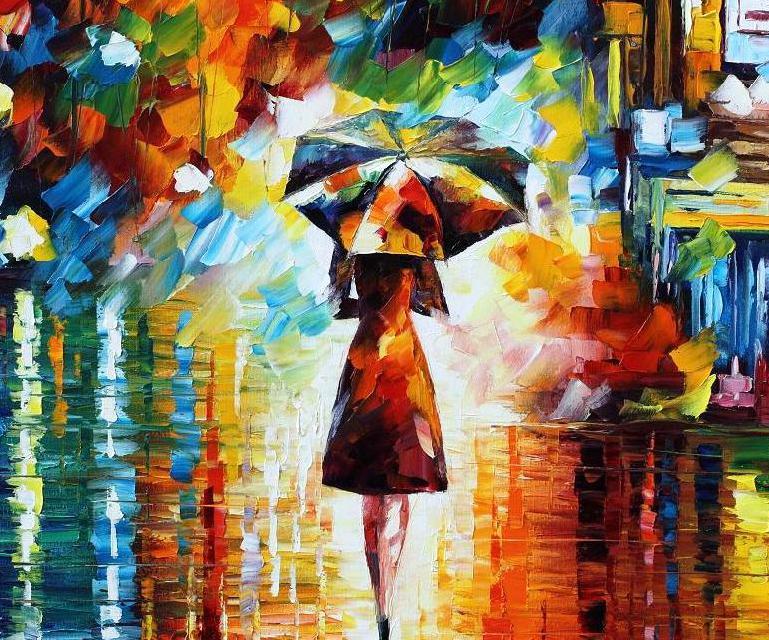} &
\cropim{.25}{Figures/teddy2naive.png}{170}{400}{25}{50} &
\cropim{.25}{Figures/teddy2ours.png}{170}{400}{25}{50} \\
d) Style image & 
e) Close-up (Naive) &
f) Close-up (Ours) 




\end{tabular}
\vspace{-0.13in}
\end{center}
   \caption{ 
        3D mesh~(a) and a style image (d), the result of naively stylizing the texture map~(b) and a magnification (e), and the result of using our method~(c) and a magnification (f). Note the visible seams shown in the magnifications of the naive method (e), whereas our method in~(f) minimizes the visibility of those seams.
   }
\label{fig:meshstyles-a}
\end{figure}

\begin{figure}
\begin{center}
\begin{tabular}{ccc}

\includegraphics[width=.30\linewidth]{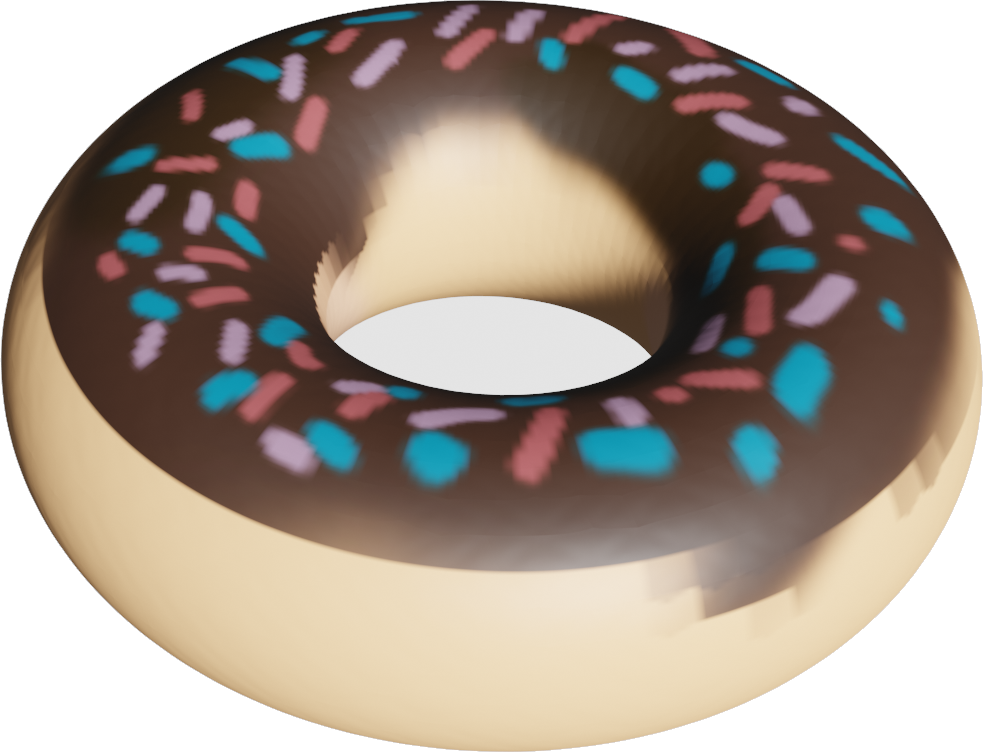} &
\includegraphics[width=.30\linewidth]{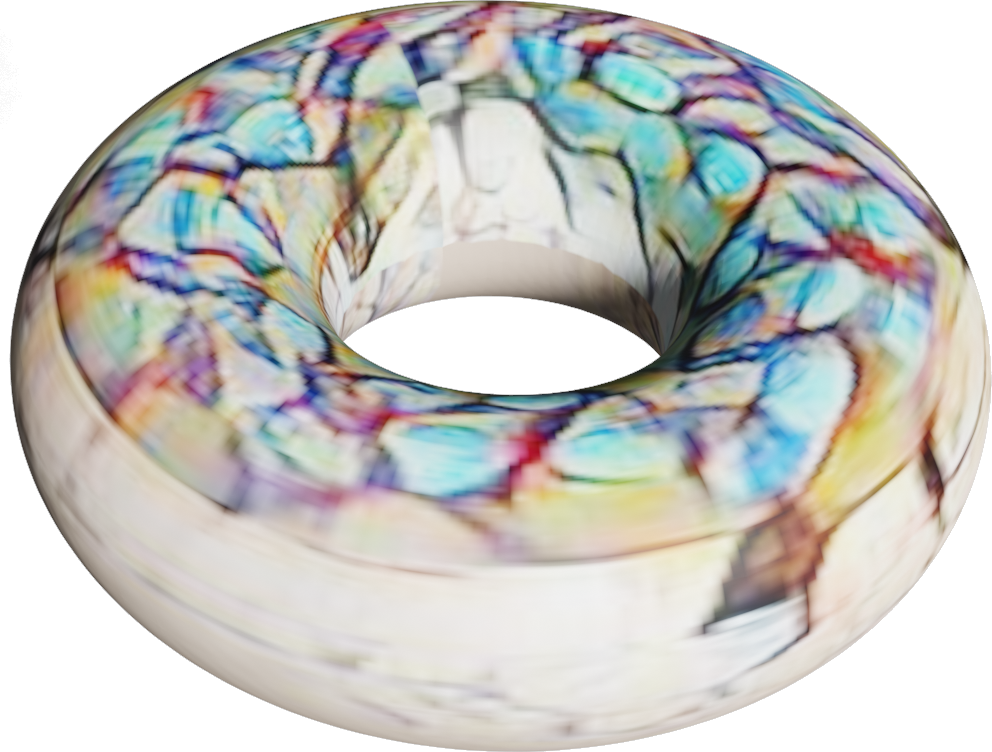} &
\includegraphics[width=.30\linewidth]{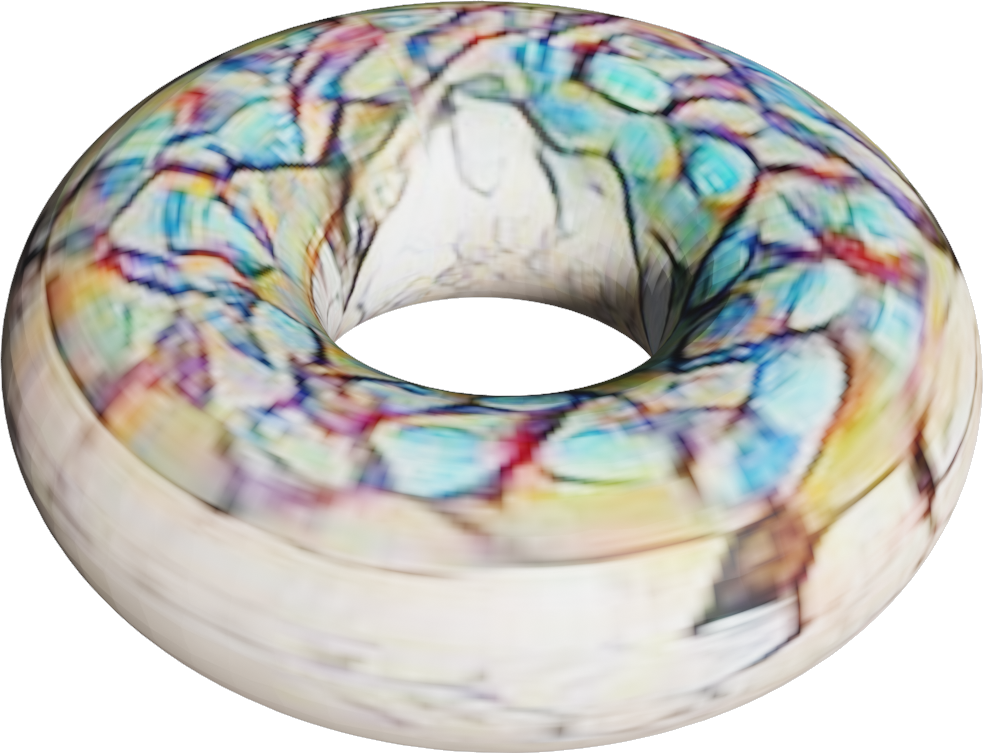} \\
a) Texture mapped model & 
b) Naively stylized &
c) Our method 
\\ \\
\includegraphics[width=.25\linewidth]{Figures/style1.jpg} &
\cropim{.28}{Figures/donut1naive_v2.png}{275}{450}{375}{50} &
\cropim{.273}{Figures/donut1ours_v2.png}{275}{450}{375}{50} \\
d) Style image & 
e) Magnification (Naive) &
f) Magnification (Ours) 
\\ \\

\end{tabular}
\vspace{-0.13in}
\end{center}
   \caption{ 
        Example of texture map stylization.
        When a texture mapped model (a) is stylized (d) by naively stylizing the entire texture map (b), visible seams result because the stylization ignores discontinuities inherent in texture mapping.
        Stylizing the texture map using a graph-based approach that respects these discontinuities removes these seams (c).
        Magnifications of (b) and (c) can be seen in (e) and (f) respectively.
        Note the clear seams in (e) for the seam running around the inner ring of the torus and around the torus from the inside to the outside. 
        While typical artifacts of texture mapping can still be seen in (f), there are not the same discontinuities in the applied style.
   }
\label{fig:meshstyles1}
\end{figure}

\begin{figure}
\begin{center}

\includegraphics[width=.20\linewidth]{Figures/style4.jpg}
\\
a) Style Image
\\
\vspace{.25in}

\begin{tabular}{ccc}

\includegraphics[width=.25\linewidth]{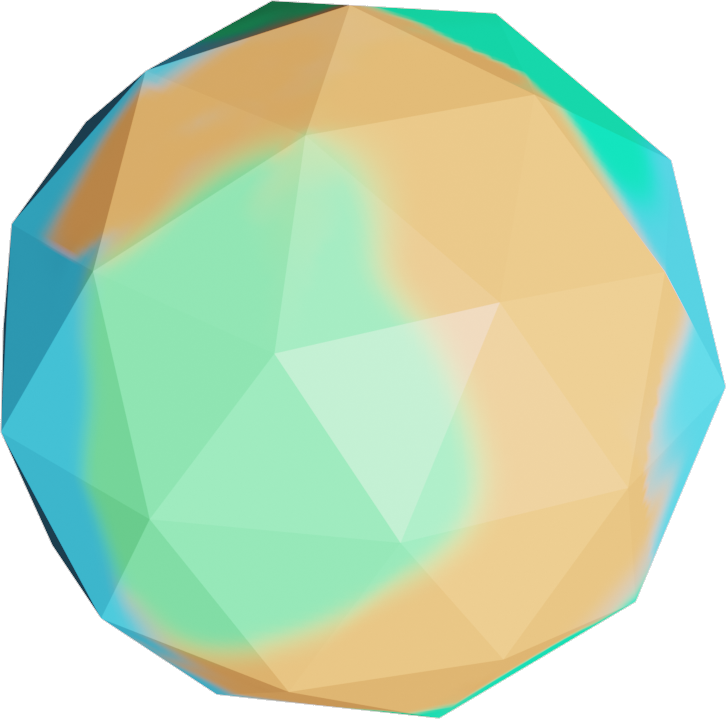} &
\includegraphics[width=.25\linewidth]{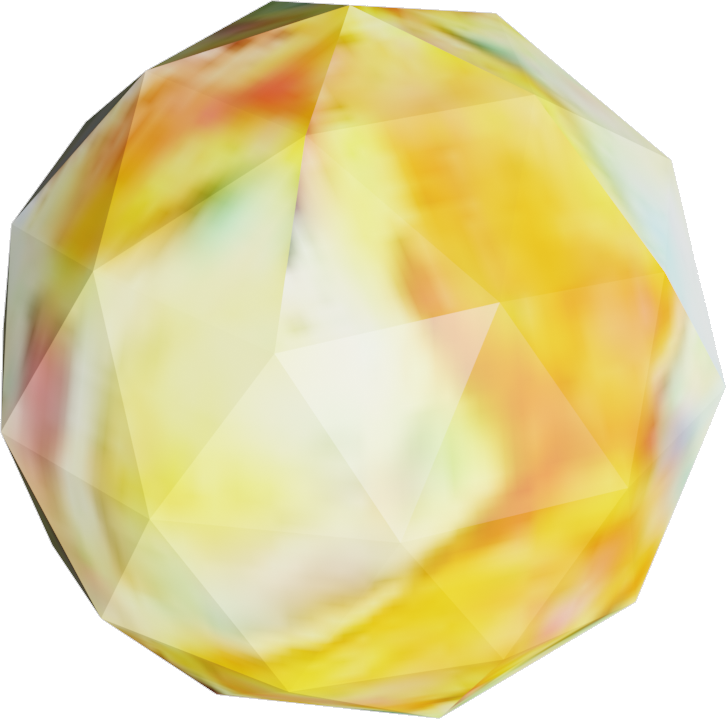} &
\includegraphics[width=.25\linewidth]{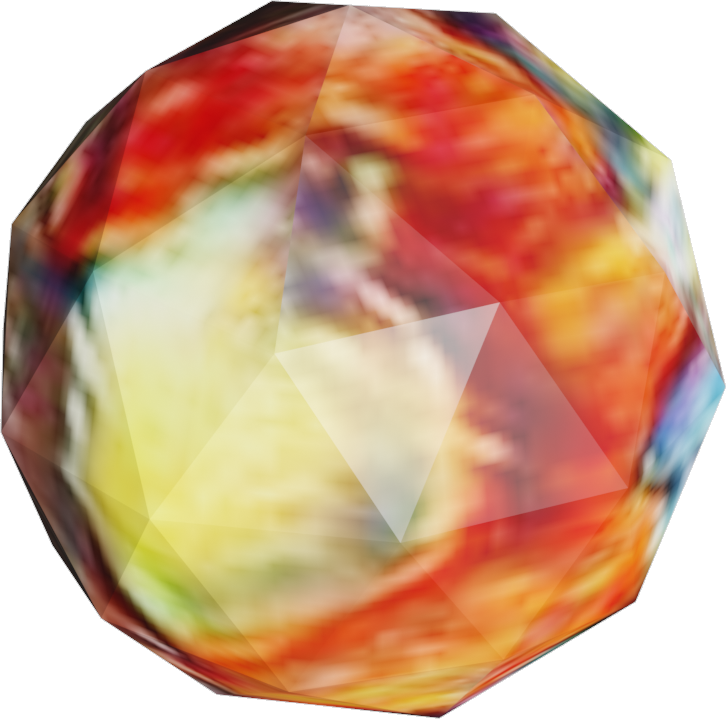} \\
b) Texture mapped model & c) Naively stylized model & d) Our method
\\
\\

\includegraphics[width=.25\linewidth]{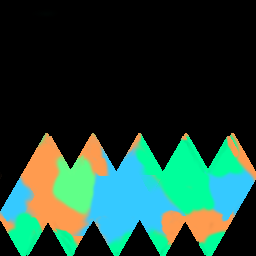} &
\includegraphics[width=.25\linewidth]{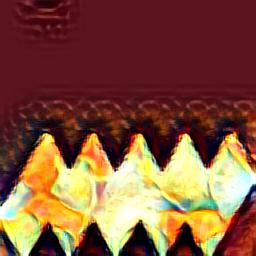} &
\includegraphics[width=.25\linewidth]{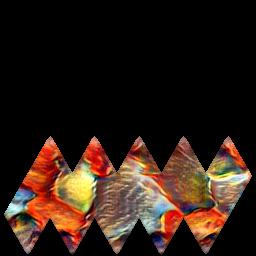} \\
e) Original texture map &
f) Naively stylized texture map &
g) Our method

\end{tabular}
\vspace{-0.13in}
\end{center}
   \caption{ 
        A given style image (a), an original 3D mesh (b) and its texture map~(e), the result of naively stylizing the texture map~(c,f), and the result of using our method~(d,g). 
        Note that the large amount of unused space on the texture map has a significant effect on the naive approach (e), mapping the darker reds of the style image to the black (unused) part of the texture map and the lighter portions to the colored (used) part of the map, resulting in a stylized model that draws primarily from the lighter colors of the style.
        Our approach (f), by stylizing only the used portion of the texture (similar to Fig.~\ref{fig:masked1}), more accurately captures the style within the region of interest, resulting in more complete use of the source style.
   }
\label{fig:meshstyles2}
\end{figure}

\begin{figure}
\begin{center}

\includegraphics[width=.20\linewidth]{Figures/style0.jpg}
\\
a) Style Image
\\

\vspace{.25in}

\begin{tabular}{ccc}
\includegraphics[width=.20\linewidth]{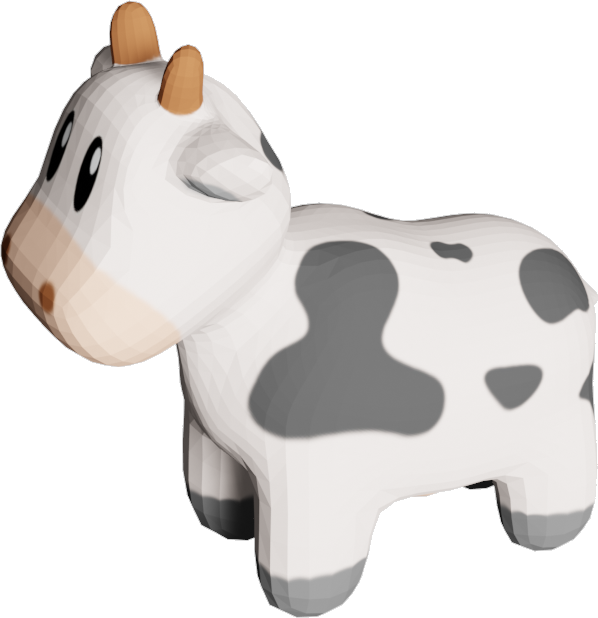} &
\includegraphics[width=.20\linewidth]{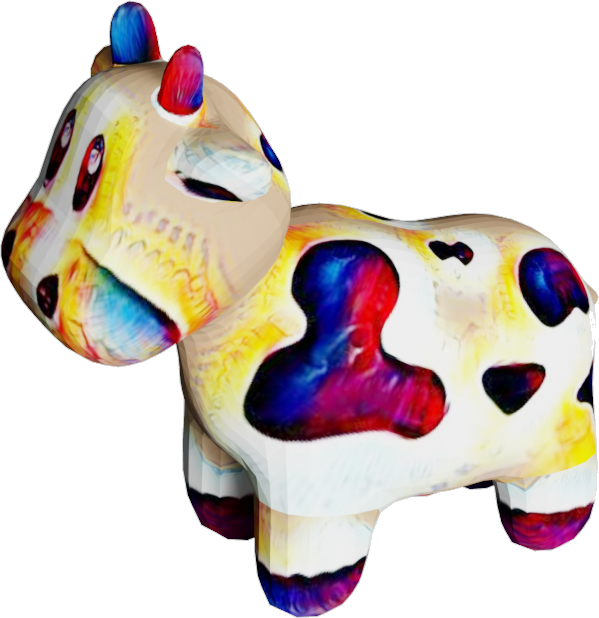} &
\includegraphics[width=.20\linewidth]{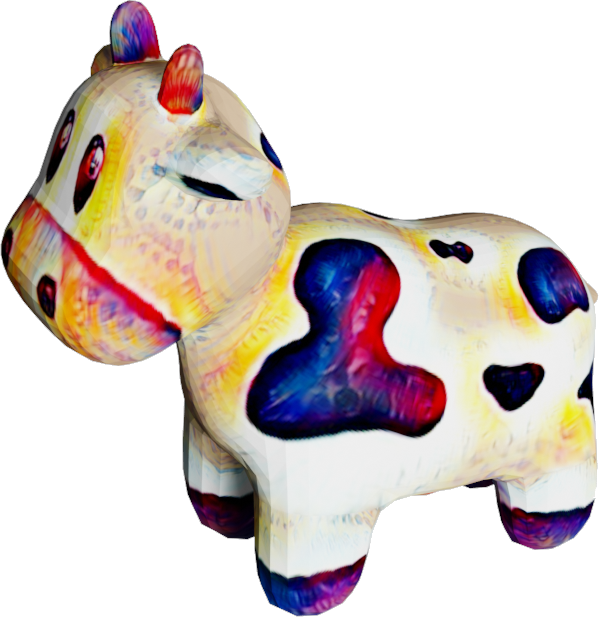} \\

b) Texture mapped model & c) Naively stylized model & d) Our method
\\
\\

\includegraphics[width=.20\linewidth]{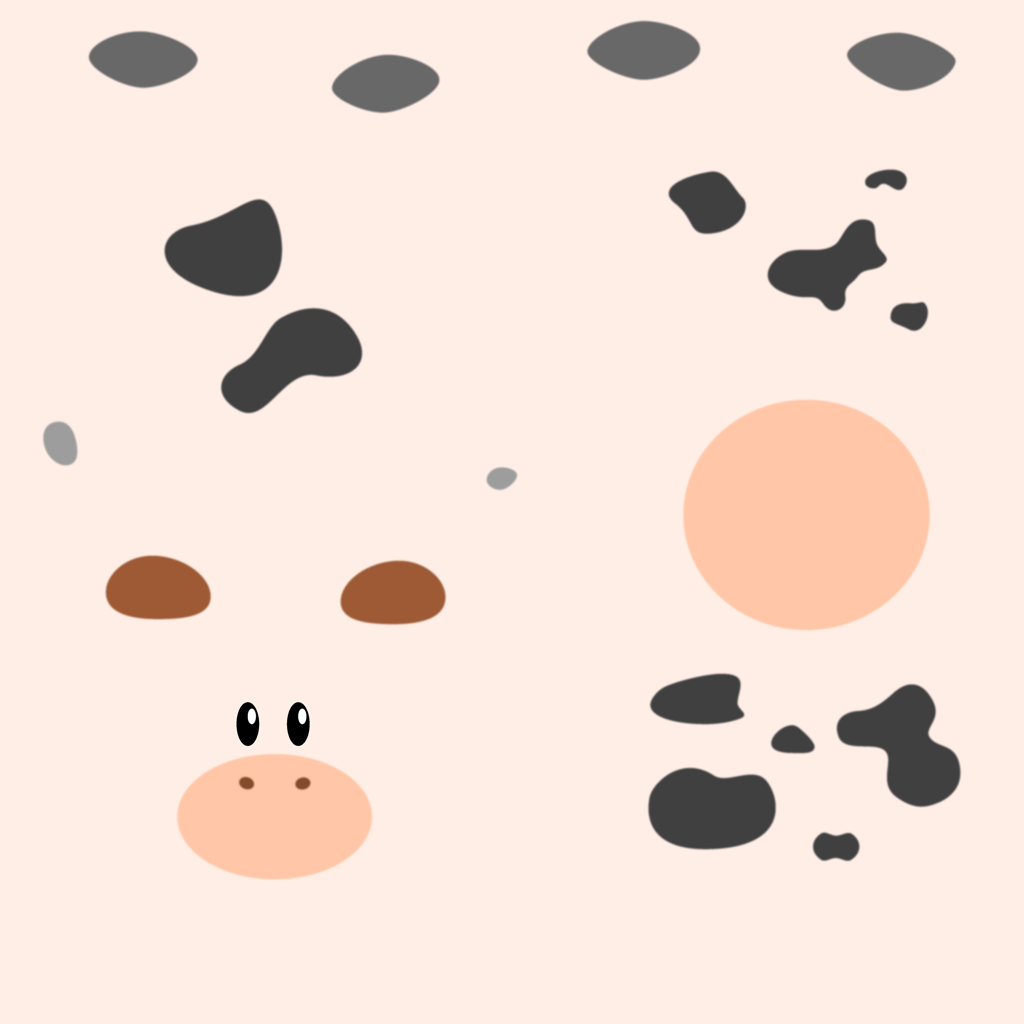} &
\includegraphics[width=.20\linewidth]{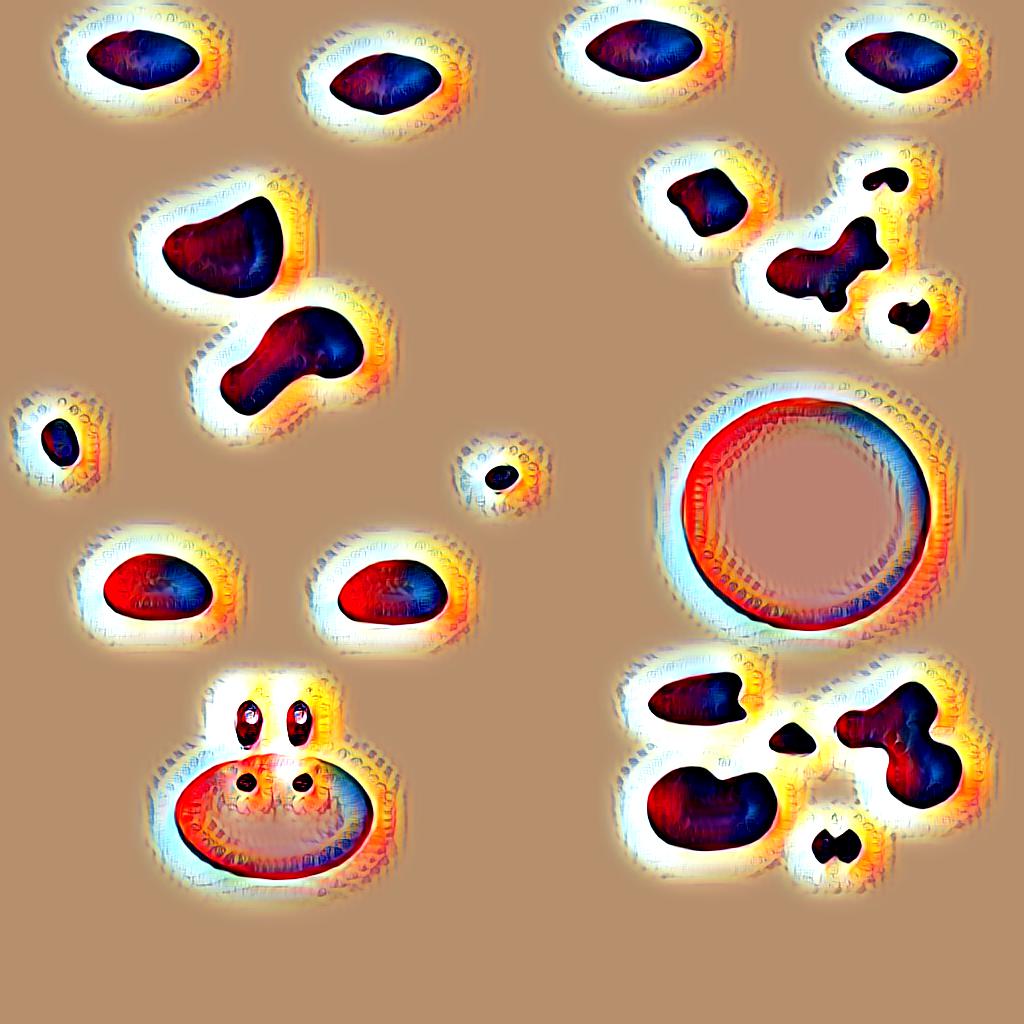} &
\includegraphics[width=.20\linewidth]{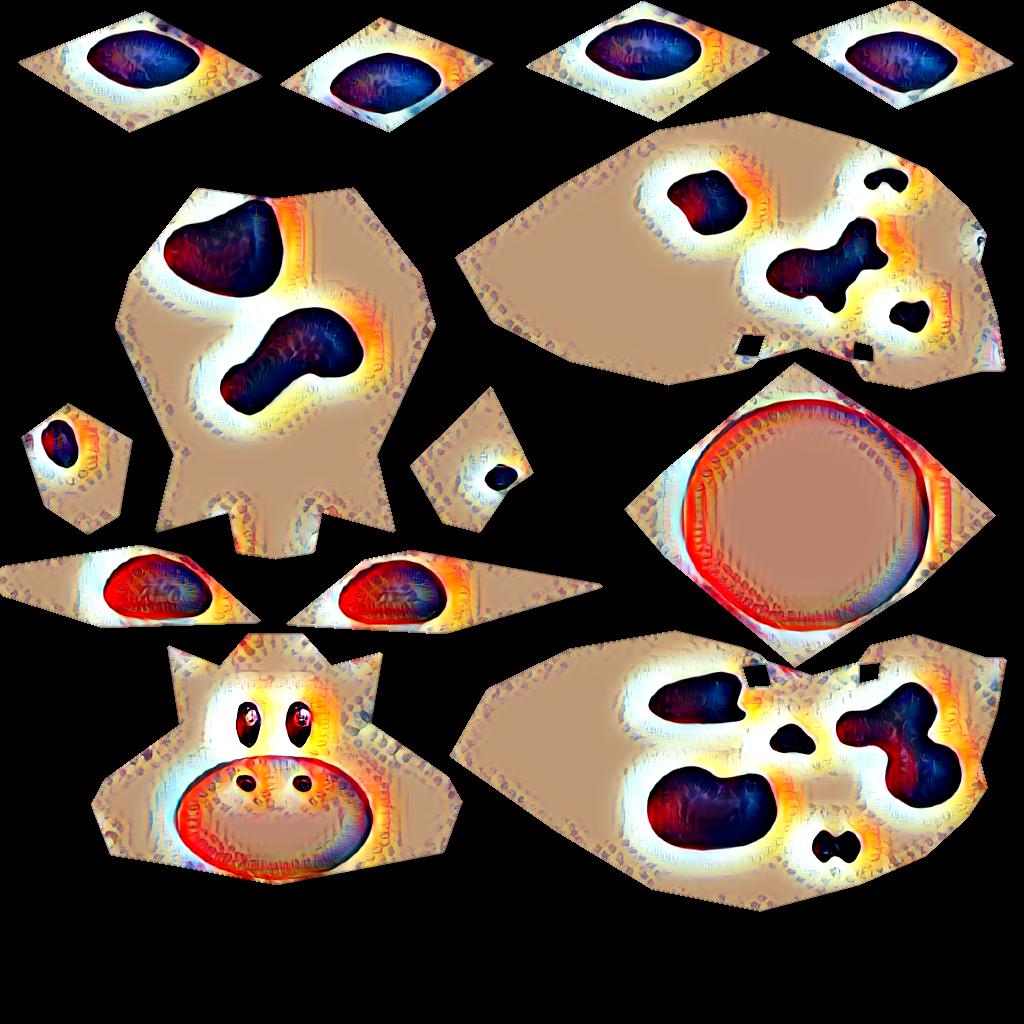} \\

e) Original texture map &
f) Naively stylized texture map &
g) Our method
\\
\\
\cropim{.20}{Figures/cow.png}{125}{360}{285}{100}
&
\cropim{.20}{Figures/cow0naive.png}{125}{360}{285}{100}&
\cropim{.20}{Figures/cow0ours.png}{125}{360}{285}{100} \\

\cropim{.20}{Figures/cow.png}{225}{0}{200}{450}
&
\cropim{.20}{Figures/cow0naive.png}{225}{0}{200}{450}&
\cropim{.20}{Figures/cow0ours.png}{225}{0}{200}{450} \\

h) Magnification (Original) &
i) Magnification (Naive) &
j) Magnification (Ours)

\end{tabular}
\vspace{-0.13in}
\end{center}
   \caption{ 
        A given style image (a), the original 3D mesh (b) and its texture map~(e), the result of naively stylizing the texture map~(c,f), and the result of using our method~(d,g). We provide magnifications of the results in two locations (h,i,j). Note the visible seams shown in the magnifications of the naive approach~(i), whereas our method in~(j) minimizes the visibility of those seams since the visible regions of the texture map can be interconnected in the graph-based approach.
   }
\label{fig:meshstyles3}
\end{figure}

\noindent
For spherical image and texture map stylization, we provide an expanded version of Figure 6 from the paper here for easier viewing (Fig.~\ref{fig:spherestyle}), and an additional spherical result in Fig.~\ref{fig:environment}. We also provide an enlarged version of Figure 10 from the paper for easier viewing (Fig.~\ref{fig:meshstyles-a}) and additional texture map results in Fig.~\ref{fig:meshstyles1} through Fig.~\ref{fig:meshstyles3}. 

~

\noindent
We also provide a supplementary video (which can be found on our \href{http://github.com/davidmhart/SelectionConv}{project website}) that includes 
visualizations of all the examples presented here.
This provides the best visualization of these results and their advantages over the naive approach.

~

\clearpage


